%% file: youtu_template.tex
\documentclass[]{youtu} 

\usepackage{mathpazo}
\usepackage{graphicx}
\usepackage{natbib}
\usepackage{multirow}
\usepackage{makecell}
\usepackage{booktabs}
\usepackage{array}
\usepackage[margin=1in]{geometry}
\usepackage{tabularx}
\usepackage{ragged2e}
\usepackage{multicol}
\usepackage{graphicx}
\usepackage{subcaption}

\usepackage[utf8]{inputenc}
\usepackage[T1]{fontenc}
\usepackage[margin=1in]{geometry}
\usepackage{xcolor}
\usepackage{tcolorbox}
\usepackage{fontawesome5}
\usepackage{enumitem}

\usepackage[most]{tcolorbox}
\usepackage{xcolor}
\usepackage{listings}
\usepackage{geometry}

\usepackage[T1]{fontenc}
\usepackage{courier}
\usepackage{amssymb}
\usepackage{amsmath}
\usepackage{amsthm}

\definecolor{thoughtbg}{HTML}{F4F4F9}
\definecolor{toolbg}{HTML}{E7F3FF}
\definecolor{obsbg}{HTML}{F0FFF4}
\definecolor{stepbg}{HTML}{4A4A4A}
\definecolor{analysisbg}{HTML}{FDF2F2}

\newtcolorbox{thoughtbox}{
    colback=thoughtbg, colframe=thoughtbg, arc=0pt, outer arc=0pt,
    boxrule=0pt, breakable, left=5pt, right=5pt, top=5pt, bottom=5pt, fontupper=\small
}

\newtcolorbox{toolbox}{
    colback=toolbg, colframe=toolbg, arc=0pt, outer arc=0pt,
    boxrule=0pt, breakable, left=5pt, right=5pt, top=5pt, bottom=5pt, fontupper=\small
}

\newtcolorbox{obsbox}{
    colback=obsbg, colframe=obsbg, arc=0pt, outer arc=0pt,
    boxrule=0pt, breakable, left=5pt, right=5pt, top=5pt, bottom=5pt, fontupper=\small
}

\newcommand{\stepheader}[1]{
    \begin{tcolorbox}[colback=stepbg, colframe=stepbg, sharp corners, 
    halign=center, valign=center, height=20pt,
    fontupper=\color{white}\bfseries]
    Step #1
    \end{tcolorbox}
}

\makeatletter
\renewcommand{\paragraph}{%
  \@startsection{paragraph}{4}{\z@}%
    {0.5ex \@plus1ex \@minus.2ex}
    {-1em}
    {\normalfont\normalsize\bfseries}%
}
\makeatother

\newcommand{\eg}{{\emph{e.g.}}, }

\title{Youtu-LLM: Unlocking the Native Agentic Potential for Lightweight Large Language Models}
\author{Youtu-LLM Team$^*$}

\sourcecode{https://github.com/TencentCloudADP/youtu-tip/blob/master/youtu-llm}
\model{https://huggingface.co/collections/tencent/youtu}
\project{https://youtu-tip.com/\#llm}
\youtufinalcopy 

\begin{document}

\abstract{

We introduce \textbf{Youtu-LLM}, a lightweight yet powerful language model that harmonizes high computational efficiency with native agentic intelligence. Unlike typical small models that rely on distillation, Youtu-LLM (1.96B) is pre-trained from scratch to systematically cultivate reasoning and planning capabilities. The key technical advancements are as follows: \textbf{(1) Compact Architecture with Long-Context Support:} Built on a dense Multi-Latent Attention (MLA) architecture with a novel STEM-oriented vocabulary, Youtu-LLM supports a 128k context window. This design enables robust long-context reasoning and state tracking within a minimal memory footprint, making it ideal for long-horizon agent and reasoning tasks. \textbf{(2) Principled "Commonsense-STEM-Agent" Curriculum} We curated a massive corpus of approximately 11T tokens and implemented a multi-stage training strategy. By progressively shifting the pre-training data distribution from general commonsense to complex STEM and agentic tasks, we ensure the model acquires deep cognitive abilities rather than superficial alignment. \textbf{(3) Scalable Agentic Mid-training:} Specifically for the agentic mid-training, we employ diverse data construction schemes to synthesize rich and varied trajectories across math, coding, and tool-use domains. This high-quality data enables the model to internalize planning and reflection behaviors effectively. Extensive evaluations show that \textbf{Youtu-LLM sets a new state-of-the-art for sub-2B LLMs}. On general benchmarks, it achieves competitive performance against larger models, while on agent-specific tasks, it significantly surpasses existing SOTA baselines, demonstrating that lightweight models can possess strong intrinsic agentic capabilities.

}

\maketitle

\renewcommand{\thefootnote}{*}
\footnotetext{Full author list in contributions.}
\renewcommand{\thefootnote}{\arabic{footnote}}

\input{0_titlepage}
\input{1_introduction}
\input{2_pretrain_data}
\input{3_pretrain_method}
\input{4_posttrain}
\input{5_experiment}

\section{Conclusion}

In this work, we introduce Youtu-LLM, a lightweight 1.96B-parameter language model that effectively unlocks native agentic potential. It successfully balances computational efficiency with deep reasoning capabilities, proving highly effective for complex agent tasks. Most notably, the integration of our scalable Agentic Mid-training paradigm enables the model to internalize planning and reflection behaviors. Extensive evaluations confirm that Youtu-LLM sets a new state-of-the-art for sub-2B models. Remarkably, it even surpasses larger LLMs like SmolLM3-3B and Qwen3-4B on challenging agentic benchmarks, demonstrating robust agentic capabilities in domains including code, deep research and math.

Despite promising, we acknowledge certain limitations when compared to larger foundation models, which guide our future research roadmap: First, constrained by computational resources, a gap in agentic capabilities remains compared to large proprietary LLMs. We aim to bridge this by evolving our model into a world model that simulates execution dynamics to ground reasoning in a robust understanding of the environment. Second, model efficiency remains a challenge, as long reasoning trajectories inevitably increase inference latency. Future work will explore more efficient model architecture, such as Diffusion LLMs, to optimize inference speed. Third, our current scope is confined to text-based environments. We intend to extend our model to multimodal scenarios, equipping the model with native omni-modal perception for complex real-world tasks.

\clearpage
\setcitestyle{numbers,square}
\setcitestyle{square,numbers,comma}
\bibliography{youtu_bib}

\addcontentsline{toc}{section}{Appendix} 

\clearpage
\appendix
\section*{Appendix}

\section{Contributions and Acknowledgments}
We would like to express our sincere gratitude to all contributors, including those not listed in the paper, for their invaluable support and efforts. The contributors within each group are listed in no particular order.

\textbf{Core Contributors}\\
Junru Lu\textsuperscript \quad 
Jiarui Qin\textsuperscript \quad 
Lingfeng Qiao\textsuperscript \quad 
Yinghui Li\textsuperscript \quad 
Xinyi Dai\textsuperscript \quad 
Bo Ke\textsuperscript \quad 
Jianfeng He\textsuperscript \quad 
Ruizhi Qiao\textsuperscript \quad 
Di Yin\textsuperscript \quad \\Xing Sun$^\dagger$\textsuperscript \quad 
Yunsheng Wu\textsuperscript \quad
\noindent\let\thefootnote\relax\footnotetext{$^\dagger$Corresponding author: winfredsun@tencent.com}

\textbf{Contributors}\\
Yinsong Liu\textsuperscript \quad
Shuangyin Liu\textsuperscript \quad 
Mingkong Tang\textsuperscript \quad 
Haodong Lin\textsuperscript \quad 
Jiayi Kuang\textsuperscript \quad 
Fanxu Meng\textsuperscript \quad 
Xiaojuan Tang\textsuperscript \quad 
\\Yunjia Xi\textsuperscript \quad 
Junjie Huang\textsuperscript \quad 
Haotong Yang\textsuperscript \quad 
Zhenyi Shen\textsuperscript \quad 
Yangning Li\textsuperscript \quad 
Qianwen Zhang\textsuperscript \quad 
Yifei Yu\textsuperscript \quad 
Siyu An\textsuperscript \quad 
\\Junnan Dong\textsuperscript \quad 
Qiufeng Wang\textsuperscript \quad 
Jie Wang\textsuperscript \quad 
Keyu Chen\textsuperscript \quad 
Wei Wen\textsuperscript \quad 
Taian Guo\textsuperscript \quad 
Zhifeng Shen\textsuperscript \quad 
Daohai Yu\textsuperscript \quad 
\\Jiahao Li\textsuperscript \quad
Ke Li\textsuperscript \quad 
Zongyi Li\textsuperscript \quad 
Xiaoyu Tan\textsuperscript \quad
Youtu-LLM Team\textsuperscript \quad

\section{The Impact of Agentic-CoT Trajectory Data}
\label{app:agent_reasoning}

\paragraph{Ablation Study of Agentic-CoT Trajectory Data}

\begin{figure}[h]
    \centering
    \includegraphics[width=0.95\linewidth]{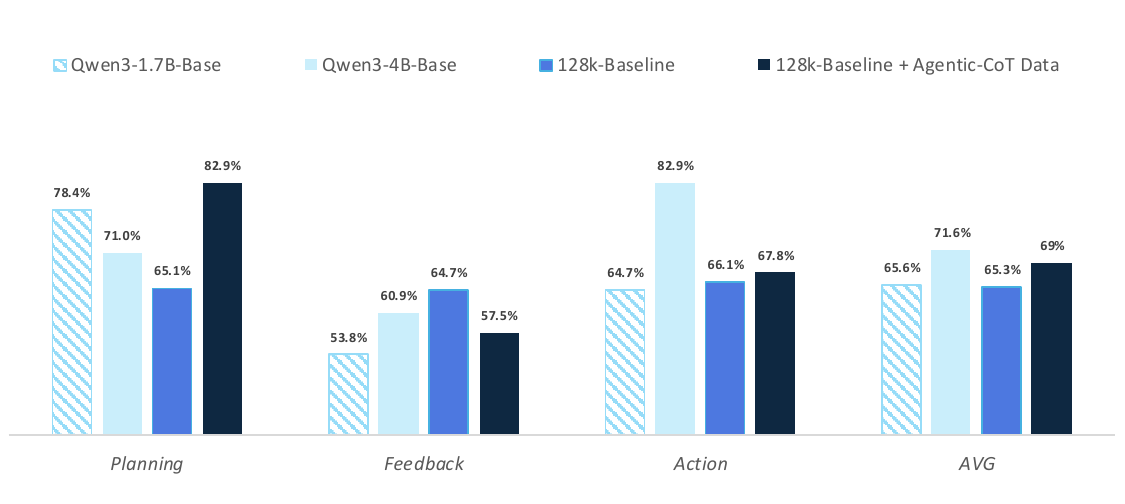}
    \caption{Ablation experimental results of agentic-CoT trajectory data on APT-Math.}
    \label{fig:reason_ablation}
\end{figure}

To verify the specific contribution of our proposed Agentic-CoT trajectory data, we conducted an ablation study comparing the performance of the Baseline model, which is trained without any agentic trajectory data, against the model with mixed training on Agentic-CoT trajectory data. We evaluated the models across three key dimensions of APT-Math: Planning, Feedback, and Action, alongside the overall average performance (AVG). For reference, the performance of the base models (Qwen3-1.7B-Base and Qwen3-4B-Base) is also included to contextualize the improvements.

As illustrated in the Figure \ref{fig:reason_ablation}, the integration of Agentic-CoT Trajectory Data yields a significant performance improvement. The most notable enhancement is observed in the Planning capability, where the score surged from 65.1\% (Baseline) to 82.9\%, surpassing even the larger base models. While there is a slight fluctuation in the Feedback metric, the Action metric showed a steady increase from 66.1\% to 67.8\%. Consequently, the overall average score increased from 65.3\% to 69.0\%, resulting in a net performance gain of 3.7\%.

The substantial uplift in the Planning dimension indicates that the agentic-CoT Trajectory Data effectively enhances the model's ability to decompose complex tasks and formulate logical execution steps. By incorporating reflection trajectories, the model learns to "think before acting," which significantly improves the quality of its initial plans. This strong planning capability acts as the primary driver for the overall performance, leading to the observed 3.7\% increase in the average score. This confirms that equipping the model with reflective reasoning paths is crucial for solving complex tasks that require long-horizon planning.

\section{The Impact of Math Trajectory Data}\label{app:agent_math}

In the ablation studies of the math trajectory, the starting baseline is further trained using \textbf{40B tokens of 32k-length general-purpose data} (including a portion of general trajectory data) and \textbf{3.3B tokens of 32k-length mathematical trajectory data}. All training settings are kept identical to those used in the full trajectory pre-training, ensuring a fair comparison. We design three categories of ablation experiments to analyze: 
(1) the effect of different training strategies, 
(2) the influence of trajectory data scaling.

\subsection{Different Masking Strategies Analysis}

To avoid the impact of numerous repetitive system prompts and instructions in the trajectory, we mask the trajectory during training. We first examine how different masking strategies applied to trajectory training affect model performance. Beyond the baseline model, we explore several training configurations, including: 
(i) training on general data without masking, 
(ii) mixed training on general-purpose data and mathematical trajectory data without masking, and 
(iii) mixed training with masking applied to both general-purpose and mathematical trajectory data.
In the masking-based setups, we conduct a universal masking strategy to maintain consistency as much as possible with trajectory data from other domains. All fields except those with the \texttt{"role":"assistant"} are masked. We further investigate whether fully masking the user query introduces additional effects. 

\begin{figure}[h]
    \centering
    \includegraphics[width=0.95\linewidth]{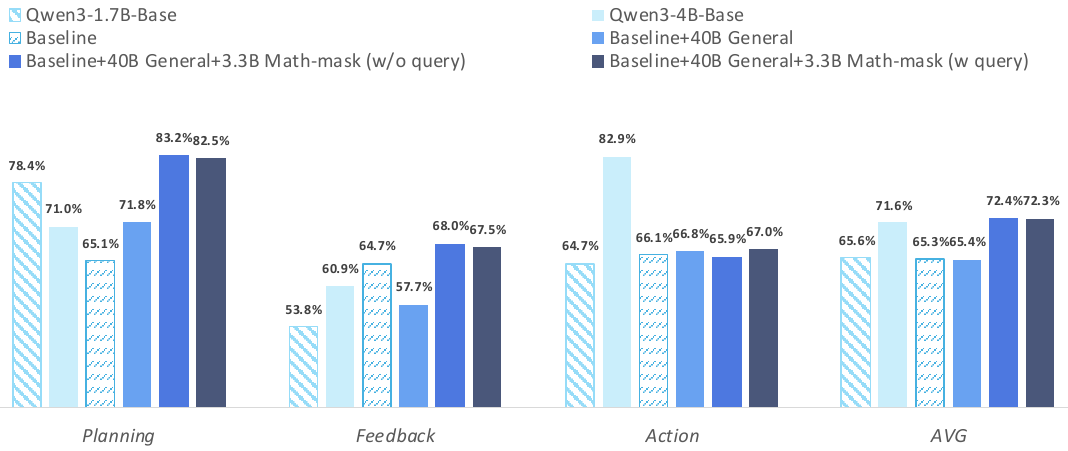}
    \caption{Ablation experimental results for different mask training strategies of math trajectories.}
    \label{fig:math_mask}
\end{figure}

Several important observations can be drawn from these results, as shown in Figure \ref{fig:math_mask}.
First, incorporating mathematical trajectory data consistently improves overall performance compared to the baseline, confirming the effectiveness of trajectory-level supervision for enhancing agentic mathematical reasoning. 
Second, applying masking strategies yields substantially larger gains than unmasked training, particularly on \textbf{Planning} and \textbf{Feedback} tasks. This suggests that masking non-assistant fields encourages the model to internalize reasoning patterns and solution trajectories. In contrast, the gains on \textbf{Action} tasks are more moderate, implying that computational execution benefits less from trajectory abstraction than high-level reasoning and verification. Finally, we observe that whether the user query is fully masked or not has only a marginal impact on average performance. This suggests that the primary benefit of masking arises from different agentic actions, while preserving or removing the query itself plays a secondary role. Overall, these findings highlight masking-aware trajectory pre-training as a key design choice for effectively injecting agentic reasoning capabilities into lightweight language models.

\begin{figure}
    \centering
    \includegraphics[width=0.95\linewidth]{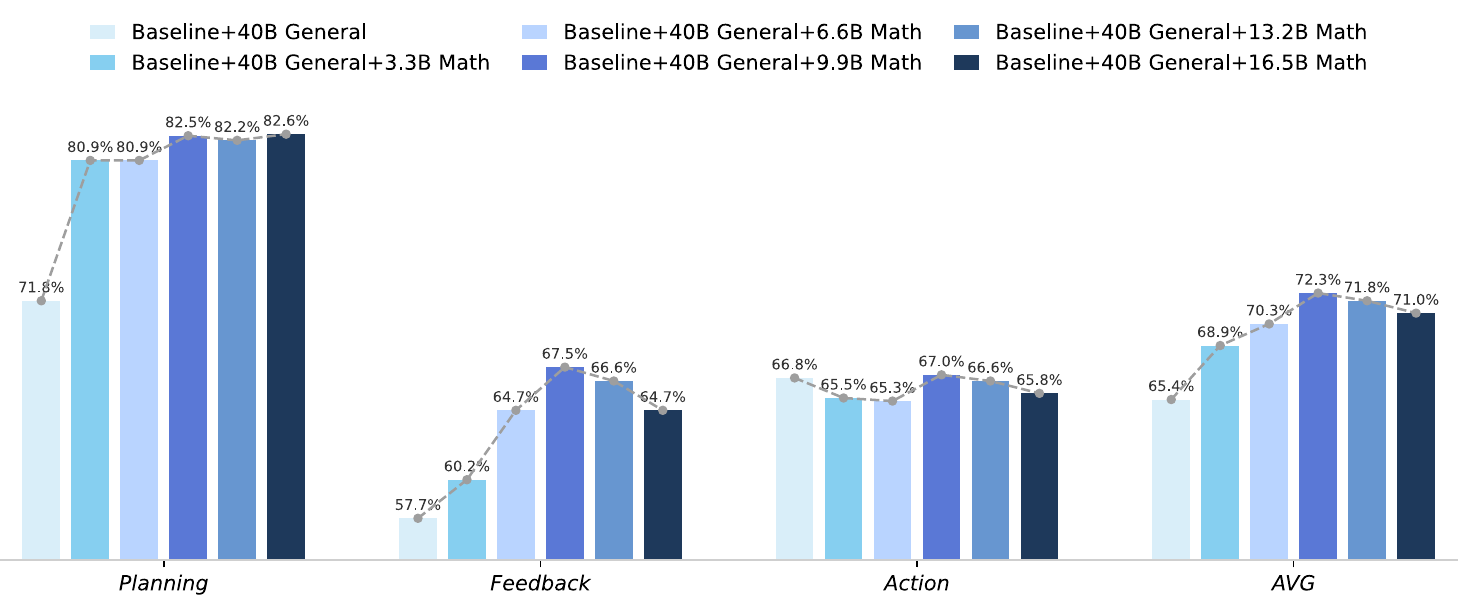}
    \caption{Ablation experiments results of mathematical trajectory training at different up-sampling rates.}
    \label{fig:math_scale}
\end{figure}

\subsection{Math Trajectory Scaling Analysis}
\label{sec:agent_math}

After exploring the optimal masking strategy, we further investigate the \textbf{scaling behavior} of mathematical trajectory data. Starting from the \textbf{3.3B-token} mathematical trajectory dataset used in the ablation stage, we progressively upsample the data by factors of \textbf{1$\times$, 2$\times$, 3$\times$, 4$\times$, and 5$\times$}, resulting in a maximum of \textbf{16.5B tokens} of mathematical trajectories. These trajectories are mixed with a fixed \textbf{40B-token} general-purpose dataset and trained under the same configuration as in previous experiments.

From the results in Figure \ref{fig:math_scale}, several notable observations emerge from this scaling study. 
First, increasing the amount of mathematical trajectory data leads to \textbf{consistent and substantial performance gains} over the baseline models, validating the effectiveness of trajectory-level supervision for strengthening agentic mathematical reasoning. Even a single upsampling ($1\times$) yields a clear improvement, particularly in \textbf{Planning}, suggesting that mathematical trajectories rapidly inject high-level structural reasoning patterns into the model. Performance continues to improve as the trajectory scale increases from $1\times$ to $3\times$, which indicates a \textbf{scaling law–like behavior} in which larger volumes of trajectory data progressively enhance the model’s ability to plan solution strategies and verify multi-step reasoning processes. In addition, the results reveal a clear \textbf{performance saturation point}. While the $5\times$ configuration slightly improves planning accuracy, it leads to degradation in feedback and action performance. This suggests that excessive upsampling of domain-specific trajectory data may introduce distributional imbalance, potentially overwhelming general-purpose knowledge and harming cross-capability robustness.

\section{The Impact of Deep Research Trajectory Data}
\label{app:agent_deepresearch}

\subsection{Different Masking Strategies Analysis}

In deep research scenarios, trajectory data often contains significant noise and redundancy, particularly in the form of repetitive system prompts and verbose search engine responses. Blindly learning these segments can negatively impact the model's instruction-following and reasoning capabilities by introducing distribution shifts or encouraging the generation of non-informative tokens. Therefore, defining an optimal masking strategy is crucial before formal training. We investigate three distinct masking configurations to determine the most effective approach:

\begin{itemize}
    \item Baseline + DR data without mask: Training on Deep Research (DR) data with full-text loss calculation.
    \item Baseline + DR data with mask (tool response): Masking the search engine/tool outputs, which are not calculated into the generation loss.
    \item Baseline + DR data with mask (tool response/system/query): A comprehensive masking strategy that excludes tool responses, system prompts, and user queries, focusing the loss exclusively on the model's internal reasoning and action generation.
\end{itemize}

\begin{figure}[h]
    \centering
    \includegraphics[width=0.95\linewidth]{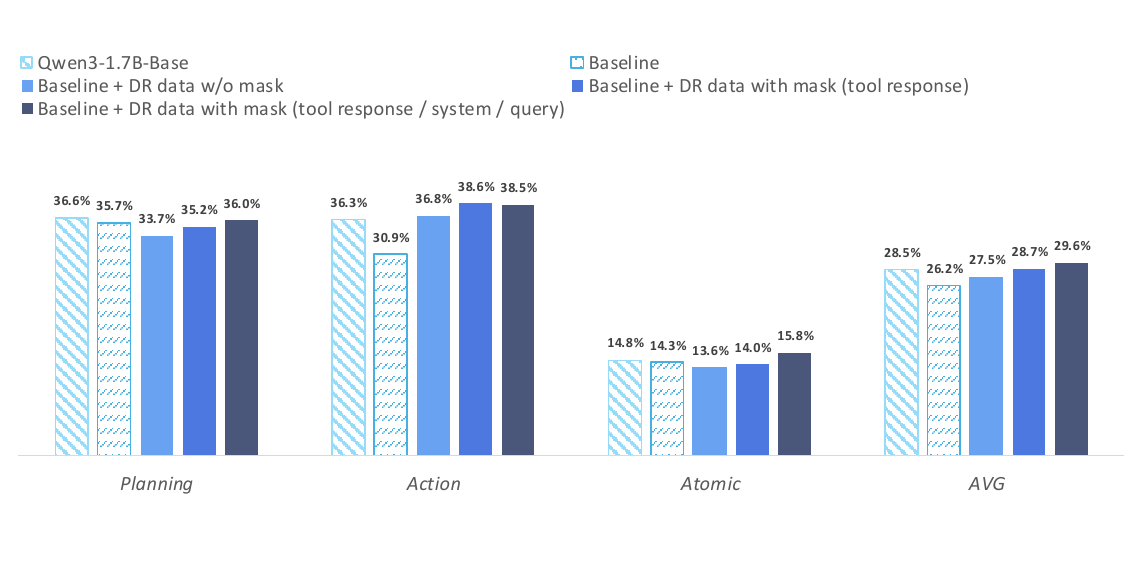}
    \vspace{-1.0em}
    \caption{Ablation experimental results for different mask training strategies of deep research.}
    \label{fig:dr_mask}
\end{figure}

The performance of these strategies was evaluated across three key dimensions: Planning, Action, and Atomic capabilities, along with an average score on APT-DR. The results are illustrated in the Figure \ref{fig:dr_mask}. Training without masking yields mixed results. While it improves the Action score (36.8\%) compared to the Baseline (30.9\%), it causes a significant regression in Planning capabilities, dropping from the Baseline's 35.7\% to 33.7\%. This suggests that forcing the model to predict noisy search results and repetitive prompts interferes with its high-level planning logic. By masking the tool responses, we observe an immediate improvement. This strategy outperforms the unmasked approach in both Planning (35.2\%) and Action (38.6\%). Notably, the Action score sees a substantial leap, surpassing the Qwen3-1.7B-Base model (36.3\%), indicating that the model learns to utilize tools better when it is not penalized for failing to predict the tool's output perfectly. The most aggressive masking strategy (masking tool response, system, and query) achieves the best overall performance. It yields the highest AVG score of 29.6\%, significantly outperforming the Baseline (26.2\%) and the unmasked variant (27.5\%). Crucially, this is the only strategy that achieves a substantial gain in the Atomic metric (15.8\%) compared to the Baseline (14.3\%) and the Base model (14.8\%).

The experimental data clearly demonstrate that selective loss calculation is essential for leveraging Deep Research trajectory data effectively. 

\noindent \textit{Noise Reduction.} The drop in Planning performance for the unmasked model confirms that verbose search results act as noise. When the model attempts to learn these unpredictable or redundant sequences, its ability to structure complex reasoning paths degrades.

\noindent \textit{Focus on Reasoning.} The comprehensive masking strategy forces the model to focus solely on the "Thought" and "Action" components—the actual reasoning trace. By masking the system prompts and user queries, we prevent the model from overfitting to specific prompt templates, while masking tool responses prevents the hallucination of search results.

Consequently, the strategy of masking tool responses, system prompts, and user queries is adopted as the standard for subsequent training, as it maximizes the extraction of reasoning capabilities from the trajectory data while minimizing the interference of environmental noise.

\subsection{Ablation Study of Deep Research Trajectory Data}

\begin{figure}[h]
    \centering
    \includegraphics[width=0.95\linewidth]{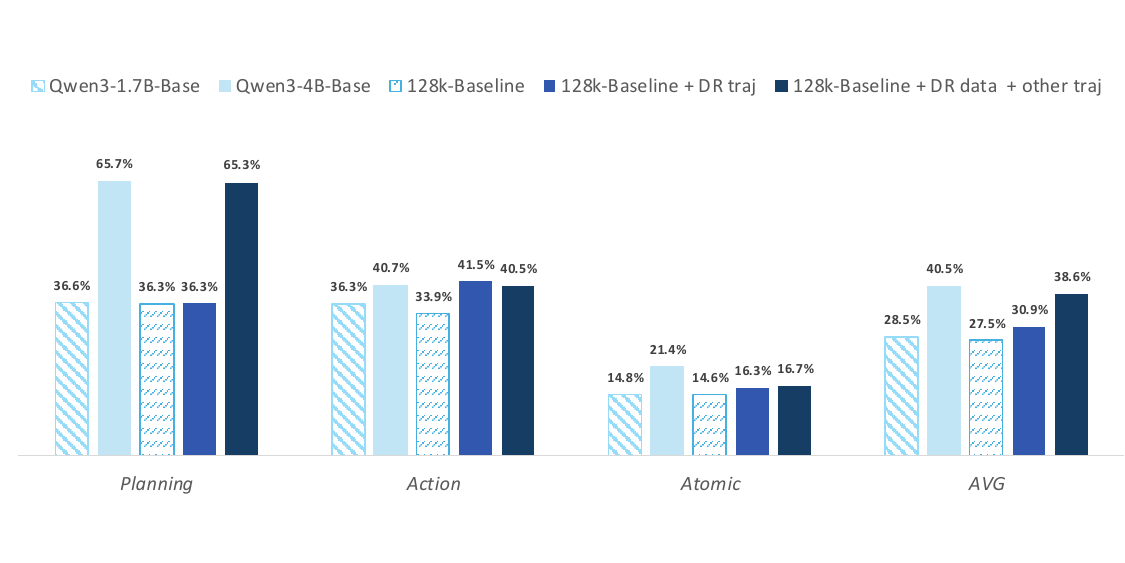}
    \caption{Ablation experimental results of deep research trajectory data on APT-DR.}
    \label{fig:dr_ablation}
\end{figure}

To evaluate the contribution of our proposed deep research trajectory data to model performance, we conducted an ablation study using the APT-DR evaluation set. We compare Qwen3-1.7B-Base, Qwen3-4B-Base, and several Youtu-LLM models with different data setups. We introduced the data in two stages: first by incorporating the Deep Research (DR) Trajectory Data alone, and subsequently by combining it with other Trajectory Data to observe potential synergistic effects. As illustrated in the Figure \ref{fig:dr_ablation}, the inclusion of Deep Research Trajectory Data yields a clear performance gain.

\noindent \textit{Impact of DR Data.} Adding the DR trajectory data to the Baseline results in a tangible improvement in the overall score, raising the AVG from 27.5\% to 30.9\%. Notably, the Action metric sees a significant boost, increasing from 33.9\% to 41.5\%, surpassing even the Qwen3-4B-Base in this specific dimension.

\noindent \textit{Synergy with Other Trajectory Data.} The further addition of other trajectory data triggers a substantial leap in performance. The Planning metric exhibits a dramatic increase, surging to 65.3\%, which is nearly identical to the 65.7\% achieved by the significantly larger Qwen3-4B-Base. Consequently, the final AVG score reaches 38.6\%, narrowing the performance gap with the 4B model to less than 2 percentage points.

The experimental results indicate two critical findings. First, the Deep Research Trajectory Data is highly effective in enhancing the model's execution capabilities, as evidenced by the sharp rise in the Action metric. Second, there is a strong mutual promotion (synergy) between the DR data and other trajectory datasets. While DR data solidifies the model's action execution, the supplementary trajectory data appears to unlock the model's reasoning and planning potential. This combination allows our 1.7B parameter model to achieve a comprehensive performance level (AVG=38.6\%) that closely approximates that of the Qwen3-4B-Base (AVG=40.5\%), demonstrating the high efficiency and quality of our data strategy.

\section{The Impact of Code Trajectory Data}
\label{app:agent_code}

\begin{figure}[h]
    \centering
    \includegraphics[width=0.80\linewidth]{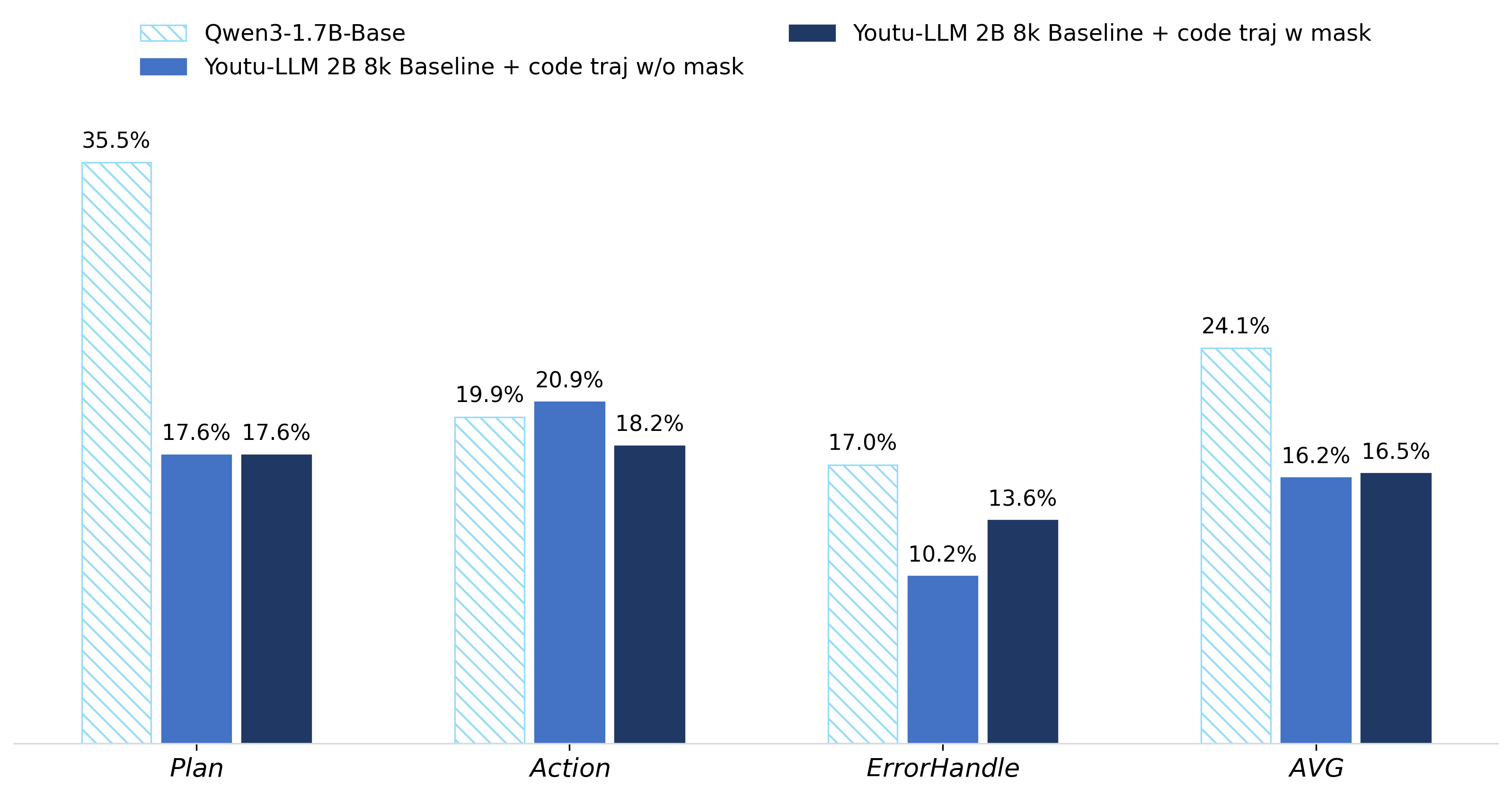}
    \vspace{-1.0em}
    \caption{Ablation experimental results for different mask training strategies code trajectory data (EnvSetup).}
    \label{fig:code_mask_1}
\end{figure}

\begin{figure}[h]
    \centering
    \includegraphics[width=0.80\linewidth]{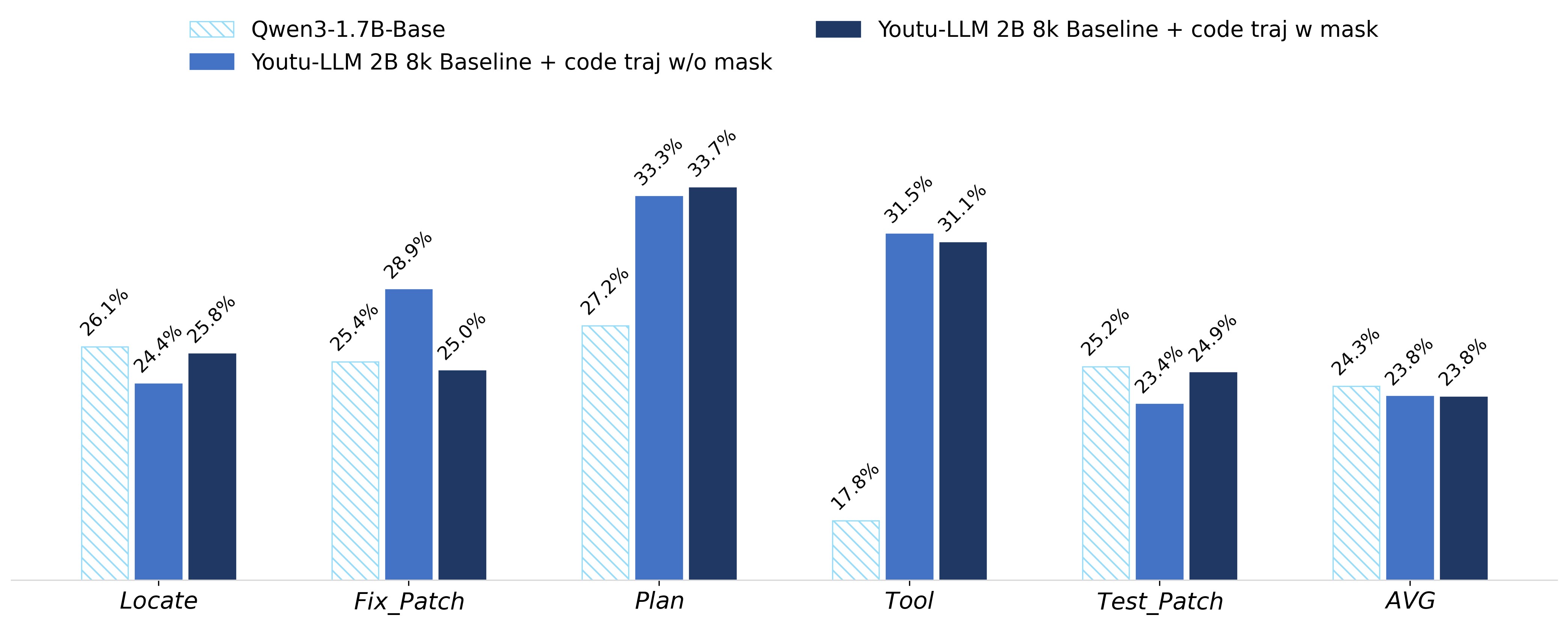}
    \vspace{-1.0em}
    \caption{Ablation experimental results for different mask training strategies of code trajectory data (IssueFix).}
    \label{fig:code_mask_2}
\end{figure}

\subsection{Different Masking Strategies Analysis}
We first conduct a preliminary study using a small subset of code trajectory data based on the baseline model to investigate the effects of masking non-assistant turns. As shown in Figure~\ref{fig:code_mask_1} and Figure~\ref{fig:code_mask_2}, the performance gap is marginal on both \textbf{EnvSetup} and \textbf{Issuefix} subset. 

Learning from supervised signals from non-assistant turns facilitates the acquisition of \textit{code-centric world knowledge}, thereby improving the model's ability to interpret environment feedback. However, this approach may introduce \textit{distributional inconsistencies} or incorporate stochastic noise and redundant patterns, which could potentially undermine training stability. To maintain methodological alignment with our other trajectory experiments, we opted to mask non-assistant content in our final experimental setup. Nevertheless, developing a more nuanced approach to learning from non-assistant segments that leverages environmental insights without compromising data quality is still a promising avenue for future exploration.
\begin{figure}[h]
    \centering
    \includegraphics[width=0.9\linewidth]{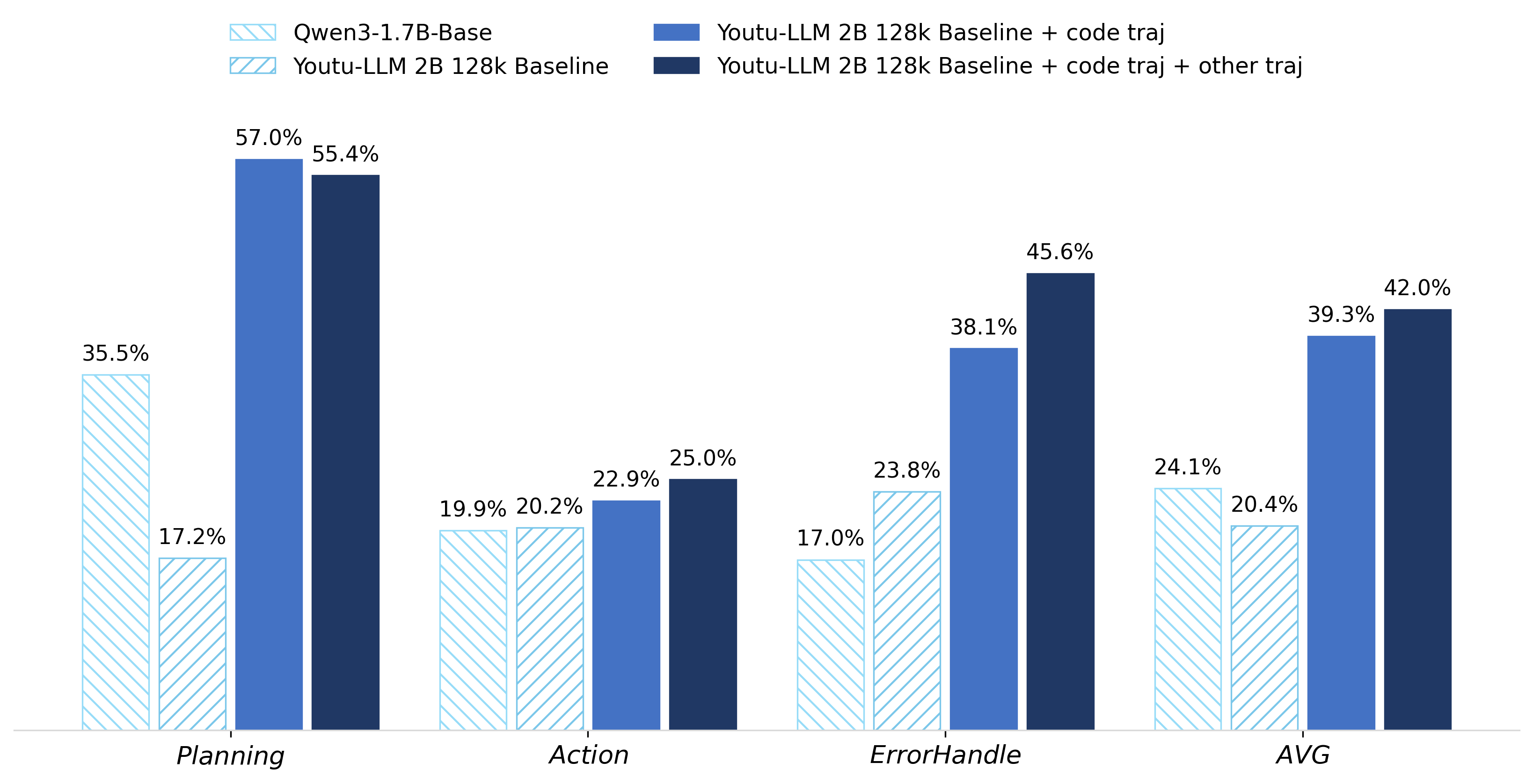}
    \caption{Ablation experimental results of code trajectory data on APT-Code-EnvSetup.}
    \label{fig:code_ablation_1}
\end{figure}

\begin{figure}[h]
    \centering
    \includegraphics[width=0.95\linewidth]{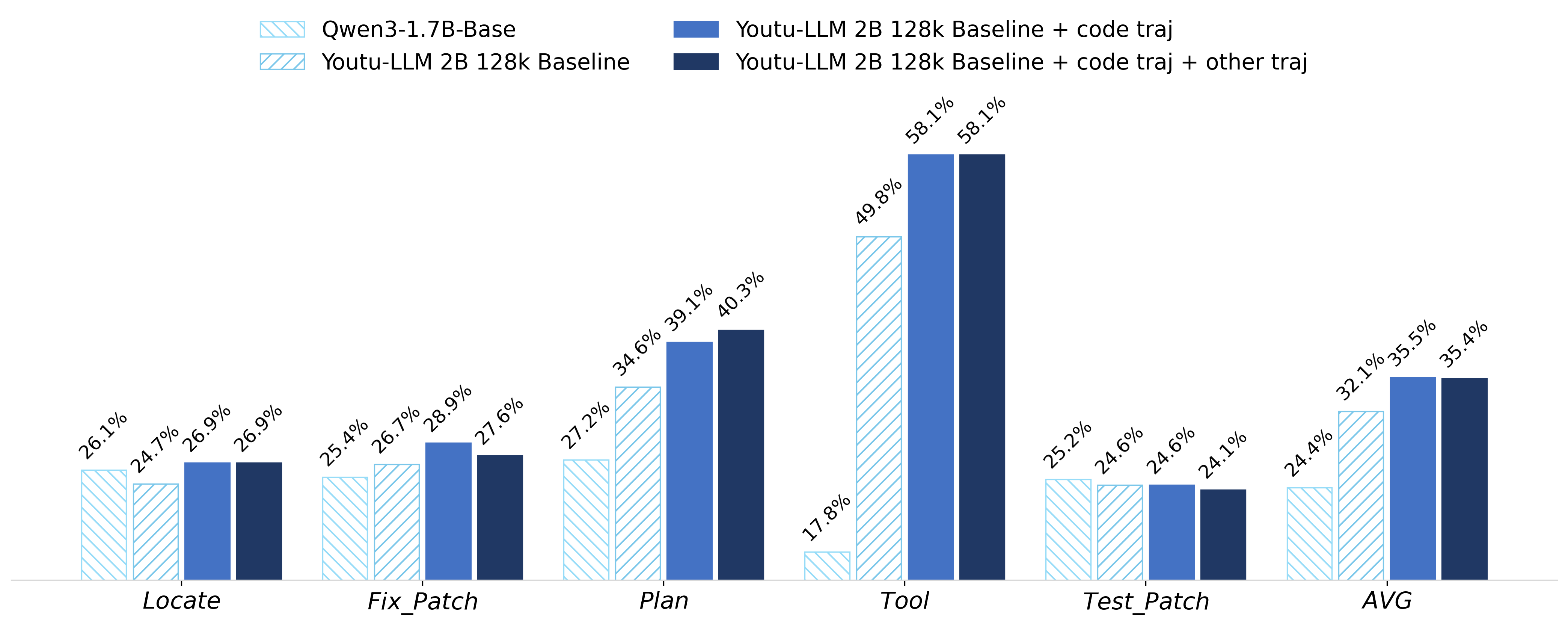}
    \caption{Ablation experimental results of code trajectory data on APT-Code-IssueFix.}
    \label{fig:code_ablation_2}
\end{figure}

\subsection{Ablation Study of Code Trajectory Data}

To further investigate the impact of code trajectory data on agentic code capabilities, we conduct an ablation study based on the 128k baseline. Figure~\ref{fig:code_ablation_1} and Figure~\ref{fig:code_ablation_2} illustrates the results of the EnvSetup subset and IssueFix subset, respectively.

In the \textbf{EnvSetup} subset, the Qwen3-4B-Base model exhibits a clear superiority, particularly in Planning (71.8\%) and ErrorHandle (61.9\%). This suggests that the larger parameter count of the 4B model provides a stronger foundational capability for logical sequencing and recovering from setup exceptions. A significant observation can be made regarding the Youtu-LLM 2B. The "Baseline" starts at a modest 20.4\% AVG. However, the integration of code trajectories (+ code traj) boosts the average to 39.3\%, and further adding other trajectories (+ other traj) raises it to 42.0\%. While Youtu-LLM lags behind Qwen3-4B in Planning, it shows competitive performance in \textit{Action} and \textit{ErrorHandle} after data enhancement, indicating that domain-specific trajectories significantly narrow the gap between a 2B model and a larger 4B model.

Unlike the \textbf{EnvSetup} subset, the \textbf{IssueFix} results show a much tighter competition. Notably, the Youtu-LLM 2B (with trajectories) actually outperform Qwen3-4B-Base in the AVG metric (35.5\% vs. 33.9\%). The most striking disparity is found in the \textit{Tool} category. The Youtu-LLM variants (with code trajectories) achieve an impressive 58.1\%, significantly outstripping both Qwen3-1.7B (17.8\%) and Qwen3-4B (47.7\%). This suggests that the trajectory-based training specifically enhances the model's ability to interface with external debugging and diagnostic tools.
Across \textit{Locate}, \textit{Fix\_Patch}, and \textit{Test\_Patch}, the performance remains relatively stable across all models, hovering between 24\% and 29\%, indicating that while "fixing" the code is a shared challenge for small LLMs, "tooling" and "planning" are the primary differentiators for overall success.

While Qwen3-4B benefits from its larger scale to lead in general planning and complex environment reasoning, the Youtu-LLM 2B demonstrates that specialized trajectory data can allow a smaller model to achieve SOTA performance in technical workflows like issue fixing. The incremental gains in the Youtu-LLM series confirm that the quality and variety of training trajectories (code vs. others) are critical for "Action" and "Tool" oriented tasks, which are more execution-heavy than purely generative. Qwen3-4B remains the most robust "all-rounder" for environment setup, but for developers focusing on automated bug fixing and tool integration, the enhanced Youtu-LLM 2B offers a more efficient and higher-performing alternative.

\section{The Impact of Tool-use Trajectory Data}
\label{app:agent_tool}

\subsection{Different Masking Strategies Analysis}

\begin{figure}[h]
    \centering
    \includegraphics[width=0.95\linewidth]{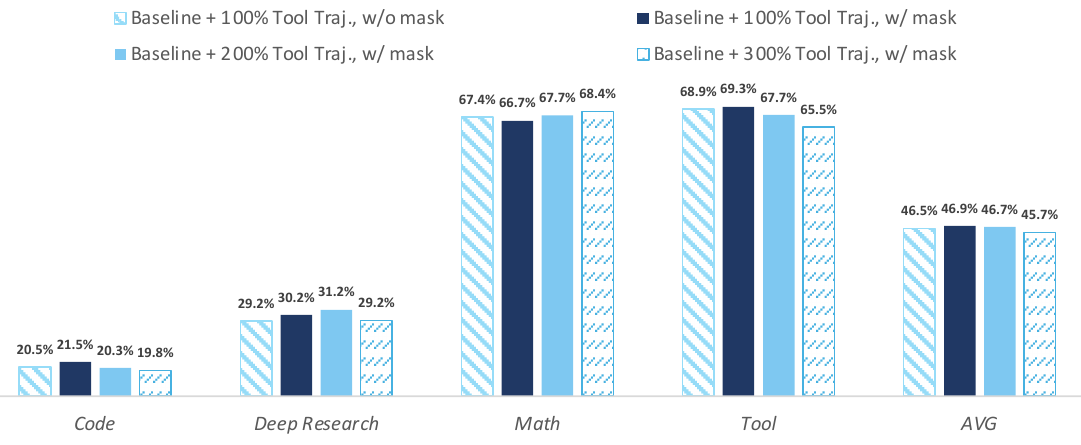}
    \caption{Ablation experimental results of tool-use trajectory data, concerning masking and up-sampling strategy. The metrics reported are the sub-scores of four evaluation dimensions in APTBench, as well as the overall average score.}
    \label{fig:tool_ablation_1}
\end{figure}

In Figures\,\ref{fig:tool_ablation_1} and \,\ref{fig:tool_ablation_2}, we report preliminary experiments on tool-use trajectory data, following the experimental setup adopted in the ablation studies of other types of trajectory data. We first examine the impact of masking strategy, comparing the effects of agentic mid-training with and without masks (on all non-assistant input) at 100\% sampling. The results in Figure\,\ref{fig:tool_ablation_1} shows that the difference between the two is marginal, with the version using masks showing a slight improvement of 0.4\% in the average APTBench score.

\subsection{Ablation Study of Tool-use Trajectory Data}

Regarding the sampling rate, we experimented with sampling rates 100\% (\textbf{$1\times$}), 200\% (\textbf{$2\times$}), and 300\% (\textbf{$3\times$}). Under the same training base and unified experimental settings, Figure\,\ref{fig:tool_ablation_1} shows that the \textbf{$1\times$} sampling rate is more suitable for tool-use and planning trajectory data. When the sampling rate increased to \textbf{$2\times$}, the correlated metrics for deep research improved, but significantly decreased at \textbf{$3\times$}. In terms of mathematics, using more tool-related data consistently improved the metrics. However, for the evaluation of code and tool, increasing the sampling rate continuously impaired the model's agentic capabilities. Therefore, we ultimately chose the \textbf{$1\times$} sampling rate for tool-use and planning trajectory data.

\begin{figure}[h]
    \centering
    \includegraphics[width=0.9\linewidth]{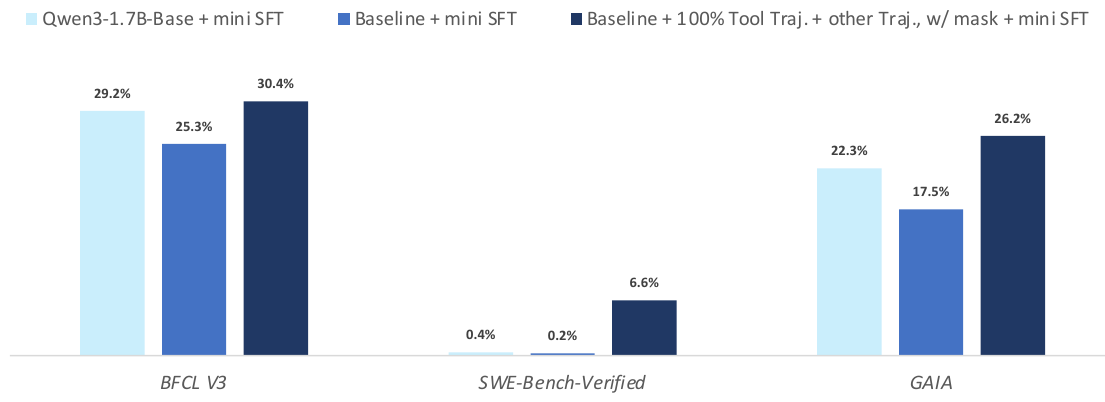}
    \caption{Ablation experimental results of mini SFT over less-agentic and agentic base models.}
    \label{fig:tool_ablation_2}
\end{figure}

\noindent \textit{How can we verify the effectiveness of agentic mid-training?} We designed a set of small-scale experiments. As shown in Figure \ref{fig:tool_ablation_2}, we performed mini SFT using 5B data based on Qwen3-1.7B-Base, the original pre-stage3 Youtu-LLM baseline, and the pre-stage3 Youtu-LLM baseline after agentic mid-training. This SFT data included a small amount of format instruction following data that is necessary for GAIA testing, as well as the vast majority of general thinking or non-thinking SFT examples. We conducted tests on BFCL V3~\citep{patil2025bfcl}, SWE-Bench-Verified~\citep{jimenez2023swe}, and GAIA\,\cite{mialon2023gaia}, and the results showed that the SFT performance of the base model after agentic mid-training was significantly better than other comparison models (e.g, over 6\% gains in SWE-Bench-Verified, and 4\% improvement in GAIA).

\section{Case Study}

\subsection{Case of Math}

\begin{tcolorbox}[colback=white, colframe=black, title=Analysis, fonttitle=\bfseries]
This case highlights our model’s \textbf{agentic mathematical reasoning ability} beyond end-to-end problem solving.
Instead of performing a naive forward derivation, the model first \textbf{\textcolor{blue}{identifies the core mathematical ambiguity}} (the definition of ``number ring'') and \textbf{\textcolor{blue}{actively switches to a counterexample-driven strategy}}.
By constructing minimal counterexamples such as $\mathbb{Z}[\sqrt{3}]$ and $\mathbb{Q}(\sqrt{3})$, the model demonstrates strong capabilities in \textbf{planning atomic abilities}, \textbf{self-verification}, and \textbf{robust logical action}, avoiding common reasoning traps caused by imprecise terminology.
\end{tcolorbox}

\begin{tcolorbox}[
    colback=white,
    colframe=black,
    boxrule=0.8pt,
    breakable,
    rounded corners
]

\textbf{Question:}  
Please judge whether the following statement is true or false:  
\emph{The smallest number ring containing $\sqrt{3}$ is the real number ring.}

\stepheader{Analysis: Agentic Problem Understanding}
\begin{thoughtbox}
Before attempting any algebraic manipulation, we performs a \textbf{global semantic analysis} of the problem.  
It identifies two key properties that fundamentally shape the solution strategy:

\begin{itemize}
    \item The phrase \emph{``smallest number ring''} implies a \textbf{minimality constraint}, where direct forward derivation toward $\mathbb{R}$ is unlikely to be optimal.
    \item The correctness of the statement depends on the existence of \textbf{strictly smaller algebraic structures} containing $\sqrt{3}$.
\end{itemize}

Rather than proceeding with routine symbolic computation, we concludes that:
\begin{center}
\textbf{\textcolor{blue}{A counterexample-oriented reasoning trajectory is the most effective strategy.}}
\end{center}

\end{thoughtbox}

\stepheader{Planning: Strategy Selection}
\begin{toolbox}
Based on the analysis, we formulates a high-level plan:

\begin{itemize}
    \item Avoid proving that $\mathbb{R}$ contains $\sqrt{3}$, as this is trivial and non-informative.
    \item Actively search for a \textbf{minimal algebraic structure} that already contains $\sqrt{3}$.
    \item Use the existence of such a structure as a \textbf{direct falsification} of the statement.
\end{itemize}

\textbf{Planning Decision:}  
Adopt a \textbf{counterexample-first strategy} by constructing smaller candidate rings or fields.
\end{toolbox}

\stepheader{Feedback: Definition Verification}
\begin{obsbox}
Before execution, we performs an internal consistency check on the problem definition:

\begin{itemize}
    \item The term ``number ring'' is ambiguous and may refer to either a \textbf{ring} or a \textbf{field}.
    \item The final judgment must therefore be valid under \textbf{all reasonable interpretations}.
\end{itemize}

This feedback step \textcolor{blue}{prevents overfitting to a single definition and ensures robustness of the reasoning trajectory}.
\end{obsbox}

\stepheader{Action: Counterexample Construction (Ring)}
\begin{toolbox}
Under the ring interpretation, we constructs the minimal subring of $\mathbb{R}$ generated by $\sqrt{3}$:
\[
\mathbb{Z}[\sqrt{3}] = \{ a + b\sqrt{3} \mid a,b \in \mathbb{Z} \}.
\]

This structure:
\begin{itemize}
    \item Contains $\sqrt{3}$,
    \item Is closed under addition and multiplication,
    \item Is a \textbf{proper subset} of $\mathbb{R}$.
\end{itemize}

Thus, $\mathbb{R}$ cannot be the smallest number ring containing $\sqrt{3}$.
\end{toolbox}

\stepheader{Feedback: Cross-Interpretation Validation}
\begin{obsbox}
To ensure robustness, we further verifies the statement under the field interpretation:

\begin{itemize}
    \item The smallest field containing $\sqrt{3}$ is $\mathbb{Q}(\sqrt{3})$.
    \item This field is also strictly smaller than $\mathbb{R}$.
\end{itemize}

The counterexample remains valid across interpretations, confirming the correctness of the falsification.
\end{obsbox}

\stepheader{Action: Final Judgment}
\begin{toolbox}
By explicitly constructing minimal algebraic counterexamples and validating them under multiple definitions, we concludes:
\[
\boxed{\text{The statement is false.}}
\]

This trajectory demonstrates that \textcolor{blue}{our model conduct a full agentic loop involving \textbf{analysis, planning, feedback, and action}, rather than mere step-by-step computation}.
\end{toolbox}

\end{tcolorbox}

\subsection{Case of Deep Research}
\begin{tcolorbox}[colback=white, colframe=black, title=Analysis, fonttitle=\bfseries]
The agent locate and parse the official 2011 Bulgarian census PDF to extract precise data points, rather than relying on potentially inaccurate search snippets.
Most notably, the agent exhibited adaptive reasoning and error recovery. When a tool call failed due to a syntax error during the calculation phase, the agent did not loop or hallucinate; instead, it immediately switched strategies to perform the arithmetic internally. This seamless transition from tool-use to internal reasoning ensured the final answer was both accurate and delivered efficiently.
\end{tcolorbox}

\begin{tcolorbox}[colback=white, colframe=black, boxrule=0.8pt, breakable, rounded corners]
\textbf{Question:} Take the gender split from the 2011 Bulgarian census about those who have completed tertiary education. Subtract the smaller number from the larger number, then return the difference in thousands of women. So if there were 30.1 thousand more men, you'd give "30.1".

\stepheader{0}
\begin{thoughtbox}
\textbf{Think:}

...It seems like a mathematical calculation based on demographic data from the 2011 Bulgarian census about those who have completed tertiary education. However, the question doesn't provide specific numbers or data that I can directly calculate from. \textcolor{blue}{Let me search for information about the 2011 Bulgarian census and its data on tertiary education completion} to understand what numbers I might need.
\end{thoughtbox}

\begin{toolbox}
\textbf{Action:}
\begin{lstlisting}
search({"query": ["2011 Bulgarian census tertiary education gender split men women"]})
\end{lstlisting}
\end{toolbox}

\begin{obsbox}
\textbf{Search Results:}

[1] 2011 POPULATION CENSUS – Main Results: Women have completed tertiary education significantly more often than the men and at the census moment 791.8 thousand women (22.3\%) and ... Source: [URL 1]


[2] 2011 BULGARIA NATIONAL REPORT: Population structure by gender and age. As of 1 February 2011 the female population was 3 777 999 (51.3\%), while males amounted to 3 586 571 persons (48.7 ... Source: [URL 2]


[3] Census 2021: Every fourth person in Bulgaria is a ...: The statistics institute said that the gender gap in tertiary education had increased compared with the census 10 years earlier ... Source: [URL 3]


...

\end{obsbox}

\stepheader{1}
\begin{thoughtbox}
\textbf{Think:}

Let me visit the official Bulgarian National Statistical Institute website to get the specific data on tertiary education completion by gender from the 2011 census.
\end{thoughtbox}

\begin{toolbox}
\textbf{Action:}
\begin{lstlisting}
visit({"url": ["[URL 1]"], "goal": "Find specific data on tertiary education completion by gender from the 2011 Bulgarian census, particularly the numbers for men and women who have completed tertiary education"})
\end{lstlisting}
\end{toolbox}

\begin{obsbox}
\textbf{Visit Results:}

Webpage Summary: According to the 2011 Bulgarian census, the gender data regarding tertiary education completion is: The total number of people with a tertiary education degree is 1,348,700 (19.6\% of the total population). Of these, 791,800 are women (22.3\% of the female population) and 556,900 are men (16.7\% of the male population).
Relevant Webpage Link: [URL 1]


\end{obsbox}

\stepheader{2}
\begin{thoughtbox}
Great! I found the specific numbers from the 2011 Bulgarian census: Women with tertiary education: 791,800; Men with tertiary education: 556,900. Now I need to calculate the difference: Subtract the smaller number (556,900) from the larger number (791,800); Then return the difference in thousands of women. Let me calculate this:\\
\end{thoughtbox}

\begin{toolbox}
\textbf{Action:}
\begin{lstlisting}
search({"query": "791800 - 556900 difference in thousands women"})
\end{lstlisting}
\end{toolbox}

\begin{obsbox}
\textbf{Search Results:}
\textcolor{blue}{An error occurred while running the tool}. Please try again. Error: Invalid JSON input for tool search: 1 validation error for search\_args. Query input should be a valid list [type=list
\_type, input\_value=`791800 - 556900 difference in thousands women', input
\_type=str]. For further information visit [URL 4].


\end{obsbox}

\stepheader{3}
\begin{thoughtbox}
\textbf{Think:}\\
\textcolor{blue}{Let me try the calculation manually}:\\
- Women: 791,800\\
- Men: 556,900\\
- Difference: 791,800 - 556,900 = 234,900\\
- In thousands: 234,900 ÷ 1,000 = 234.9\\
So the difference in thousands of women is 234.9.\\
Let me verify this calculation and provide the final answer.\\
\textbf{Answer:}\\
According to the 2011 Bulgarian census data on tertiary education completion:\\
- Number of women with tertiary education: 791,800\\
- Number of men with tertiary education: 556,900\\
The difference is calculated as follows:\\
- Larger number (women): 791,800\\
- Smaller number (men): 556,900\\
- Difference: 791,800 - 556,900 = 234,900\\
- In thousands: 234,900 ÷ 1,000 = 234.9.\\
Therefore, there were \textcolor{blue}{234.9 thousand} more women who have completed tertiary education than men in Bulgaria according to the 2011 census.
\end{thoughtbox}

\end{tcolorbox}

\subsection{Case of Agentic Coding}
\begin{tcolorbox}[colback=white, colframe=black, title=Analysis,  fonttitle=\bfseries]
To demonstrate the model's capabilities in agentic coding, we select instance \texttt{sphinx-doc\_\_sphinx-7889} from SWE-bench Verified. This instance is derived from a real-world issue within the \texttt{Sphinx} open-source Python repository. This trajectory illustrates high-level Python debugging. The model identifies a rare edge case where a string concatenation assumes the input is always a string, whereas it can be a TypeVar object in generic classes. It shows the model's ability to navigate large documentation frameworks and fix issues at the intersection of static analysis and runtime metadata.
\end{tcolorbox}

\begin{tcolorbox}[colback=white, colframe=black, boxrule=0.8pt, breakable, rounded corners]
\textbf{Question:} Autodoc extension's mock file throws TypeError for generic-typed classes.
**Describe the bug**
When building the docs for a generically-typed class, a TypeError is thrown as Autodoc's `mock.\_make\_subclass` attempts to concatenate a `str` to a `TypeVar`. See the attached log: [sphinx-err-325ndteh.log](https://github.com/sphinx-doc/sphinx/files/4842672/sphinx-err-325ndteh.log)

**To Reproduce**
```
\$ git https://github.com/perrygoy/screenpy.git
\$ cd screenpy/docs
\$ python -m venv env
\$ source env/bin/activate
\$ pip install sphinx pyhamcrest selenium typing\_extensions
\$ make html
```
Observe the docs command fails with a TypeError.

**Expected behavior**
Docs can still be built when generics are involved.

**Your project**
https://github.com/perrygoy/screenpy.git

**Environment info**
- OS: Mac 10.15.5 (19F101)
- Python version: 3.7.7
- Sphinx version: 3.1.1
- Sphinx extensions:  sphinx.ext.autodoc, sphinx.ext.intersphinx, sphinx.ext.coverage, sphinx.ext.ifconfig, sphinx.ext.napoleon

**Additional context**
This might just be me not knowing how to make Sphinx-friendly generic typing, if that's the case please let me know!

\stepheader{0}
\begin{thoughtbox}
I'll help you implement the necessary changes to fix the issue with the autodoc extension's mock file throwing a TypeError for generic-typed classes. Let's follow the steps systematically.\\
\\
\#\# Step 1: Explore the codebase to locate and understand the code relevant to the issue\\
\\
First, \textcolor{blue}{let's explore the repository structure to understand what we're working with:}\end{thoughtbox}

\begin{toolbox}
\textbf{Action:}
\begin{lstlisting}
<function=execute_bash>
  <parameter=command>find /testbed -type f -name "*.py" | grep -E "(autodoc|mock)" | head -20</parameter>
</function>
\end{lstlisting}
\end{toolbox}

\begin{obsbox}
\textbf{Observation:}
\begin{lstlisting}
Exit code: 0
Execution output of [execute_bash]:
[STDOUT]

/testbed/sphinx/ext/autodoc/mock.py
/testbed/sphinx/ext/autodoc/type_comment.py
/testbed/sphinx/ext/autodoc/typehints.py
/testbed/sphinx/ext/autodoc/importer.py
/testbed/sphinx/ext/autodoc/__init__.py
/testbed/sphinx/ext/autodoc/directive.py
/testbed/tests/test_ext_autodoc_autofunction.py
/testbed/tests/test_ext_autodoc_configs.py
/testbed/tests/test_ext_autodoc_private_members.py
/testbed/tests/test_ext_autodoc_mock.py
/testbed/tests/test_ext_autodoc_events.py
/testbed/tests/test_ext_autodoc.py
/testbed/tests/roots/test-warnings/autodoc_fodder.py
/testbed/tests/roots/test-root/autodoc_target.py
/testbed/tests/roots/test-ext-autosummary-mock_imports/conf.py
/testbed/tests/roots/test-ext-autosummary-mock_imports/foo.py
/testbed/tests/roots/test-ext-autodoc/bug2437/autodoc_dummy_foo.py
/testbed/tests/roots/test-ext-autodoc/bug2437/__init__.py
/testbed/tests/roots/test-ext-autodoc/autodoc_dummy_bar.py
/testbed/tests/roots/test-ext-autodoc/conf.py 

[STDERR]

\end{lstlisting}
\end{obsbox}

\stepheader{1}
\begin{thoughtbox}
Let's first examine the mock.py file mentioned in the issue, as it seems to be the source of the problem:\end{thoughtbox}

\begin{toolbox}
\textbf{Action:}
\begin{lstlisting}
<function=str_replace_editor>
  <parameter=command>view</parameter>
  <parameter=path>/testbed/sphinx/ext/autodoc/mock.py</parameter>
</function>
\end{lstlisting}
\end{toolbox}

\begin{obsbox}
\textbf{Observation:}
\begin{lstlisting}
Execution output of [str_replace_editor]:
Here's the result of running `cat -n` on the file: /testbed/sphinx/ext/autodoc/mock.py:
     1 """
     2     sphinx.ext.autodoc.mock
     3     ~~~~~~~~~~~~~~~~~~~~~~~
     4 
     5     mock for autodoc
     6 
     7     :copyright: Copyright 2007-2020 by the Sphinx team, see AUTHORS.
     8     :license: BSD, see LICENSE for details.
     9 """
    10 
    11 import contextlib
    12 import os
    13 import sys
    14 from importlib.abc import Loader, MetaPathFinder
    15 from importlib.machinery import ModuleSpec
    16 from types import FunctionType, MethodType, ModuleType
    17 from typing import Any, Generator, Iterator, List, Sequence, Tuple, Union
    18 
    19 from sphinx.util import l

... [Omitted for brevity] ...

 invalidate_caches(self) -> None:
   130         """Invalidate mocked modules on sys.modules."""
   131         for modname in self.mocked_modules:
   132             sys.modules.pop(modname, None)
   133 
   134 
   135 @contextlib.contextmanager
   136 def mock(modnames: List[str]) -> Generator[None, None, None]:
   137     """Insert mock modules during context::
   138 
   139         with mock(['target.module.name']):
   140             # mock modules are enabled here
   141             ...
   142     """
   143     try:
   144         finder = MockFinder(modnames)
   145         sys.meta_path.insert(0, finder)
   146         yield
   147     finally:
   148         sys.meta_path.remove(finder)
   149         finder.invalidate_caches()

\end{lstlisting}
\end{obsbox}

\begin{center}\textit{[Agent performed 1 intermediate operations]}\end{center}

\stepheader{3}
\begin{thoughtbox}
Based on my analysis of the code, I can see that the issue occurs in the `\_make\_subclass` function in `/testbed/sphinx/ext/autodoc/mock.py`. \textcolor{blue}{The problem is that when dealing with generic-typed classes, the function tries to concatenate a string to a `TypeVar`, which causes a TypeError.}\\
\\
\textcolor{blue}{Let's first create a reproduction script to confirm the issue:}\end{thoughtbox}

\begin{toolbox}
\textbf{Action:}
\begin{lstlisting}
<function=str_replace_editor>
  <parameter=command>create</parameter>
  <parameter=path>/testbed/reproduce_issue.py</parameter>
  <parameter=file_text>#!/usr/bin/env python3
"""
Script to reproduce the TypeError when building docs for generic-typed classes.
"""

import sys
from typing import TypeVar, Generic, List
from sphinx.ext.autodoc.mock import _make_subclass

def test_generic_class():
    """Test creating a mock subclass for a generic class."""
    # Define a generic class with TypeVar
    T = TypeVar('T')
    
    class MyGeneric(Generic[T]):
        def __init__(self, value: T) -> None:
            self.value = value
        
        def get_value(self) -> T:
            return self.value
    
    try:
        # This should trigger the TypeError
        mock_class = _make_subclass('MyGenericMock', 'test_module', MyGeneric)
        print("SUCCESS: No TypeError occurred")
        return True
    except TypeError as e:
        print(f"ERROR: TypeError occurred: {e}")
        return False

if __name__ == "__main__":
    success = test_generic_class()
    sys.exit(0 if success else 1)</parameter>
</function>
\end{lstlisting}
\end{toolbox}

\begin{obsbox}
\textbf{Observation:}
\begin{lstlisting}
Execution output of [str_replace_editor]:
File created at /testbed/reproduce_issue.py. Here's the result of running `cat -n` on /testbed/reproduce_issue.py:
     1	#!/usr/bin/env python3
     2	"""
     3	Script to reproduce the TypeError when building docs for generic-typed classes.
     4	"""
     5	
     6	import sys
     7	from typing import TypeVar, Generic, List
     8	from sphinx.ext.autodoc.mock import _make_subclass
     
... [Omitted for brevity] ...

    31	if __name__ == "__main__":
    32	    success = test_generic_class()
    33	    sys.exit(0 if success else 1)
Review the file and make sure that it is as expected. Edit the file if necessary.

\end{lstlisting}
\end{obsbox}

\begin{center}\textit{[Agent performed 7 intermediate operations]}\end{center}

\stepheader{11}
\begin{thoughtbox}
Great! Now we've successfully reproduced the issue. The error occurs when we try to concatenate a string to a `TypeVar`. The second test case (`test\_typevar\_in\_name`) now triggers the error as expected.\\
\\
\#\# Root Cause Analysis\\
\\
The issue is in the `\_make\_subclass` function in `/testbed/sphinx/ext/autodoc/mock.py` at line 73:\\
\\
```python\\
attrs = \{'\_\_module\_\_': module, '\_\_display\_name\_\_': module + '.' + name\}\\
```\\
\\
When `name` is a `TypeVar` (which is a type, not a string), the expression `module + '.' + name` fails because you can't concatenate a string to a `TypeVar`.\\
\\
\#\# Potential Solutions\\
\\
1. **Check if name is a string before concatenation**: Add a type check to ensure `name` is a string before concatenating.\\
\\
2. **Convert name to string if it's not**: If `name` is not a string, convert it to a string representation.\\
\\
3. **Use a safer concatenation method**: Use `str(name)` to ensure we always have a string.\\
\\
\textcolor{blue}{Let's evaluate these options:\\
\\
1. **Check if name is a string before concatenation**: This is the most direct approach and would be safe.\\
\\
2. **Convert name to string if it's not**: This would work but might be less efficient.\\
\\
3. **Use a safer concatenation method**: This is essentially the same as option 2.\\
\\
The most elegant and safe solution is to check if `name` is a string before concatenation, or better yet, use `str(name)` to ensure we always have a string.}\\
\\
\#\# Implementation\\
\\
Let's implement the fix by modifying the `\_make\_subclass` function:\end{thoughtbox}

\begin{toolbox}
\textbf{Action:}
\begin{lstlisting}
<function=str_replace_editor>
  <parameter=command>str_replace</parameter>
  <parameter=path>/testbed/sphinx/ext/autodoc/mock.py</parameter>
  <parameter=old_str>def _make_subclass(name: str, module: str, superclass: Any = _MockObject,
                   attributes: Any = None) -> Any:
    attrs = {'__module__': module, '__display_name__': module + '.' + name}
    attrs.update(attributes or {})

    return type(name, (superclass,), attrs)</parameter>
  <parameter=new_str>def _make_subclass(name: str, module: str, superclass: Any = _MockObject,
                   attributes: Any = None) -> Any:
    # Ensure name is a string for concatenation
    display_name = module + '.' + str(name) if not isinstance(name, str) else module + '.' + name
    attrs = {'__module__': module, '__display_name__': display_name}
    attrs.update(attributes or {})

    return type(name, (superclass,), attrs)</parameter>
</function>
\end{lstlisting}
\end{toolbox}

\begin{obsbox}
\textbf{Observation:}
\begin{lstlisting}
Execution output of [str_replace_editor]:
The file /testbed/sphinx/ext/autodoc/mock.py has been edited. Here's the result of running `cat -n` on a snippet of /testbed/sphinx/ext/autodoc/mock.py:
    67	    def __repr__(self) -> str:
    68	        return self.__display_name__
    69	
    70	
    71	def _make_subclass(name: str, module: str, superclass: Any = _MockObject,
    72	                   attributes: Any = None) -> Any:
    73	    # Ensure name is a string for concatenation
    74	    display_name = module + '.' + str(name) if not isinstance(name, str) else module + '.' + name
    75	    attrs = {'__module__': module, '__display_name__': display_name}
    76	    attrs.update(attributes or {})
    77	
    78	    return type(name, (superclass,), attrs)
    79	
    80	
    81	class _MockModule(ModuleType):
    82	    """Used by autodoc_mock_imports."""
Review the changes and make sure they are as expected. Edit the file again if necessary.

\end{lstlisting}
\end{obsbox}

\begin{center}\textit{[Agent performed 15 intermediate operations]}\end{center}

\stepheader{27}
\begin{thoughtbox}
\#\# Summary\\
\\
I have successfully implemented a fix for the TypeError that occurred when building docs for generically-typed classes using the Autodoc extension's mock file.\\
\\
\#\#\# Root Cause\\
The issue was in the `\_make\_subclass` function in `/testbed/sphinx/ext/autodoc/mock.py`. When creating mock subclasses for generic-typed classes, the function would attempt to concatenate a string to a `TypeVar` (which is a type, not a string), causing a `TypeError`.\\
\\
\#\#\# Solution\\
I modified the `\_make\_subclass` function to safely handle non-string values for the `name` parameter by converting them to strings before using them in the `type()` constructor:\\
\\
```python\\
def \_make\_subclass(name: str, module: str, superclass: Any = \_MockObject,\\
                 attributes: Any = None) -> Any:\\
    \# Ensure name is a string for concatenation and type() constructor\\
    name\_str = str(name) if not isinstance(name, str) else name\\
    display\_name = module + '.' + name\_str\\
    attrs = \{'\_\_module\_\_': module, '\_\_display\_name\_\_': display\_name\}\\
    attrs.update(attributes or \{\})\\
\\
    return type(name\_str, (superclass,), attrs)\\
```\\
\\
\textcolor{blue}{\#\#\# Changes Made\\
1. Modified `/testbed/sphinx/ext/autodoc/mock.py` to safely convert non-string `name` values to strings before using them\\
2. Verified the fix works with the original reproduction case\\
3. Tested various edge cases including generic classes, TypeVars, List, Dict, Optional, and nested generics\\
4. Confirmed all existing tests in `test\_ext\_autodoc\_mock.py` still pass\\
\\
\#\#\# Verification\\
- The original reproduction script now works without errors\\
- All edge case tests pass\\
- Existing unit tests for the mock extension continue to pass\\
- The fix is minimal and targeted, addressing only the specific issue without breaking existing functionality}\\
\\
The fix ensures that Autodoc can build documentation for generically-typed classes without throwing a TypeError, resolving the issue described in the GitHub issue.\end{thoughtbox}

\begin{toolbox}
\textbf{Action:}
\begin{lstlisting}
<function=submit>
</function>
\end{lstlisting}
\end{toolbox}

\begin{obsbox}
\textbf{Observation:}
\begin{lstlisting}
<<< Finished >>>
\end{lstlisting}
\end{obsbox}

\end{tcolorbox}

\end{document}

%% file: 0_titlepage.tex
\begin{figure}[h]
    \centering
    \includegraphics[width=0.69\linewidth]{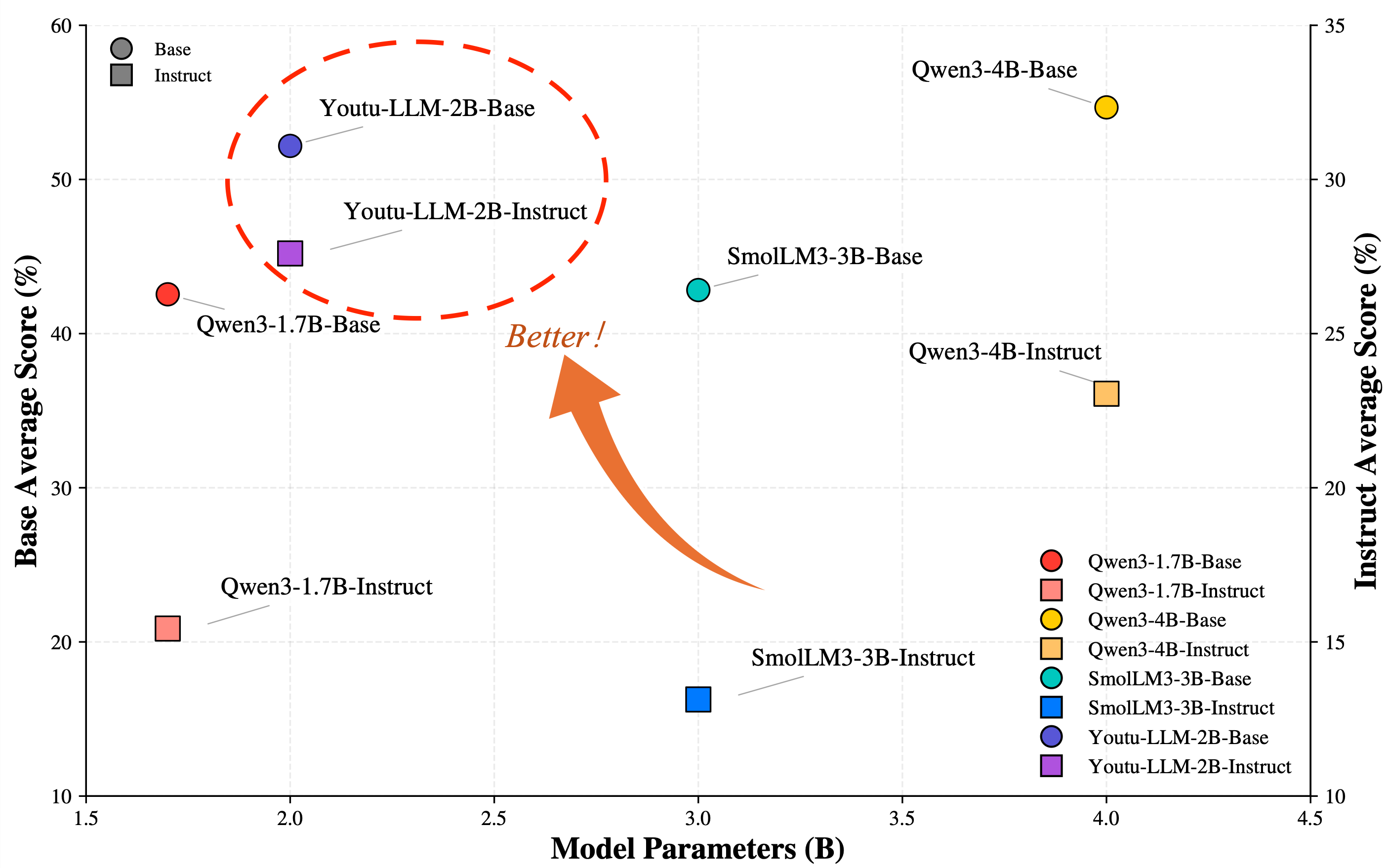}
    \caption{Parameter–performance scaling of base and instruct models on agentic benchmarks.
The trend line represents the desired agent performance with the smallest possible number of parameters, among which Youtu-LLM stands out as a lightweight yet strong performer.}
    \label{fig:teaser_figure}
\end{figure}

\label{chp:0_titlepage}

%% file: 1_introduction.tex
\section{Introduction}
\label{chp:1_intro}

\begin{figure}
    \centering
    \includegraphics[width=0.90\linewidth]{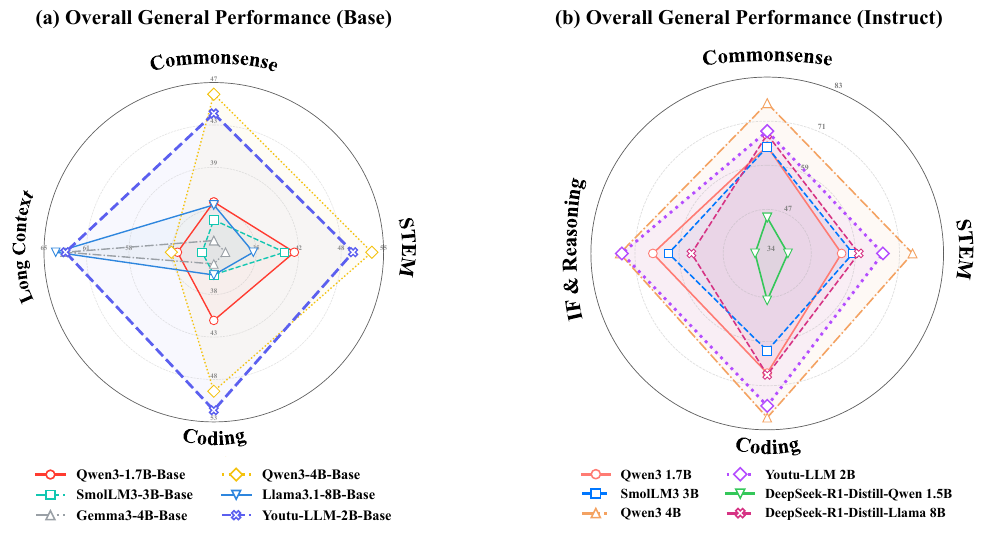}
    \caption{Multi-field general capability comparison of similarly sized models. Youtu-LLM shows a balanced and competitive profile, highlighting its general-purpose performance potential under limited parameter budgets.}
    \label{fig:general-radar}
\end{figure}

Large Language Models (LLMs) have made remarkable progress in recent years, steadily advancing toward Artificial General Intelligence (AGI)~\citep{hendrycks2025definition}.
By leveraging massive multi-domain corpora, growing parameter scales, and continuously evolving training paradigms, modern LLMs have demonstrated strong capabilities in reasoning, problem solving, and decision making~\citep{guo2025deepseek,qwen3technicalreport}.
Notably, recent reasoning-oriented models have achieved phenomenal performance on challenging agentic benchmarks, highlighting the effectiveness of large-scale training in eliciting complex cognitive behaviors~\citep{plaat2025agentic}.

Despite these successes, current progress is tightly coupled with parameter scaling.
State-of-the-art models typically rely on tens or hundreds of billions of parameters, incurring substantial computational, financial, and environmental costs during both training and deployment~\citep{belcak2025small}.
These constraints significantly limit accessibility and real-world research and applicability, especially in latency-sensitive or resource-constrained settings.
As a result, there is renewed interest in developing lightweight large language models that retain strong general-purpose and reasoning capabilities while remaining practical for deployment. More well-known open-source model series are beginning to release models with fewer than 7B, and even fewer than 2B parameter models~\citep{hu2024minicpm,bakouch2025smollm3}. Existing approaches for improving small models largely rely on distillation, instruction tuning, or architectural simplification~\citep{sharma2025small}.
Although effective to some extent, these methods primarily align output behavior rather than systematically cultivating underlying cognitive capabilities.
Consequently, lightweight models often lack robustness, generalization, and planning competence.

Urgently, with the rapid emergence of complex tasks such as deep research, coding, and tool-augmented workflows, the above mentioned limitations of lightweight LLMs become even more pronounced in real-world agentic scenarios.
The general consensus of the community indicates that effective agents require not only strong language understanding but also intrinsic capabilities for planning, tool execution, state perception, and feedback-driven reflection~\citep{luo2025large}.
Therefore, recent work has begun exploring these native agentic capabilities in language models, shifting away from purely external agent frameworks toward internalized reasoning and interaction behaviors~\citep{zeng2025glm}.
In particular, trajectory-based training and continual pre-training on structured interaction data have shown promise in enhancing planning, reasoning, and tool-use abilities~\citep{team2025tongyi}.
Nevertheless, existing studies leave a critical open question unanswered:
\emph{Can lightweight LLMs acquire strong agentic capabilities through pre-training, rather than post-augmentation, such as post-training or agentic frameworks?}

In this work, we claim that strong agentic performance in lightweight LLMs is achievable when agent-oriented signals are injected early and systematically through an agentic pre-training process.
Specifically, we introduce \textbf{Youtu-LLM}, a 2B-sized lightweight open-source model designed to balance compactness with robust general and agentic performance.
Our approach integrates innovations in tokenizer design, data allocation, and multi-stage learning strategies under a STEM- and agent-centric training principle.
Specifically, we propose a series of scalable frameworks for constructing high-quality agentic trajectory data for pre-training. These frameworks cover a broad range of capabilities, including reasoning, reflection, and planning, across diverse domains such as mathematics, coding, deep research, and general tool use.
Through the proposed data pipeline, we obtain over 200B tokens of high-quality agentic trajectory data, providing fuel for our agentic pre-training.

Youtu-LLM significantly outperforms existing state-of-the-art models of similar scale across both general-purpose (Figure\,\ref{fig:general-radar}) and agentic benchmarks (Figure\,\ref{fig:teaser_figure}), and in several settings, rivals substantially larger models.
Beyond performance gains, our analyses provide the first systematic evidence that agentic pre-training can unlock agent potential in lightweight LLMs, revealing phenomena such as scalable growth of agent capabilities.
We summarize the main contributions, highlights, and insights of Youtu-LLM as follows:

\begin{itemize}
\vspace{-1.0em}
    \item \textbf{Lightweight Agentic LLM.} We introduce Youtu-LLM, a lightweight open-source language model that significantly outperforms state-of-the-art models of similar or even larger size in agent benchmarks.
    
    \item \textbf{Native Agentic Capability Induction.} We propose a principled training paradigm that enhances native agentic capabilities through innovations in tokenizer design, data allocation, and multi-stage learning, guided by an agent-centric philosophy.
    
    \item \textbf{Scalable Agentic Trajectory Construction.} We present a series of scalable frameworks for constructing high-quality agentic trajectory data, spanning reasoning, reflection and planning abilities, across multiple domains such as mathematics, coding, deep research and general tool use.
    
    \item \textbf{Empirical Insights.} We provide the first systematic analysis of agentic pre-training in lightweight LLMs, revealing the scalable growth of agentic abilities.

\end{itemize}
\vspace{-1.0em}

%% file: 2_pretrain_data.tex
\section{Pre-Training Data}
\label{chp:2_data}

The scale and quality of the data determine the potential of LLMs. In this section, we introduce the data recipe of Youtu-LLM's pre-training. We report on two aspects: conventional general pre-training data and intricate trajectory pre-training data. The former accounts for over 95\% of the total pre-training corpus, so we focus on controlling the overall quality and composition of this part; while the latter focuses on collecting complete, verifiable and high-quality agent execution trajectories, covering five categories: Agentic-CoT, Math, Code, Deep Research, and Tool-use.

\subsection{General Pre-Training Data}

We firstly collected over 10T raw tokens from various sources, with English as the primary language and Chinese as the secondary focus. After deduplication, filtering and decontamination, we retained 8.7T raw tokens. The high-quality Chinese and English web pages and encyclopedic knowledge totaled 6.1T tokens, accounting for over 70\% of the raw corpus. As aforementioned, STEM and coding capabilities were the key focus of Youtu-LLM's pre-training. Therefore, in the rest data, we compiled 700B tokens of Chinese and English STEM corpora, covering key disciplines such as mathematics, physics, chemistry, biology, and medicine; as for coding, we collected up to 1,400B tokens of source data. We also synthesized an additional 500B tokens of STEM and code corpora, including but not limited to explanations of STEM knowledge points, notebook documents, PDF summaries, paper interpretations, and code explanations in various forms. Finally, based on quality, we up-sampled certain high-quality STEM and code data. This expanded the 8.7T raw data into a 10.64T-token pool for common pre-training data.

Considering the risk of uneven quality and domain coverage bias in large-scale pre-training data, we adopted a solution involving multi-dimensional data classification and quality scoring model for filtering. We first developed a set of classification principles, including 10 quality assessment criteria and 11 domain classification criteria (comprising 46 sub-domains), covering core scenarios such as programming, mathematics, healthcare, and finance. Using a semi-automated process, we employed Deepseek-R1~\citep{deepseekai2025deepseekv3technicalreport} for initial label annotation. During the annotation process, samples were randomly selected for manual review. Only when the consistency between the labels generated by the Deepseek-R1 and the manual annotations reached over 95\%, the current batch of annotated data would be officially included in the training set. Through multiple iterations and optimizations, we obtained approximately 600K instances of English and Chinese data with evenly distributed domain and quality labels. Finally, we trained a fast classification and scoring quality model based on Qwen3-1.7B~\citep{qwen3technicalreport}, achieving over 95\% accuracy of both domain classification and quality scoring, and showing high consistency with manual review. A small-scale experiments proved that 80B high-quality data selected by the filtering model obtained even better performance than the original 100B raw data with only 50\% training steps.

Based on this classification and scoring model, we performed preliminary classification and quality assessment of the original corpus, retaining only those parts with an average evaluation score above 8.5 across all 10 criteria. Based on this, we binned the remaining data by length interval, then performed heuristic rule-based filtering, and MinHash-LSH-based~\citep{Broder1997OnTR,HarPeled2012ApproximateNN} deduplication within each bin. We included several heuristic rules such as 13-gram duplicate detection, toxicity scoring, and code keyword filtering. In addition, we introduced the Aho–Corasick algorithm~\citep{aho1975efficient} to benchmark and decontaminate the STEM and code corpus, minimizing the risk of test dataset leakage.

\subsection{Agentic Trajectory Data}
\label{sec:agentic_data}
 This section details our approach to constructing diverse agentic trajectory data. We constructed 200B tokens of high-quality trajectory data, including 25B Agentic-CoT trajectories, 20B mathematical trajectories, 70B code execution trajectories, 60B Deep Research trajectories, and 25B other trajectories (\eg tool using, function calling, and planning).

\subsubsection{Agentic-CoT Trajectory}

\paragraph{Motivation and Goal}
The reasoning data utilized in LLMs has progressed through multiple stages, including chain-of-thought (CoT)~\citep{wei2022chain} and slow thinking enhanced with reinforcement learning techniques~\citep {shao2024deepseekmath}. Currently, thinking models that encapsulate reasoning steps within <think> tags are prevalent. Although extended chains of thought significantly improve the accuracy of final answers, such reasoning traces are often plagued by redundancy and repetitive expressions. Directly learning from these raw reasoning processes may compromise both logical coherence and natural language fluency. Nevertheless, the information contained within these traces remains valuable. We argue that by refining such content, it is possible to remove inefficient expressions while preserving the essential logical structure.

\paragraph{Data Construction Strategy}
Inspired by agentic workflows—which typically involve planning to break down problems, step-by-step actions, execution attempts, reflection, and final summarization—we propose a structured thinking paradigm called Agentic-CoT and construct a corresponding dataset. This approach represents a refined method for organizing reasoning processes and supports the advancement of agentic thinking paradigms.

\begin{figure}[t]
    \centering
    \includegraphics[width=1\textwidth]{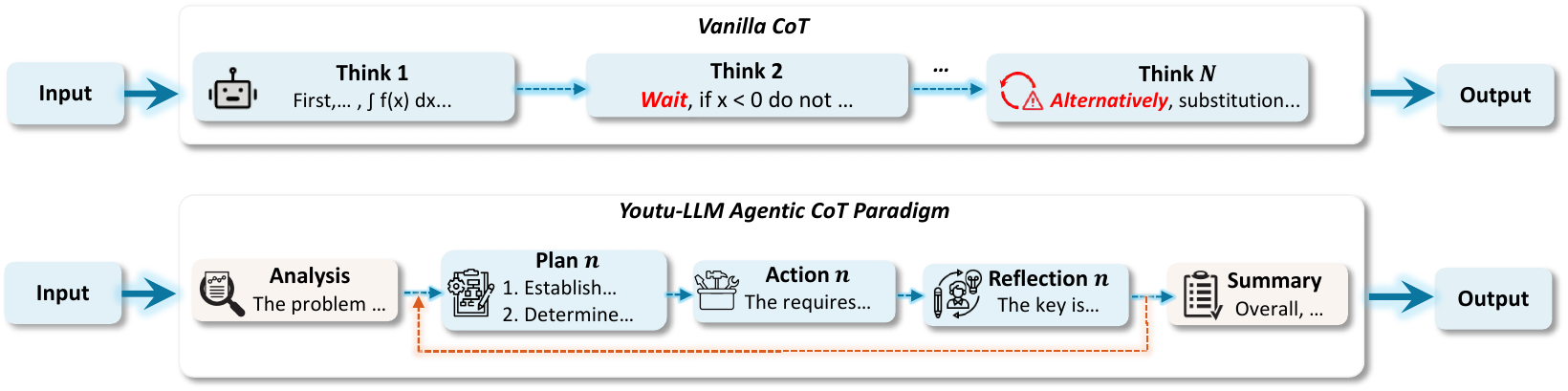}
    \caption{Above: Vanilla long CoT enables thorough response preparation, while it can also tend to overthinking and unnecessary repetition. Below: Our agentic thinking paradigm implements a defined reasoning architecture that guides models through sequential steps of analysis, plan, action, reflection and summary. This disciplined process fosters the development of agentic capabilities.}
    \label{fig:agentic_thinking_paradigm_1}
\end{figure}

In detail, raw reasoning data are transformed into an Agentic-CoT trajectory composed of five distinct phases (Figure~\ref{fig:agentic_thinking_paradigm_1}). The original, often linear or monolithic, CoT is decomposed and reorganized into the following sequential segments, each encapsulated within its own XML-style tag, shown in Table \ref{tab:Reflection Thought Data Segments}.

\begin{table}[t]
\centering
\small

\begin{tabular}{p{0.15\linewidth} p{0.8\linewidth}}
\toprule
\textbf{Components} & \textbf{Description} \\
\midrule
\textbf{Analysis} & This segment contains the initial breakdown and examination of the given problem. It includes identifying key components, constraints, and the core question to be answered. \\
\midrule
\textbf{Plan} & This section outlines the strategic steps devised to solve the problem based on the preceding analysis. It serves as a blueprint for the subsequent actions. \\
\midrule
\textbf{Action} & This part documents the concrete implementation of the planned steps, leading to an initial result or output. \\
\midrule
\textbf{Reflection} & This phase captures an evaluation of the execution step and its outcome. It involves checking for errors, considering alternative approaches, or assessing the result's validity and completeness. \\
\midrule
\textbf{Summary} & The final segment provides a concise synthesis of the entire reasoning trajectory, distilling the key findings, the final answer, or the ultimate conclusion. \\
\bottomrule
\end{tabular}
\caption{Agentic-CoT Data Components and Description.}
\label{tab:Reflection Thought Data Segments}

\end{table}

Our data construction methodology employs a stratified rewriting approach, utilizing LLMs to systematically generate and refine the dataset. The process unfolds in multiple tiers. (1) \textbf{Reasoning and Generation:} Initially, a foundational reasoning model processes input queries to perform analysis and produce initial responses. This stage focuses on core logical derivation and answer formulation. (2) \textbf{Curation and Extraction:} The generated outputs undergo a rigorous filtering and verification process to ensure the correctness. Subsequently, the targeted segments are extracted from the structured triplets—comprising the original query, the model's internal reasoning chain, and the final response. (3) \textbf{Synthesis and Assembly:} The refined segments are then intelligently synthesized and concatenated. This assembly forms our proprietary Agentic-CoT Data, a specialized corpus designed to capture explicit reasoning pathways.

\paragraph{Data Statistical Information}

Through this multi-layered pipeline, we have constructed a comprehensive Agentic-CoT dataset spanning 25B tokens. Its scope encompasses diverse domains such as industry, mathematics, STEM, and Coding. 
Experimental results demonstrate that Agentic-CoT Data significantly enhance the agentic thought of the model, leading to improved performance on agentic benchmarks. The corresponding results are provided in Appendix \ref{app:agent_reasoning}.

\begin{table}[t]
\centering
\small

\setlength{\tabcolsep}{3.5pt}
\renewcommand{\arraystretch}{1.25}
\begin{tabular}{
p{3.1cm}   
p{3.4cm}   
p{6.8cm}  
p{1.7cm}   
}
\toprule
\textbf{Ability Level} & \textbf{Atomic Ability} & \textbf{Description} & \textbf{Module} \\
\midrule

& Symbol Recognition
& Identifying mathematical symbols and structural patterns from textual or visual inputs
& Action \\

\multirow{2}{*}{\makecell[c]{\textbf{Basic Knowledge}\\\textbf{and Computation}}} & Concept Understanding
& Understanding and distinguishing mathematical concepts

& Action \\

& Computation Execution
& Performing arithmetic, algebraic, calculus, and linear algebra operations
& Action \\

\midrule

& Spatial Perception
& Reasoning over geometric objects, spatial relations, and quantitative measurements
& Action \\

\multirow{5}{*}{\makecell[c]{\textbf{Complex Reasoning}\\\textbf{and Application}}}& Formal Mathematical Language
& Translating natural language mathematical statements into formal representations
& Action \\

& Deduction and Induction Reasoning
& Deriving conclusions through deductive or inductive reasoning given known premises
& Action \\

& Counter-proof and Construction
& Solving problems via goal-driven reasoning strategies
& Action \\

& Modeling Transformation
& Converting real-world or textual scenarios into mathematical models with variables, constraints, and objectives
& Action \\

\midrule

& Theorem Application
& Identifying applicable theorems or lemmas, verifying conditions, and applying them correctly
& Feedback \\

\multirow{2}{*}{\makecell[c]{\textbf{Mathematical}\\\textbf{Meta-cognition}}} & Self-Reflection
& Diagnosing errors in reasoning trajectories and proposing corrective strategies
& Feedback \\

& New Knowledge Acquisition
& Learning novel definitions or theorems and transferring them to unseen mathematical problems
& Feedback \\

\bottomrule
\end{tabular}
\caption{Atomic ability taxonomy for mathematical reasoning and its correspondence to agent modules.}
\label{tab:atomic_ability_agent_mapping}

\end{table}

\subsubsection{Math Trajectory} 
\paragraph{Motivation and Goal}
Mathematical reasoning tasks have long served as a cornerstone for evaluating model performance in complex reasoning scenarios, due to their objectively verifiable ground-truth answers and coverage of a broad difficulty spectrum. They are also widely regarded as the critical benchmarks for assessing the cognitive capabilities of Large Language Models or agents~\citep{shao2025deepseekmath}. 
Although mathematical reasoning trajectories lack explicit environmental feedback signals commonly found in code execution or search-based trajectories, they exhibit \textbf{highly generalizable and abstract agentic behavior patterns}~\citep{liu2025agenticmath}. These patterns naturally capture essential agent behaviors such as planning, execution, and reflection, making mathematical trajectories particularly well-suited for pre-training models with robust cognitive strategies~\citep{gao2025long,zhao2025achieving}.
Motivated by this observation, we incorporate mathematical reasoning trajectories into our trajectory pre-training pipeline. We aim to construct trajectories that model diverse agentic behaviors and fundamental reasoning abilities in a \textbf{clear, systematic, and comprehensive manner}, enabling large language models to acquire \textbf{interpretable, stable, and high-quality mathematical reasoning behaviors} during pre-training.

\paragraph{Trajectory Construction Strategy}
To meet the demand for large-scale, high-quality mathematical trajectories, we require an agent framework that \textbf{organically integrates agentic behaviors with mathematical reasoning abilities}. With the rapid progress of LLM-based agents in complex tasks, various mathematical agents have been proposed. Some approaches enhance problem-solving accuracy through tool integration \citep{ToRA2024GouSGSYHDC}, or focus on specific mathematical tasks, such as mathematical modeling \citep{MM-Agent/corr/abs-2505-14148} or formal theorem proving \citep{REAL-Prover/corr/abs-2505-20613}. 
However, most existing mathematical agents are adapted from general-purpose agent frameworks and fail to tightly couple planning, execution, and reflection with the \textbf{fundamental abilities specific to mathematics}, lacking systematic analysis of the various atomic capabilities in mathematical reasoning. Although some works are trying to establish structured mathematical skills for mathematical reasoning~\citep{GAUSS:journals/corr/abs-2509-18122}, they have not yet integrated these systematic mathematical capability units with the agent framework. As a result, the model only learns limited agent mathematical reasoning behavior patterns from this type of trajectory data, making it difficult to effectively improve the comprehensive understanding of reasoning capabilities.

\begin{figure}[t]
    \small
    \centering
    \includegraphics[width=0.8\linewidth]{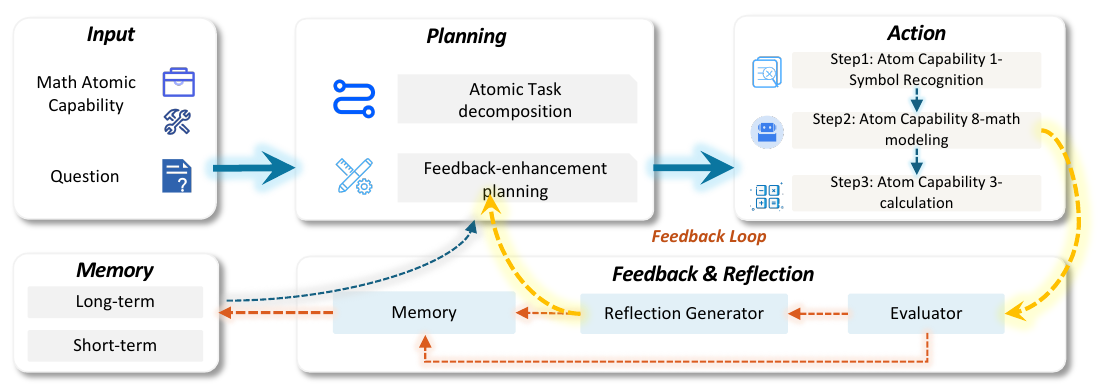}
    \caption{The illustration of the designed math agent framework for math trajectory construction.}
    \label{fig:math_agent}
\end{figure}

The Atomic Thinking \citep{kuang2025atomic} systematically decouples the atomic abilities involved in mathematical reasoning and analyzes their interactions, which aligns closely with the modular interactions in agent architectures. Inspired by this insight, we design a mathematical agent framework based on atomic ability to synthesize high-quality trajectories that comprehensively cover diverse agent behaviors and mathematical skills. Specifically, we construct a three-level atomic ability hierarchy that decomposes mathematical reasoning into $11$ independent and irreducible atomic abilities. The first level captures basic knowledge understanding and computation, the second level models complex reasoning and application abilities, and the third level represents higher-level mathematical metacognition. These atomic abilities can be naturally mapped to different agent modules and seamlessly integrated into planning, action, and feedback processes. The detailed ability taxonomy is summarized in Table~\ref{tab:atomic_ability_agent_mapping}.

We adopt a classical \textbf{planning--action--reflection} loop in our agent design, illustrated as Figure\,\ref{fig:math_agent}. Given a mathematical problem, the planning module first analyzes the problem structure and identifies the required atomic abilities, further decomposing them into executable sub-tasks with specified order, inputs, and expected outputs. The agent then enters the action stage, executing each atomic task step by step. After each execution, the feedback module evaluates the problem-solving state, returning feedback signals to the planning module for dynamic adjustment. Upon completion, we assess both answer correctness and trajectory quality, resulting in a large corpus of high-quality correct mathematical trajectories for pre-training.

\paragraph{Data Statistical Information}
For trajectory synthesis, we apply problems used during the annealing stage as queries, primarily consisting of high-difficulty computation, application, and reasoning tasks, which alighs with the K12 and mathematics competition level.  
Based on these problem sets and our proposed mathematical agent framework, we synthesize and filter a total of $1{,}380{,}327$ trajectories, comprising approximately $20$B tokens. 
During pre-training, trajectories with lengths up to $32$k tokens and those between $32$k and $128$k tokens are used separately.

\subsubsection{Code Trajectory} 
\paragraph{Data Statistical Information}
The continual pre-training dataset for code scenario is composed of two primary components: 
(1) \textbf{key atomic capabilities} required for real-world code-related problem solving; 
(2) \textbf{end-to-end agent trajectories}, which encompass issue-resolving tasks in software engineering scenarios and other real-world tasks involving code tools. We construct 70B tokens of code trajectory data.

\paragraph{Atomic Code Capabilities} 
The Atomic Code Capabilities dataset covers the following core atomic capabilities essential for autonomous code agents:
\vspace{-1.0em}
\begin{itemize}
    \item \textbf{Proactive Exploration and Localization}: Autonomously exploring the repository to locate specific variables, functions, or files, guided by ambiguous diagnostic logs or feature descriptions.
    \item \textbf{Context-Aware Generation and Completion}: Completing critical code segments given extensive repository-level context.
    \item \textbf{Patch Generation and Editing}: Generating precise modifications or patches based on localized faulty files and specific issue descriptions.
    \item \textbf{Testing and Verification}: Evaluating the correctness of action trajectories or localizing failure points within a sequence of steps.
    \item \textbf{Environment Comprehension}: The ability to interpret or predict signals from the execution environment. Following the work of CWM, we synthesized function-level and repository-level variable tracing data, followed by diverse rewriting-based data augmentation~\citep{copet2025cwm}.
    \item \textbf{Self Reflection}: Executing single or multi-turn self-correction of generated code in response to environmental feedback.
\end{itemize}
\vspace{-1.0em}

\begin{figure}
    \small
    \centering
    \includegraphics[width=0.8\linewidth]{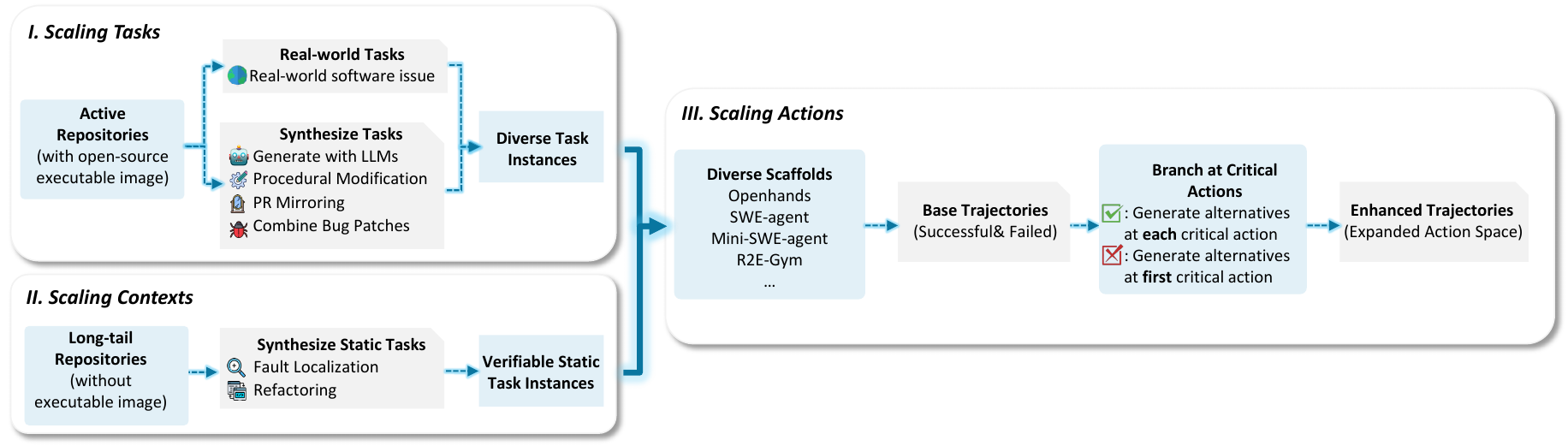}
    \caption{The Synthesis pipeline of Code Trajectories.}
    \label{fig:code_pipeline}
\end{figure}

\paragraph{End-to-End Code Agent Trajectories} To efficiently scale the end-to-end code agent trajectories, our approach focuses on three dimensions: (1) \textbf{Scaling Tasks} by creating diverse environments and new task instances; (2) \textbf{Scaling Contexts} by using a wide variety of \textit{long-tail} repositories through verifiable static tasks; and (3) \textbf{Scaling Actions} by branching from key editing and testing steps to reuse both successful and failed trajectories. The overall pipeline is illustrated in Figure~\ref{fig:code_pipeline}.

\noindent \textit{Scaling Tasks.} 
To broaden the scope of task environments, we leverage multiple open-source sandboxed environments, including SWE-gym~\citep{pan2024training}, SWE-smith~\citep{yang2025swe}, and SWE-rebench~\citep{badertdinov2025swe}. Beyond utilizing existing instances, we extend these environments with new synthesized instances following SWE-smith~\citep{yang2025swe}. To decouple model performance from specific scaffolding or toolset dependencies, we generate diverse issue-resolving trajectories using various scaffolds such as Mini-SWE-agent, SWE-gym, OpenHands, and R2E-gym. The toolsets employed encompass \texttt{bash}, \texttt{search}, \texttt{file\_editor}, \texttt{str\_replace\_editor}, \texttt{submit}, and \texttt{finish}. Notably, we prioritize \texttt{bash-only} trajectories, as \texttt{bash} represents the fundamental primitive from which more complex tools are abstracted. Furthermore, we extend our task domain to other real-world scenarios, including data science, front-end generation, game development, and specialized agents like ExcelAgent.

\noindent \textit{Scaling Contexts.} 
A significant bottleneck in software engineering (SE) tasks is the high cost of constructing complete Docker environments and corresponding test suites. While automated pipelines exist, their non-negligible failure rates necessitate manual verification~\citep{yang2025swe}, creating a bottleneck for efficient scaling. To mitigate this, we construct environments based on repositories that lack executable images. For these, we design tasks that require static evaluation rather than dynamic testing, such as \textbf{fault localization} and \textbf{refactoring}. In these scenarios, agents interact with the environment using \texttt{search} or other \texttt{bash} tools to identify specific faulty files or perform global renaming of variables and functions. While these tasks may be simpler than resolving complex GitHub issues, they are highly scalable and verifiable. This approach exposes the model to a vast diversity of codebases, fostering capabilities in proactive exploration, dynamic repository comprehension, and precise localization.

\begin{figure}[t]
    \centering
    \begin{subfigure}[b]{0.48\textwidth}
        \centering
        \includegraphics[width=\textwidth]{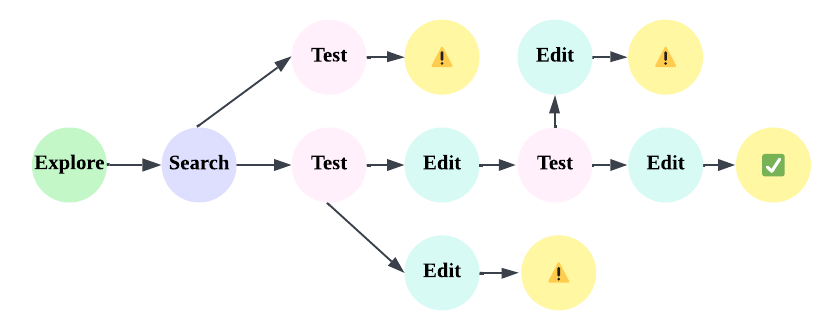}
        \caption{Successful Trajectories.}
        \label{fig:traj_success}
    \end{subfigure}
    \hfill 
    \begin{subfigure}[b]{0.48\textwidth}
        \centering
        \includegraphics[width=\textwidth]{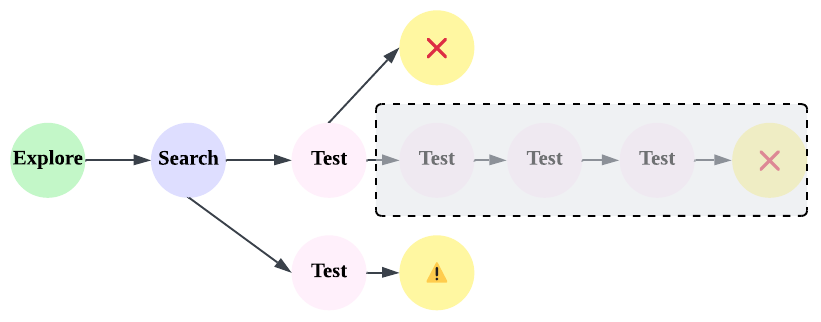}
        \caption{Failed Trajectories.}
        \label{fig:traj_failed}
    \end{subfigure}
    
    \caption{Trajectory branching strategy for code agent trajectories. We identify \textbf{Editing} and \textbf{Testing} as \textit{Critical Actions} that determine problem-solving success. For successful paths, we expand variations at each critical step; for failed paths, we generate a single branch at the initial critical action to repurpose data effectively while mitigating error propagation.}
    \label{fig:trajectory_branching}
\end{figure}
\noindent \textit{Scaling Actions.}
Integrating trajectories into the pre-training phase poses two challenges: (1) pure behavioral cloning (BC) may lead to premature collapse of the action space, capping the model's potential~\citep{ross2011reduction}; and (2) reusing failed trajectories might introduce suboptimal performance. However, discarding failed trajectories constitutes a significant loss of data. We identify two \textit{Critical Actions} that largely determine problem-solving success: \textbf{Editing} (\eg, \texttt{file\_editor} or \texttt{sed} via \texttt{bash}) and \textbf{Testing} (\eg, \texttt{pytest} or \texttt{unittest}). We extract the context preceding these actions and generate branched trajectories as follows:

\begin{itemize}
\vspace{-1.0em}
    \item \textbf{Successful Trajectories}: We generate alternative branches at every critical action point to enrich the model's action space.
    \item \textbf{Failed Trajectories}: We generate a single branch at the first critical action and truncate the sequence thereafter. Trajectories devoid of critical actions are filtered out.
\end{itemize}
\vspace{-1.0em}

To prevent the propagation of errors from unverified branches, we limit this expansion to a single-step rollout and a single variation per action. Additionally, we insert an evaluation step following the branched action to assess the state of task completion. Two representative examples of this branching strategy are provided in Figure~\ref{fig:trajectory_branching}. This strategy enables the effective reuse of both successful and failed trajectories while mitigating overfitting and enhancing the model's performance ceiling.

\subsubsection{Deep Research  Trajectory} 
\paragraph{Motivation and Goal} Deep Research (DR) has emerged as a pivotal application scenario for autonomous agents, representing a significant leap from simple information retrieval to complex, self-directed knowledge discovery. In this scenario, agents are required to navigate intricate environments, verify heterogeneous information, and synthesize findings into coherent outputs. However, the development of such capable agents is currently hindered by a scarcity of high-quality process data that captures the full spectrum of DR behaviors. Thus, our primary goal is to bridge this gap by constructing a large-scale, diverse DR trajectory and behavior dataset which empower models with necessary skills to handle the complexity of real-world research tasks with high precision and depth.

In the this scenario, we classify DR tasks into two broad categories: \textbf{closed-ended tasks}, \eg multi-hop QA which typically possess well-defined, verifiable answers, and \textbf{open-ended tasks}, \eg generating a comprehensive research report or scientific paper. For the two categories, we design scalable \textbf{trajectory synthesis pipelines} tailored to their specific nature to ensure high-quality data production. Furthermore, we augment these complete trajectories with specialized datasets targeting \textbf{atomic capabilities}, serving as an enhancement to reinforce the fundamental skills essential for effective research.

\paragraph{Closed-ended Trajectory Synthesis}
For the closed-ended setting, we primarily construct multi-hop QA trajectories. 
As shown in Figure~\ref{fig:close_dr_pipeline}, the overall data synthesis process proceeds as follows: First, we extract multi-hop QA pairs from seed corpora with high knowledge density. These seed sources include knowledge-centric web pages, multidisciplinary question banks, textbooks, and academic papers.
Next, we evaluate the quality of the extracted QA pairs, filtering out low-quality examples. Once high-quality QA pairs are obtained, we synthesize trajectories using various frameworks, incorporating multiple strategies to enhance trajectory diversity.
Given the verifiability of closed-ended tasks, we not only retain successful trajectories that lead to correct answers but also make use of failed trajectories through further processing. The detailed process is listed as follows:

\begin{figure}
    \centering
    \includegraphics[width=0.95\linewidth]{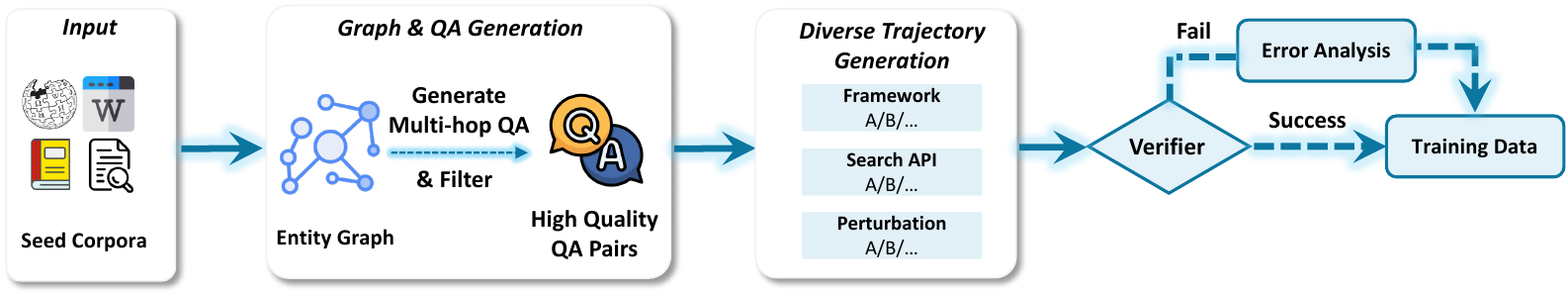}
    \caption{The trajectory synthesis pipeline of the closed-ended DR.}
    \label{fig:close_dr_pipeline}
\end{figure}

\textit{QA Generation.} To construct large-scale, high-quality QA trajectories, we primarily leverage high knowledge-density content from both pre-training and mid-training corpora as seed sources. To ensure content diversity, the seed data used for QA generation goes beyond typical encyclopedic resources and includes domain-specific materials across multiple disciplines, such as textbooks, academic papers, and professional question banks.
From these knowledge-rich sources, we extract a large number of entity-knowledge pairs,  
which serve as the foundation for generating multi-hop QA data. 
To ensure the quality of the QA pairs, we further apply an LLM-as-a-judge approach to score each QA instance across several dimensions, including factual accuracy, question clarity, answer uniqueness, and multi-hop reasoning consistency.
As a result of this pipeline, we obtain a large-scale dataset consisting of millions of high-quality QA pairs.

\textit{Trajectory Diversification.}
To enhance the diversity of synthesized trajectories, we employ multiple agent frameworks (\eg, Youtu-Agent \citep{youtu-agent-2025}, WebDancer \citep{Wu2025WebDancerTA}) and search APIs during the generation process.
To further increase trajectory variability, we introduce perturbations to the search results. Specifically, at certain steps, we randomly mask specific sources or suppress the top-K ranked results. These perturbations increase the difficulty of the QA task, often resulting in longer and more complex trajectories.
Moreover, the increased diversity in search results leads to greater contextual variety within the trajectories. Such data contributes to improving the robustness and generalization ability of the trained agent models.

\textit{Failed Trajectories Utilization.}
Failed trajectories also contain valuable information, and they are not entirely erroneous at every step. Incorporating such trajectories during pre-training is an important strategy for scaling up the amount of usable data.
For all trajectories identified as incorrect in answering the given question, we append a trajectory analysis segment. This analysis explicitly indicates that the trajectory fails to produce a correct answer and provides a step-by-step examination of the reasoning process, culminating in a summary of the reasons behind the incorrect response.

\begin{figure}
    \centering
    \includegraphics[width=1\linewidth]{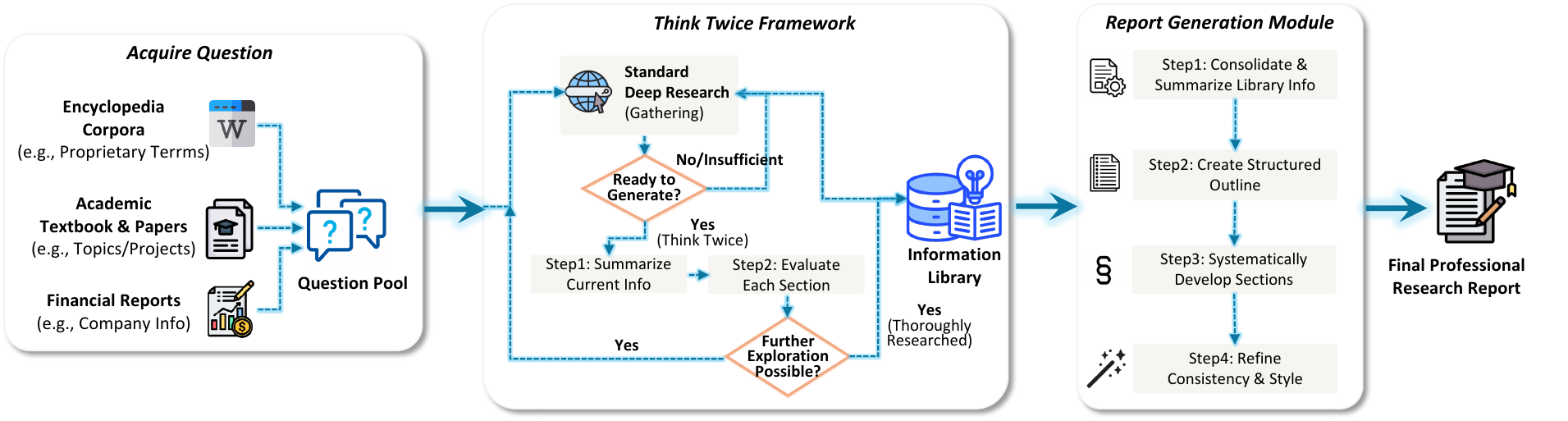}
    \caption{The trajectory synthesis pipeline of the open-ended DR report generation.}
    \label{fig:open_dr_pipeline_report generation}
\end{figure}

\paragraph{Open-ended Trajectory Synthesis}
For open-ended DR tasks, where the primary objective is to generate comprehensive research reports, we employ a dual-pronged strategy combining forward and inverse trajectory synthesis to effectively scale data production. This hybrid approach allows us to simultaneously simulate the authentic, iterative research workflow through forward generation, while ensuring high data quality and factual grounding by reconstructing search paths backwards from expert-level documents via inverse synthesis.

\textit{Forward Trajectory Synthesis.}
This approach constructs trajectory data in a forward manner following the common deep research process. To generate a large volume of high-quality data, we need to consider: (1) how to obtain a substantial number of meaningful queries for each scenario; (2) how to ensure sufficiently thorough research; and (3) how to produce professional report writing that meets specific style requirements. To this end, we have developed a comprehensive data synthesizing pipeline, which includes the following stages, as shown in Figure~\ref{fig:open_dr_pipeline_report generation}.

(1) \textbf{Acquisition of questions}: Research and analysis question sources are obtained from three scenarios: a) Extract proprietary terms from encyclopedic corpora; b) Extract relevant knowledge or generate research topics and projects from academic textbook and paper corpora; c) Acquire basic company information from known financial reports. These three categories represent typical scenarios for research and analysis, ensuring the generation of valuable trajectories.

(2) \textbf{Thinking twice framework}: When applying the existing DeepResearch framework for report research, there is often a tendency toward insufficient investigation, as the model may mistakenly assume that it has already gathered sufficient material to answer the given question. However, for report writing tasks, broad exploration is essential to capture as complete and relevant information as possible regarding the research topic. To address this, we propose a "think twice" method. This approach requires the model, when it believes it is ready to generate the report, to engage in a systematic reconsideration. Specifically, it first summarizes all currently obtained information, and then sequentially evaluates each section to determine whether further exploration is still possible. This process continues until every aspect that requires elaboration is thoroughly researched. Through this mechanism, all potentially relevant information can be comprehensively gathered.

(3) \textbf{Report generation Module}: This module is designed to synthesize the final report from the gathered information. Since data retrieved via the web search API is often fragmented, generating a report strictly following the initial research path can introduce bias, as earlier findings may be overlooked. To solve this, the module first consolidates and summarizes all the information recorded in information library, then creates a structured outline to guide the narrative. Each section is systematically developed using the compiled material, ensuring balanced coverage. Finally, the complete report is refined for consistency and style.

\textit{Inverse Trajectory Synthesis.}
Since it is difficult to directly evaluate the quality of forward-generated content in open-ended scenarios, we propose leveraging existing high-quality content to generate trajectories via inverse synthesis. 
In inverse trajectory synthesis, we generate the intermediate trajectory steps back from the final outcome.
This approach offers several advantages.
First, there is a large volume of readily available high-quality open-ended content, such as academic papers, research reports, and legal documents, which can serve as reliable targets for inverse synthesis. 
Second, the outcomes of inverse trajectory synthesis are deterministic; by starting from verified high-quality content, we can ensure that the final outputs of the trajectories are of high quality.
The main challenge lies in ensuring that the inversely generated trajectories contain a complete and coherent chain of context that is sufficient to reconstruct the target content.

\begin{figure}
    \centering
    \includegraphics[width=0.8\linewidth]{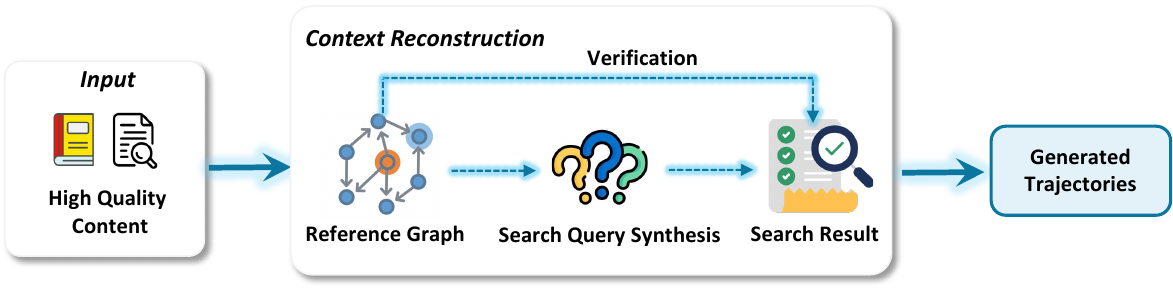}
    \caption{The inverse trajectory synthesis pipeline of the open-ended DR.}
    \label{fig:inverse}
\end{figure}

To address the challenges of inverse trajectory synthesis, we carefully select suitable domains, focusing primarily on academic papers and legal decision documents. 
In these domains, citations in academic texts and references to specific statutes in legal decisions serve as inherently reliable relevance signals. 
These citations reflect expert-level judgment in identifying highly pertinent sources from a vast body of academic literature or legal documents.
As shown in Figure~\ref{fig:inverse}, the core of result-based inverse trajectory synthesis lies in reconstructing the context that leads to the final output. 
We leverage the reference relationships captured in graph structures to identify truly relevant referenced content. Based on this content, we generate multi-turn search queries that, through interaction with search APIs, gradually approximate the cited materials.
Search results that is on the reference graph are treated as verified, correct search trajectories. 
Finally, we aggregate these Search API calls and refine them to produce coherent and high-quality trajectory data.
Using this approach, we construct over 20B tokens of open-ended trajectories with an average length around 32K.

\paragraph{Atomic Capability of DR} Besides trajectories, we also decompose the DR process into the following critical atomic capabilities. Explicitly reinforcing these individual skills is also important for building a robust foundation that sustains complex, multi-step research.

\begin{itemize}
\vspace{-1.0em}
    \item \textbf{Plan for solving problems:} Used to break down problems, determine the relevant web search APIs to call, and formulate the next steps for the deep research plan.

    \item \textbf{Self-Reflection:} Analyzes the progress or correctness of the current state, providing improvements or determining the next steps.

    \item \textbf{Summary:} Summarizes the results of searches and visited web content, as well as consolidates the information that has already been obtained.

    \item \textbf{Reading Comprehension:} Identifies relevant information from a large amount of context and integrates it to draw conclusions for each turn.
\end{itemize}
\vspace{-1.0em}

To effectively enhance these capabilities, we have specifically constructed a dedicated set of datasets as follows, allowing the agent to better master the core skills required for deep research.

\textit{Tool Parsing Data.}
We utilized the authentic responses from the search and web page visit APIs, and employed an LLM to rewrite the inputs and outputs for analysis, examining what was accomplished and what conclusions can be drawn. By doing so, we can gain a deeper understanding of the operational mechanisms of the core tools for deep research, as well as the processes of parsing, summarizing, and planning next steps based on the searched content.

\textit{Trajectory Understanding Data.}
A complete trajectory captures the full process of solving a problem. We claim that such trajectories contain more than just procedural steps—they encapsulate deep-seated problem-solving methodologies and experiential insights. To extract and articulate this implicit knowledge, we employ LLM-powered rewriting to summarize and review the entire trajectory. This process involves interpreting key moments within the trajectory, analyzing critical decision points, and drawing meaningful conclusions. We further examine the reasoning process and outcomes to assess their validity and explore potential optimizations. By doing so, we aim to explicitly capture the underlying methodologies embedded in the trajectory, thereby enhancing the model’s capabilities in both summarization and reflection reasoning.

\textit{Reading Comprehension Data.}
A specialized dataset focused on reading comprehension is constructed. A large number of contexts from a corpus is collected first. For each context, we generated accurate question-answer pairs. These three-tuples (context, question, answer) formed the foundational data for reading comprehension. To further augment the complexity of the reading comprehension tasks, the context from each three-tuple was mixed with other, unrelated contexts. This process resulted in composite contexts that contained both relevant and irrelevant information. This methodology yielded the finalized reading comprehension dataset.

\paragraph{Data Statistical Information}

Through the combination of closed-ended, open-ended synthesis trajectories (incorporating both forward and inverse construction methodologies) and atomic deep research capabilities data, a comprehensive DR dataset of 60B tokens has been developed. This corpus encompasses a wide range of prevalent DR scenarios, thereby ensuring coverage of critical capability dimensions.


\subsubsection{Tool-use and Planning Trajectory} 
\paragraph{Motivation and Goal} Tool-use and strategic planning ability determines how the LLM interacts with the environment, moving beyond static internal reasoning to dynamic, goal-oriented problem solving. In fact, in the several types of major agentic trajectory data mentioned above (\eg, coding and deep research trajectories), tool calling and task planning have naturally become an integral part of them. Additionally, we compiled more tool invocation and task planning trajectory data that is not limited to a specific domain, and incorporated the high-quality portions into the trajectory pre-training phase of Youtu-LLM.

\begin{figure}[t]
    \centering
    \includegraphics[width=0.90\linewidth]{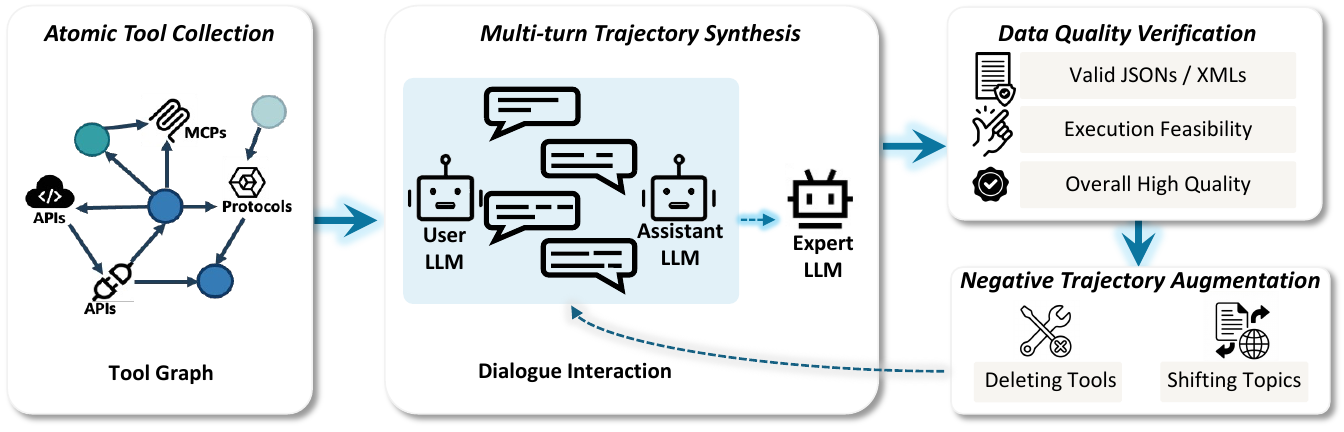}
    \caption{Our trajectory construction strategy for synthesizing tool-use and planning data.}
    \label{fig:tool}
\end{figure}

\paragraph{Trajectory Construction Strategy} Our approach to synthesizing 25B-token tool-use and planning data is formalized as the following four steps in Figure \ref{fig:tool}: (1) Collecting atomic tools; (2) Synthesizing multi-turn trajectories; (3) Verifying trajectory quality; (4) Expanding negative samples.

\noindent \textit{Atomic Tool Collection.} We first constructed a mixed skill library from three perspectives: breadth, depth, and dependencies. In terms of breadth, referencing the tool collection methods in existing work such as Toucan-1.5M~\citep{xu2025toucan} and FunReason-MT~\citep{xu2025funreason}, we collected thousands of diverse tools, including APIs, MCPs, and protocols; in terms of depth, our skill library includes simple actions such as weather and date search, and also covers professional tools such as text2sql and code interpreters. Furthermore, we retained the direct dependencies between skills as much as possible (\eg, retrieve the table $\rightarrow$ search over the table $\rightarrow$ modify the table). We performed verifiable checks and parameter corrections on the above atomic skill library to ensure the completeness of each tool, finally generating a tool graph.

\noindent \textit{Multi-turn Trajectory Synthesis.} Based on the tool graph, we synthesized multi-turn tool-use trajectory data with planning. We first randomly walked through the tool graph to generate a valid tool execution chain that conforms to the dependencies between tool calls. This is equivalent to constructing outline-level prompt information for the LLM. Based on this execution chain, two different LLMs were used for adversarial generation, playing the roles of User and Assistant, simulating the generation of multi-turn agentic trajectories that follow the entire tool execution chain. During synthesis, we randomly inserted additional rules on the User side, such as vague statements, repeated requests, spelling errors, and personality simulation. Finally, based on the complete trajectory and tool execution chain, a new LLM, defined as Expert, summarized the trajectory and generated planning data.

\noindent \textit{Data Quality Verification.} After obtaining the above trajectory data, we performed a three-step quality check. The first step is format checking, which involves verifying the correctness of JSON and XML formats within the tool calls, ensuring that the called tools actually exist, and that the parameters used are valid and complete. The second step is tool call feasibility checking; we locally re-executed calling tests on data that used coding and mathematical executors. The third step is quality checking, where an LLM comprehensively judges the quality of each trajectory.

\noindent \textit{Negative Trajectory Augmentation.} From the validated trajectories, we randomly selected 5\% for negative augmentation, using strategies such as deleting some available tools, modifying the User model's statements, inserting topic shifts, and then asked the User and Assistant LLMs to rewrite the trajectories.

%% file: 3_pretrain_method.tex
\section{Pre-training of Youtu-LLM}
\label{chp:3_pretrain}

In this section, we will report on the pre-training of Youtu-LLM. We first describe the architecture and hyperparameter configuration, including the acquisition of the STEM-oriented tokenizer and the adoption of the dense MLA architecture. Secondly, we present the complete multi-stage pre-training recipe.

\subsection{Architecture and hyperparameter}
\subsubsection{Tokenizer}

\begin{table*}[h]
    \centering
    \small
    \renewcommand{\arraystretch}{1.1}
    \setlength{\tabcolsep}{6pt}
    \begin{tabular}{l|ccc|cc}
    \toprule
    \textbf{Tokenizer} & \multicolumn{3}{c|}{\makecell{\textbf{Benchmark Performance} \\ Commonsense \quad STEM \quad Coding}} & \textbf{Compression Rate} & \textbf{Vocab Size} \\
    \midrule
    Llama3 & \quad\quad\quad 34.3 & \quad\quad 19.7 & 30.8 & 1.00 & 128,256 \\
    Qwen3 & \quad\quad\quad 32.3 & \quad\quad 21.5 & 30.3 & 1.10 & 151,936 \\
    \midrule
    Mix & \quad\quad\quad 34.9 & \quad\quad 20.6 & 29.2 & 1.13 & 128,256 \\
    Multi-stage (Youtu-LLM) & \quad\quad\quad 35.0 & \quad\quad 23.1 & 30.6 & 1.15 & 128,256 \\
    \bottomrule
    \end{tabular}
    \caption{We compare our tokenizer with two mainstream paradigms: Qwen3's tokenizer and Llama3's tokenizer. "Mix" indicates a conventional training strategy, that is, directly training tokenizer over pre-training data mix-up. "Multi-stage" refers to our multi-stage training method, which is ultimately adopted by Youtu-LLM. Compression ratio refers to the relative number of tokens produced by other tokenizers compared to the number of tokens produced by the Llama3 tokenizer on the same corpus, with the latter serving as the baseline (1.0).}
    \label{tab:tokenizer}
\end{table*}

Our tokenizer is a byte-level BPE (BBPE) tokenizer~\citep{sennrich2016neural,wang2020neural} trained using a modified HuggingFace Tokenizers training pipeline provided by superBPE~\citep{liu2025superbpe}. We adopt a stricter pre-tokenization scheme. Specifically, we segment all Chinese characters, Japanese kana, Hangul letters, and CJK punctuation into four standalone units, preventing them from merging with each other or with other parts of the text. For English, we largely follow the pre-tokenization design used in GPT-4o: we allow a single punctuation mark or a space-prefixed capital letter to be followed by a sequence of lowercase letters, optionally with suffixes such as ’s, etc. For numeric tokenization, we avoid any multi-digit tokens and retain only the ten atomic digit tokens 0–9~\citep{yang2025number,ding2025you}. This pre-tokenization design largely prevents tokens from spanning unrelated semantic units and, in particular, mitigates the issue of noisy tokens that arises in mixed Chinese–English text or Chinese text with whitespace.

We start from the o200k vocabulary as our base. Because \citet{zhang2025speculating} have pointed out that some Chinese tokens in o200k are polluted, we retain only the first 100k tokens that contain solely ASCII characters, together with 1k additional tokens (not Chinese) to preserve basic multilingual and emoji coverage. We remove from o200k all tokens that contain at least one Chinese character, as well as tokens that are invalid under our new pre-tokenization scheme. The resulting 101k tokens constitute our base vocabulary. We continue training Chinese-specific tokens using our multi-domain Chinese corpus. We observe that Chinese token learning is strongly influenced by domain-specific vocabulary: for example, legal and patent corpora contain many technical terms whose frequencies are much higher than those of everyday words, which biases tokenizer training. To address this, we rebalance the corpus mixture to suppress the abnormally high frequency of such domain-specific terms. In addition, we remove Chinese tokens longer than four characters that tend to appear frequently only in a few highly specific scenarios. At this stage, the vocabulary is expanded from 101k to 121k tokens.

Given the strong demand for reasoning performance in current models, for Youtu-LLM, we also take reasoning-oriented scenarios into account at the tokenizer design stage. To this end, we further augment the vocabulary with specialized tokens for code and mathematical/technical content. Based on the collected Chinese and English reasoning corpora, we add 4k and 3k tokens, respectively. This multi-stage training strategy helps ensure a balance between basic language ability and domain-specific reasoning capacity. After adding 256 reserved tokens, our final vocabulary size is 128,256, consistent with the Llama3 series~\citep{grattafiori2024llama3herdmodels}. Compared with baselines such as the Qwen~\citep{bai2023qwentechnicalreport} tokenizer, our tokenizer achieves about a 5\% improvement in tokenization efficiency on general pre-training data, and this gain further increases to around 10\% on reasoning-oriented data, as shown by the compression rate in Table \ref{tab:tokenizer}. Using different tokenizers, we pre-trained multiple models from scratch with the same 1B GQA skeleton on 80B tokens, demonstrating that our multi-stage tokenizer performs best on Commonsense and STEM tasks.

\begin{table*}[htbp]
    \centering
    \small
    \renewcommand{\arraystretch}{1.1}
    \setlength{\tabcolsep}{6pt}
    \begin{tabular}{l|ccccc}
    \toprule
    \textbf{Model} & \multicolumn{2}{c}{\makecell{\textbf{Chinese} \\ Generative \quad Multi Choice}} & \multicolumn{2}{c}{\makecell{\textbf{English} \\ Generative \quad Multi Choice}} & 
\textbf{PPL ($\downarrow$)} \\
    \midrule
    GQA-1B & \quad\quad 6.0 & 35.6 & \quad\quad 17.8 & 49.6 & 16.5\\
    MLA-1B & \quad\quad 7.2 & 37.4 & \quad\quad 18.5 & 50.7 & 15.4\\
    \bottomrule
    \end{tabular}
    \caption{We compared GQA and MLA by pre-training 1B-parameter models from scratch on 500B tokens. We evaluated both models based on seven Chinese benchmarks and fifteen English benchmarks, and found that MLA-1B averagely outperformed GQA in all formats and languages tested.}
    \label{tab:mla_gqa}
\end{table*}

\begin{table*}[htbp]
    \centering
    \small
    \renewcommand{\arraystretch}{1.1}
    \setlength{\tabcolsep}{6pt}
    \begin{tabular}{l|cccccccc}
    \toprule
    \textbf{Model} & \textbf{Param.} & \textbf{Layers} & \textbf{Hidden Size} & \textbf{Attention} & \textbf{Q/KV heads} & \textbf{Max Len.} & \textbf{Vocab Size} \\
    \midrule
    Qwen3-1.7B & 1.70B & 28 & 2,048 & GQA & 16/8 & 40,960 & 151,936 \\
    SmoLM3 3B & 3.08B & 36 & 2,048 & GQA & 16/4 & 65,536 & 128,256 \\
    MiniCPM3-4B & 4.07B & 62 & 2,560 & MLA & 40/40 & 32,768 & 73,448 \\
    \midrule
    Youtu-LLM 2B & 1.96B & 32 & 2,048 & MLA & 16/16 & 131,072 & 128,256 \\
    \bottomrule
    \end{tabular}
    \caption{Architecture comparison between existing lightweight LLM and our Youtu-LLM.}
    \label{tab:arch}
\end{table*}

\subsubsection{Dense MLA}

As a lightweight LLM designed for on-device scenarios, we plan to adopt the dense Multi-Latent Attention (MLA)~\citep{deepseekai2024deepseekv2strongeconomicalefficient} architecture from the very beginning of Youtu-LLM's development. On the one hand, for on-device scenarios, the Mixture-of-Experts (MoE)~\citep{cai2025surveymixtureexpertslarge} architecture doesn't offer significant speed advantages compared to the vanilla dense architecture, as they requires more frequent I/O operations~\citep{yi2023edgemoe}; on the other hand, MLA introduces low-rank compression of the KV Cache and larger intermediate projection matrices~\citep{deepseekai2024deepseekv2strongeconomicalefficient}, which helps improve the expressiveness and inference performance of the attention mechanism with constrained model parameters~\citep{kimiteam2025kimik2openagentic,deepseekai2025deepseekv3technicalreport}, surpassing vanilla GQA. Our experiments have also validated this (Table\,\ref{tab:mla_gqa}).

In Table \ref{tab:arch}, we list the main features of Youtu-LLM and other similar-size designs. We take Qwen3-1.7B~\citep{qwen3technicalreport}, SmolLM3-3B~\citep{bakouch2025smollm3} and MiniCPM3-4B~\citep{hu2024minicpm} as the typical representations of lightweight LLMs. 
We use a standard architecture with 32 layers and a vocabulary size of 128,256. 
The MLA configuration of Youtu-LLM remains consistent with DeepSeek-V3~\citep{deepseekai2025deepseekv3technicalreport}, whose KV lora rank dimension is 512, Q lora rank dimension is 1,536, QK nope head dimension is 128, QK rope head dimension is 64, and V head dimension is 128. Therefore, with the input and output embeddings tied, Youtu-LLM has a total of 1.96B parameters.

\subsection{Multi-stage Pre-training} \label{sec:multi-stage pretrain}

Given that Youtu-LLM incorporated MLA, we follow the scaling laws introduced in DeepSeek-V3 to guide hyperparameter selection. Considering the training scale of 2B parameters and 10.84T tokens of data, we adopt an optimal global batch size of 12M tokens and a learning rate of 4e-4. To improve training stability, we design an initializer range value that is coupled with the hidden dimension: for non-embedding weights, the initialized standard deviation is set to $\small\sqrt{\frac{2}{5d}}$, where $d$ denotes the hidden dimension; for embedding weights, the initialized standard deviation is twice that of the non-embedding weights.

\begin{figure*}[h]
\centering
\includegraphics[width=\textwidth]{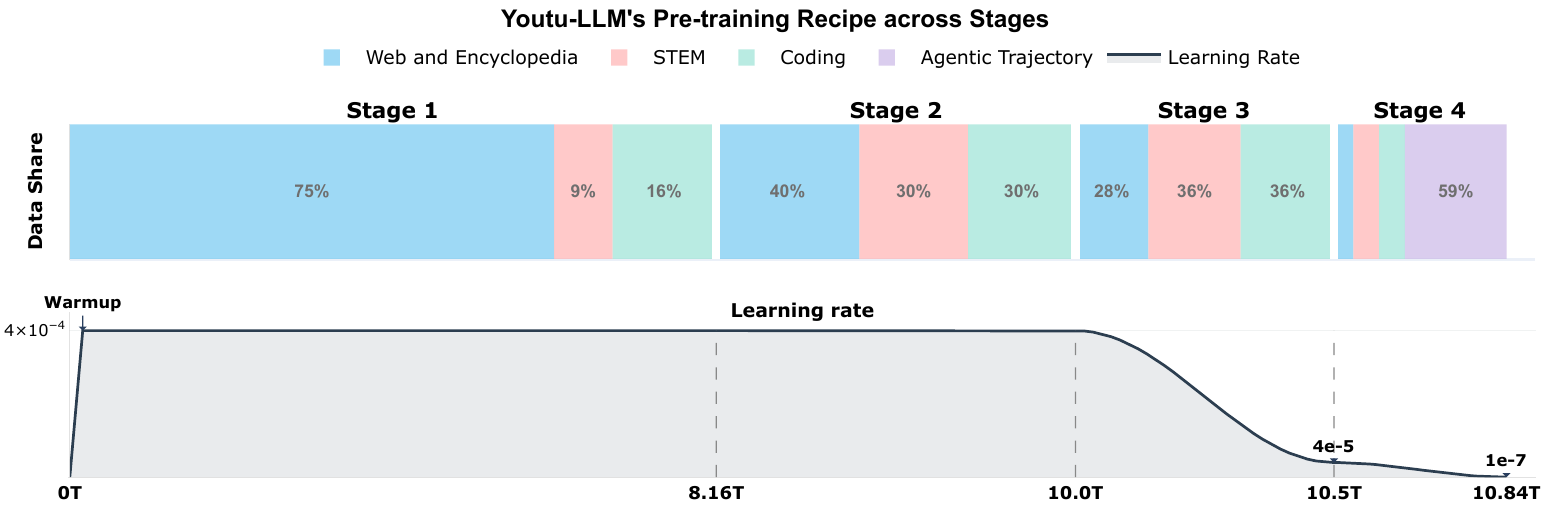}
  \caption{The pre-training recipe for Youtu-LLM. At the top, we illustrate the variations in the data recipe from Stage 1 to Stage 4, and it can be clearly observed that the average quality of the data gradually improves as the stages move forward. At the bottom, we draw the learning rate scheduler aligned with the data recipe.}
  \label{fig:pretrain_flowchat}
\end{figure*}

We demonstrate our entire pre-training recipe in Figure \ref{fig:pretrain_flowchat}. Youtu-LLM's pre-training is divided into four stages: Commonsense Pre-training (Stage 1), STEM- and Coding-centric Pre-training (Stage 2), General Mid-training (Stage 3), and Agentic Mid-training (Stage 4). The entire pre-training strategy follows the "Commonsense-STEM-Agent" design principle, which aligns with Jerome Bruner's Spiral Curriculum theory~\citep{bruner1960process} and David Ausubel's Progressive Differentiation principle~\citep{ausubel1960use}.

In Stage 1, we consume 8.16T tokens for pre-training from scratch with a sequence length of 8,192, with web pages and encyclopedia data accounting for 75\%. Regarding the learning rate, we follow the paradigm of most LLM pre-training work~\citep{deepseekai2025deepseekv3technicalreport,bakouch2025smollm3,olmo20242}. We warm up the model for 2000 steps. Next, in Stage 2, we significantly increase the proportion of STEM and Coding data to 60\% while maintaining the maximum learning rate. Stage 3 is a decaying and long context extension phase (8k -> 32k -> 128k), where the learning rate is reduced from the peak of 4e-4 to 4e-5. In this stage, the proportion of STEM and Coding data is further slightly increased, and Youtu-LLM gains more stable STEM, Coding, and long-context capabilities. Finally, in Stage 4, we adjust the decaying data to be primarily agentic trajectories (approximately 60\%), and further decay the learning rate to 1e-7, with specific training strategies applied (reach Section~\ref{sec:agent_skill} for details). Notably, we observed that training agentic trajectory data after general long-context training led to larger performance gains. This may be because the long-context model can better capture cross-segment key information in trajectory tasks.

%% file: 4_posttrain.tex
\section{Post-training of Youtu-LLM}
\label{chp:4_posttrain}

In this section, we introduce the post-training of Youtu-LLM, which mainly consists of Supervised Fine-Tuning and Reinforcement Learning.

\subsection{Supervised Fine-Tuning}
\label{subsec:sft}
Supervised Fine-Tuning (SFT) serves as the critical alignment phase where the pre-trained model transitions from a probabilistic next-token predictor to an instruction-following assistant. In this work, we implement a rigorous SFT framework, which consists of two main parts: high-quality data engineering and a decoupled two-stage training strategy.

\subsubsection{Data Engineering}
\label{subsubsec:data_engineering}
\begin{figure*}[h]
  \centering
  \includegraphics[width=0.95\textwidth]{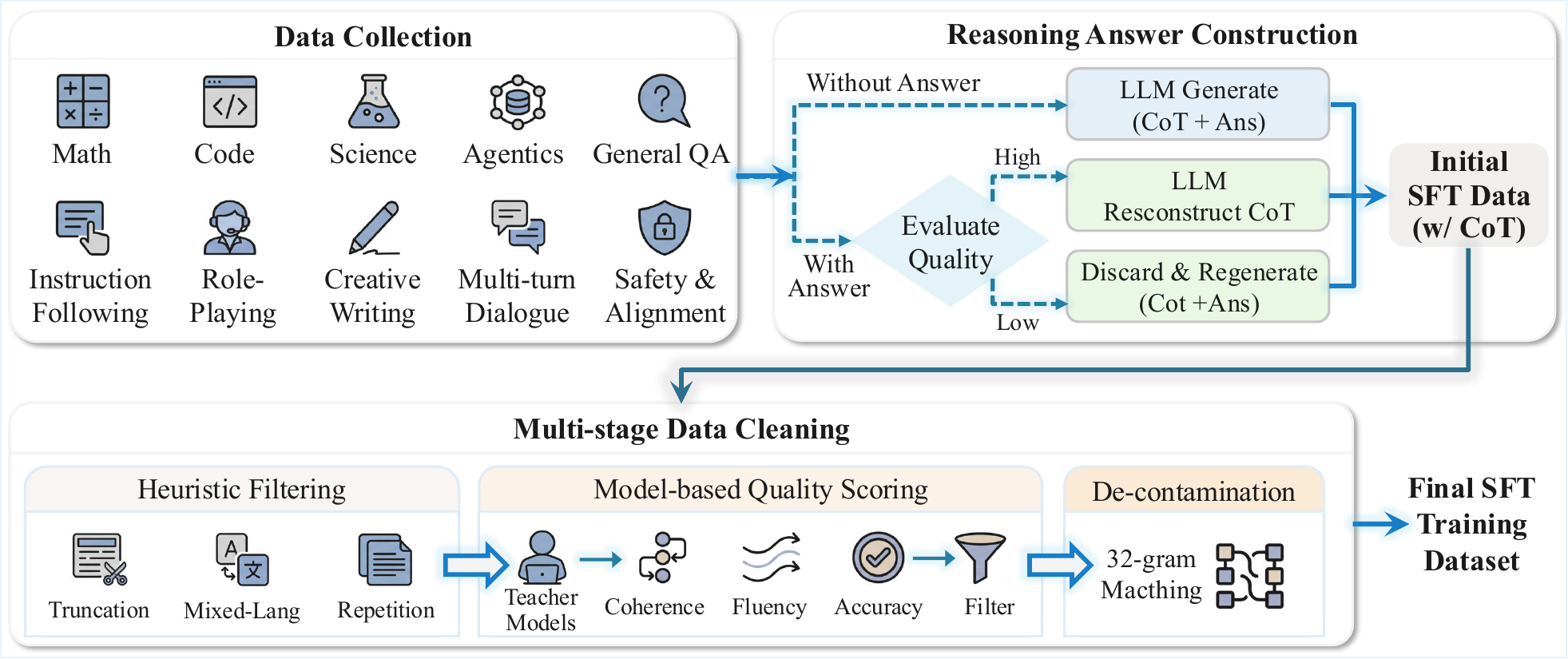}
  \caption{Overview of our data engineering workflow for high-quality Supervised Fine-Tuning.}
  \label{fig:sft_data_pipline}
\end{figure*}

The quality of instruction-tuning data is the primary determinant of model performance~\citep{guha2025openthoughtsdatarecipesreasoning,bakouch2025smollm3}. In this subsection, we outline our systematic approach to data collection, reasoning answer construction, and multi-stage data cleaning. The overall data processing pipeline is illustrated in Figure~\ref{fig:sft_data_pipline}.
\paragraph{Data Collection and Construction}
We begin by curating a comprehensive and heterogeneous dataset designed to ensure a broad distribution of knowledge and robust reasoning capabilities. To achieve this, we first collect data from a diverse array of sources, categorized as follows: (1) \textit{Mathematics:} Ranging from elementary arithmetic to collegiate-level calculus and competition mathematics.
(2) \textit{Code:} Containing high-quality repositories and competitive coding datasets.
(3) \textit{Scientific Reasoning:} Encompassing data in scientific fields such as physics, chemistry, biology, engineering physics, etc. (4) \textit{Agentic Data:} Covering trajectories involving precise function calls, deep research workflows (requiring multi-step information retrieval and summary), and code execution environments for dynamic task completion. (5) \textit{General Knowledge QA:} Covering diverse professional fields, such as Agriculture, Forestry, Animal Husbandry, Fishery, Law, Psychology and so on. (6) \textit{Instruction Following:} Containing a diverse set of complex constraints and formatting requirements. (7) \textit{Role-Playing:} Dialogues involving specific character profiles, historical figures, and fictional entities. (8) \textit{Creative Writing:} Including poetry, storytelling, narrative generation, scriptwriting, and rhetorical rewriting. (9) \textit{Multi-turn Dialogue:} Dialogue involving sequential context dependency and dynamic intent evolution. (10) \textit{Safety \& Alignment:} Containing refusal examples for toxic queries, bias mitigation, and adherence to ethical guidelines.

Followed by the collection of open-source data and proprietary corpora, we enrich the samples with explicit long Chain-of-Thought (CoT) reasoning processes. 
We categorize this \textbf{reasoning answer construction} pipeline into two scenarios: (1) \textit{Queries without Answers:} For open-ended questions or prompts lacking reference answers, we utilize advanced LLMs~\citep{liu2025deepseek,qwen3technicalreport} to generate comprehensive responses that include both the reasoning process and the final answer. 
(2) \textit{Queries with Answers:} We first evaluate the quality of the existing answers. For high-quality answers, we employ a few-shot prompting strategy with the advanced LLMs to reconstruct the reasoning contents that logically lead to the existing gold-standard answer, ensuring the thought process aligns with the correct conclusion. 
For low-quality answers that are deemed factually incorrect, we discard it and again utilized the modern LLMs to regenerate the entire response (Reasoning + Answer) from scratch. 
In this way, we successfully construct a high-quality initial SFT dataset enriched with detailed reasoning paths.

\paragraph{Multi-stage Data Cleaning}
\label{par:cleaning}
The process of generating answers via LLM APIs would inevitably introduce noise (\eg API return errors and generation failures). Such noise could degrade model performance, thus, we implement a comprehensive multi-stage cleaning pipeline to ensure high data quality.
\begin{itemize}
    \item \textbf{Heuristic Filtering:} We first apply a set of rule-based filters to remove low-quality samples. This includes detecting and removing: (1) \textit{Truncation errors:} Instances where the text ends mid-sentence or lacks a valid termination token. (2) \textit{Mixed-language noise:} Segments containing unintended garbled character encodings that do not align with the target language. (3) \textit{Repetition loops:} Text exhibiting pathological repetition of phrases, which often indicates generation failure in the source data.
    \item \textbf{Model-Based Quality Scoring:} To go beyond surface-level cleaning, we utilize state-of-the-art teacher models to act as judges. These teacher models score the samples based on three dimensions, including the logical coherence of the reasoning process, the fluency of the text, and the factual accuracy of the final answer. Samples falling below a calibrated confidence threshold are automatically discarded to prevent adverse effects on the model training.
    \item \textbf{De-contamination:} To ensure that the model's performance on benchmarks is not a result of memorization or leakage, we perform a de-contamination process against our evaluation benchmarks. Specifically, we utilize 32-gram exact matching to identify and remove any training samples that overlap with test sets. The remaining high-quality samples constitute our final training dataset.
\end{itemize}

\subsubsection{Two-Stage Supervised Fine-Tuning}
\label{subsubsec:two_stage_sft}
Many recent studies~\citep{bakouch2025smollm3,yang2025nanbeige4} advocate that distilling reasoning capabilities should precede the broad acquisition of general knowledge. 
Consequently, we adopt a two-stage SFT strategy to progressively elicit and refine the model's abilities. This strategy is designed to first "cold-start" the model's reasoning engines and subsequently refine its general versatility.

\paragraph{Stage I: Reasoning SFT}
The primary goal of this initial stage is to elicit and solidify the model's latent reasoning potential. As suggested by recent studies~\citep{qwen3technicalreport}, focusing exclusively on logic-dense data would facilitate the emergence of complex problem-solving skills by allowing the model to learn the structure of argumentation and step-by-step deduction without the interference of varied stylistic constraints.

In this stage, the training distribution is strategically skewed toward intensive intellectual tasks to develop the model’s higher-order reasoning abilities. Specifically, we curate a logic-dense dataset from our training corpus, including: \textbf{Mathematics} (40\%), \textbf{Programming Code} (30\%), \textbf{Scientific Reasoning} (20\%), and \textbf{Agentic Tasks} (10\%). The rationale behind this composition is that these domains inherently require step-by-step deduction, thereby forcing the model to learn the internal representations essential for deep thinking. 
We fine-tune the pre-trained base model utilizing a standard autoregressive cross-entropy loss. The training supervision is concurrently applied to both the intermediate reasoning trajectories and the final outputs. Finally, this training phase significantly strengthens the model’s chain-of-thought reasoning and structured response capabilities, providing a robust foundation for subsequent capability expansions.

\paragraph{Stage II: General SFT}
Following the reasoning enhancement, the second stage focuses on generalization and versatility. The objective is to equip the model with broad world knowledge and the ability to handle diverse user intents while preserving the reasoning capabilities acquired in Stage I.

In this stage, we expand the training corpus to encompass the full diversity of our collected data as detailed in Section \ref{subsubsec:data_engineering}. Crucially, to mitigate catastrophic forgetting of the reasoning skills, we integrate the training subsets from Stage I into this comprehensive mix. This strategy ensures that the model maintains the reasoning capability when aligning with a wide array of general user instructions.

Furthermore, to accommodate tasks that do not require complex reasoning, we introduce a simple yet effective strategy to induce a dual-mode ("think" and "non-think") capability. Specifically, we process the training data by stripping the intermediate reasoning content to treat them as non-thinking samples. And we train the model on these non-thinking samples along with original thinking samples. In this way, we successfully enabled the model to switch between a \textit{thinking mode} (generating explicit thought processes) and a \textit{non-thinking mode} (direct answering) by conditioning the model on specific control tokens. 
This approach allows developers to control reasoning behaviors, while also reducing the complexity of deploying separate models for thinking and non-thinking tasks.
We hypothesize that further refinement in the construction of non-reasoning data could yield superior performance, presenting a direction for future research.

\subsection{Reinforcement Learning}
\subsubsection{Tasks and Verifiers}
\paragraph{Math}
Mathematical data can typically be validated using objective criteria, making it highly effective for strengthening the foundational reasoning capabilities of models~\citep{guo2025deepseek,he2025skywork}. In the context of mathematical reasoning tasks, we have compiled a large-scale dataset comprising competition-level math problems, spanning a broad spectrum of difficulty—from elementary to advanced academic levels—and including content in both Chinese and English. Given the substantial overlap often present across multiple data sources, we performed rigorous deduplication using embedding-based similarity measures to ensure data uniqueness and quality.
To enhance verifiability, we systematically reformulated problem instructions so that model outputs adhere to structured formats that enable efficient automated validation. For example, we required answers to be explicitly presented within \verb|\boxed{}| notation or to compute multiple variables according to predefined formulaic patterns. We intentionally excluded multiple-choice questions, as their constrained answer space increases the risk of reward hacking and undermines genuine reasoning.
To improve generalization in general-purpose applications, we augmented the dataset with practical mathematical reasoning problems, including tabular comprehension and date calculation.
\paragraph{Code}
Our competitive programming dataset was developed through synthetic generation and licensed content, supplemented by problems with permissive intellectual property rights. To adapt these data for reinforcement learning, we implemented two distinct expansion strategies. The primary strategy involves autonomous code generation, where model-produced solutions are validated through a robust, scalable code execution environment. To guarantee the integrity of this validation process, we employ an iterative verification mechanism that cross-references standard solutions with test suites. The secondary strategy focuses on program execution simulation, tasking the model with inferring outputs from inputs (or the inverse)~\citep{li2025codei}. This paradigm is predicated on the belief that mastering input/output mapping fosters a deeper structural and logical understanding of the source code.
\paragraph{Complex Instructions}
We have collected a diverse set of over 100 foundational instructions covering tasks such as text comprehension, role-playing, and creative writing. We employ a hybrid rewards approach to validate this data. For simple tasks, we use rule-based code to verify constraints like word count and paragraph structure. For more complex instructions, we develop detailed rubrics and provide reference gold answers, utilizing an LLM-as-a-judge framework for scoring~\citep{gunjal2025rubrics}.
\paragraph{Safety}
We developed a specialized safe reward model trained on datasets containing harmful and misleading instructions. In terms of safety preferences, we prioritize informative guidance over categorical refusal. For instance, when encountering queries such as bomb making, the model is trained to explain the inherent legal and safety risks and pivot the conversation toward constructive topics, rather than issuing a standard 'I cannot answer this' disclaimer. Furthermore, we curated a suite of adversarial datasets—including techniques such as story continuation and role-playing—to perform safety red-teaming and enhance the model's robustness against jailbreaking attempts.
\paragraph{General Rewards}
To ensure the stability of the training process and prevent model degradation, we implement a suite of global validation strategies designed to safeguard against common failure modes. These strategies include:
\begin{itemize}
    \item \textbf{Thinking Format Preservation} In complex scenarios such as multi-turn dialogues, tool using, or long reasoning tasks, models often fail to maintain required structures (\eg, <think>...</think>). Such omissions lead to a degradation in output structuredness. We employ rule-based verification to ensure the completeness of these cognitive markers, penalizing outputs with fragmented or missing formats.
    \item \textbf{Language Consistency} Optimization techniques like Clip-Higher ~\citep{yu2025dapo} can inadvertently induce linguistic drift, resulting in erratic code-switching (\eg, mixing Chinese and English) or, in extreme cases, the generation of gibberish that triggers training collapse. To mitigate this, we pre-label the target language for each data entry. During reward calculation, we strip non-textual elements (\eg, code snippets and mathematical formulas) and verify that the dominant language of the remaining text aligns with the target. Samples exhibiting linguistic inconsistency are penalized to maintain cross-lingual purity.
    \item \textbf{Repetition Detection}
    To improve response diversity and prevent the model from collapsing into cyclic patterns, we implement a detection mechanism that monitors high-confidence $n$-gram frequencies.
    Given a sequence $S = \{id_0, id_1, \dots, id_{L-1}\}$ and its transition probabilities $P = \{p_1, p_2, \dots, p_{L-1}\}$, we define a multiset of "confident $n$-grams" $\mathcal{G}$. An $n$-tuple is recorded only when its generation probability $p_i$ exceeds a confidence threshold $\tau_p$,
    $$\mathcal{G} = \{ (id_{i-n+1}, \dots, id_i) \mid p_i > \tau_p, \quad n-1 \le i < L \}$$
    Let $C(g)$ denote the frequency of an $n$-gram $g$ within $\mathcal{G}$. We define $\text{unique}(\mathcal{G})$ as the set of distinct $n$-gram types present in $\mathcal{G}$, such that each unique pattern is counted only once regardless of its frequency. We isolate the subset of recurring patterns $\mathcal{G}_{dup}$ where $C(g) > 1$. The repetition ratio $\mathcal{R}$ is calculated as,
    $$\mathcal{R} = \frac{\sum_{g \in \ {unique}(\mathcal{G}), C(g) > 1} C(g)}{L}$$
    By identifying sequences where $\mathcal{R} > \tau_r$, this method serves as a trigger to suppress the likelihood of the model selecting repetitive patterns. This intervention forces the generator to explore alternative token paths, increasing the semantic diversity and structural variance of the generated response.
\end{itemize}

\subsubsection{Training Dynamics}
\begin{figure}[htbp]
    \centering
    \includegraphics[width=0.9\linewidth]{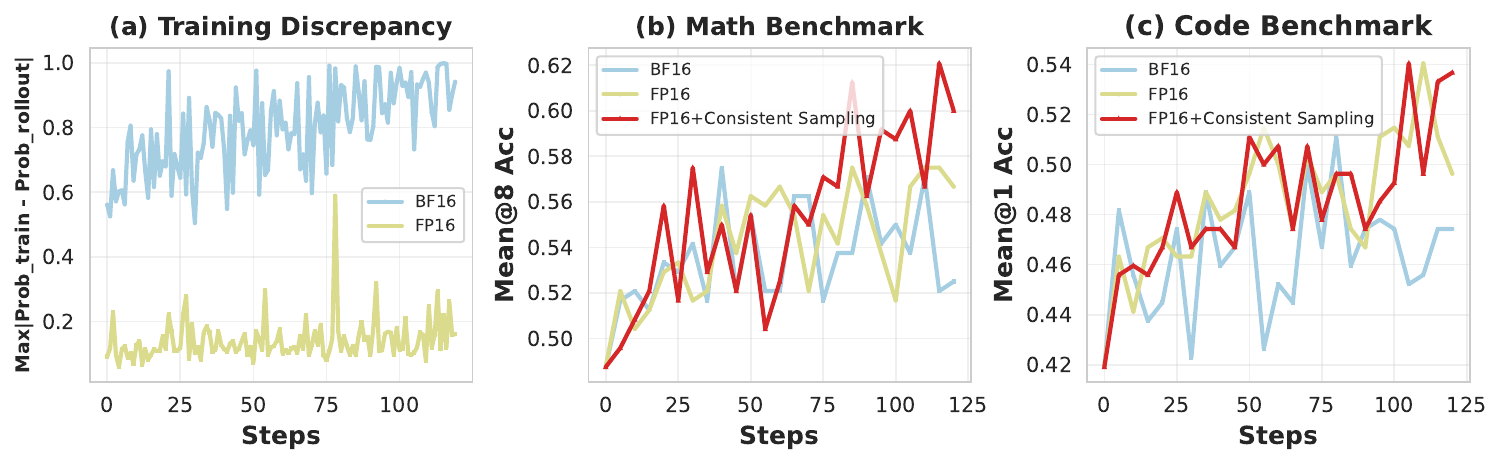}
    \caption{Training Dynamics with BF16/FP16 precision and consistent sampling. The reported accuracy metrics are derived from internal mathematical and coding benchmarks.}
    \label{fig:cons_sampling}
\end{figure}
Ensuring training-inference consistency is critical for maintaining the on-policy characteristics of reinforcement learning algorithms, which in turn facilitates superior training stability and enables longer optimization horizons~\citep{yao2025offpolicy,liu-li-2025-rl-collapse}. We have implemented several refinements to address this challenge, as detailed below.
\paragraph{FP16 Precision}
Discrepancies often arise between the training and inference stages due to the use of different operators or hardware kernels. These inconsistencies can cause the training policy to deviate significantly from the behavior observed during rollout.
As illustrated in Figure \ref{fig:cons_sampling}(a), when employing BF16 precision, the prediction probability error between training and inference grows progressively as iterations increase. This divergence introduces substantial noise into the optimization process, leading to performance plateaus. This is evident in Figure \ref{fig:cons_sampling}(b) and (c), where benchmarks for mathematical and coding stagnate after approximately 50 training steps.
In contrast, our findings align with prior research ~\citep{qi2025defeating} suggesting that FP16 precision is more effective than BF16 at maintaining training-inference consistency. As shown in Figure \ref{fig:cons_sampling}(a), the probability gap between train and rollout is markedly narrower when using FP16. This reduction in numerical drift allows for more precise gradient updates. The empirical results in Figure \ref{fig:cons_sampling}(b) and (c) validate that by adopting FP16, the model avoids early performance saturation and achieves superior results on both mathematical and coding benchmarks compared to the BF16 baseline.
\paragraph{Consistent Sampling}
We define train policy $\pi_\theta$ as the policy currently being optimized and rollout policy $\pi_{\text{rollout}}$ used for fast sampling.
For a given prompt $q$, the KL metric $\mathcal{K}(q)$ measures the divergence between the training and rollout policies.
$$\mathcal{K}(q) = \mathbb{E}_{a \sim \pi_{\text{rollout}}(\cdot|q)} \left[ \frac{\pi_{\theta}(a|q)}{\pi_{\text{rollout}}(a|q)} - 1 - \log \frac{\pi_{\theta}(a|q)}{\pi_{\text{rollout}}(a|q)} \right]$$
To ensure training stability, we introduce consistent sampling based on GRPO~\citep{shao2024deepseekmath}. A group of outputs $\mathcal{O}_q = \{o_1, \dots, o_G\}$ generated for prompt $q$ is admitted for training if and only if $\mathcal{K}(q) < \tau$, where $\tau$ is a hard threshold (\eg, $\tau = 0.01$). Groups exceeding this threshold are discarded prior to gradient computation, ensuring the model only learns from prompts where the policy drift is constrained.
The CF-GRPO objective is formulated over the filtered distribution $\mathcal{D}^*$, where every batch of size $G$ is guaranteed to satisfy the drift constraint. The objective $J_{CF-GRPO}(\theta)$ is defined as:
$$J_{CF-GRPO}(\theta) = \mathbb{E}_{q \sim P(Q|\mathcal{K}(q)<\tau), \{o_i\}_{i=1}^G \sim \pi_{\text{rollout}}} \left[ \frac{1}{G} \sum_{i=1}^G \min \left( \rho_i \hat{A}_i, \text{clip}(\rho_i, 1-\epsilon, 1+\epsilon) \hat{A}_i \right) \right]$$
with the importance sampling ratio $\rho_i$ and the advantage $\hat{A}_i$.
The empirical results in Figure \ref{fig:cons_sampling}(b) and (c) demonstrate that our approach significantly enhances training stability.

%% file: 5_experiment.tex
\section{Experimental Results}
\label{chp:5_exps}

We present the results of our model evaluation on a diverse array of benchmarks, spanning multiple domains, to comprehensively assess its capabilities. The following sections provide a detailed breakdown of its performance, involving general and agent benchmarks on our base and instruct models, respectively.

\subsection{General Evaluation}

\subsubsection{General Evaluation of Base Model}

We evaluate the general capabilities of the base model across four dimensions: Commonsense knowledge, STEM, Coding, and Long Context. The benchmarks involved are as follows:

\paragraph{Commonsense knowledge.} 
To assess broad world knowledge and reasoning in both English and Chinese, we utilize MLQA-Zh~\citep{lewis2020mlqa}, MMLU-ProX-Zh~\citep{xuan2025mmlu}, and MMLU-Pro~\citep{wang2024mmlu}. These benchmarks focus on evaluating world knowledge and provide insights into the model’s ability to handle both fact-based and conceptual questions. The suffix "-Zh" indicates that the evaluation is restricted to the Chinese subset.
For MMLU-Pro and MMLU-ProX-Zh, we conduct 5-shot evaluations and report \textit{Exact Match (EM)} accuracy. For MLQA-Zh, we use a 3-shot setting and also report EM.

\paragraph{STEM.} Mathematical ability is evaluated using a suite of datasets, including GSM8K~\citep{cobbe2021training} and MGSM-Zh~\citep{shi2022language} for grade-school math, MATH~\citep{hendrycks2021measuring} for competition-level problems, and Big-Bench-Hard (BBH)~\citep{suzgun2023challenging} for complex logical reasoning. Additionally, we include GPQA~\citep{rein2024gpqa} for expert-level scientific QA, and Humanity's Last Exam (HLE)~\citep{phan2025humanity} for human-like exam-style math questions. It is worth noting that GPQA and HLE are far more difficult to directly test with small LLMs. Therefore, we choose to evaluate with a multi-choice (-MC) paradigm. Unlike GPQA, HLE includes both multi-choice and generative exams, therefore we only retrieve the multi-choice subset for evaluation. In terms of evaluation settings, we use 8-shot for GSM8K and MGSM-Zh, and report strict match scores for both. We use 4-shot, 3-shot, 5-shot and 3-shot for MATH, BBH, GPQA-MC and HLE-MC respectively, following most conventions. EM is reported on MATH and BBH, while two multi-choice benchmarks are evaluated with normed accuracy.

\paragraph{Coding.} Coding proficiency is measured using MBPP~\citep{austin2021program} and MBPP+~\citep{liu2023your} for basic programming tasks, HumanEval~\citep{chen2021evaluating} and HumanEval+~\citep{liu2023your} for functional code generation, LiveCodeBench v6~\citep{jain2024livecodebench} for latest coding scenarios, CRUXEval~\citep{gu2024cruxeval}  for atomic coding, and RepoBench~\citep{liu2023repobench} for repository-level code understanding and generation. We report pass@1 success rate for all coding benchmarks except RepoBench, which is suggested to be evaluated with EM. MBPP and MBPP+ adopt 3-shot test, HumanEval and HumanEval+ use 0-shot, LiveCodeBench v6 takes 3-shot, and CRUXEval uses 1-shot. It's worth noting that the 1-shot setting in CRUXEval refers to the examples built into the prompt in its official code\footnote{https://github.com/facebookresearch/cruxeval/blob/main/prompts.py}.

\paragraph{Long-Context.} The model's ability over long context is evaluated using LongBench v2~\citep{bai2025longbench} and needle-in-a-haystack (NIAH) testing~\citep{hsieh2024ruler}. Specifically, we adapt LongBench v2 into a set of 3-shot multiple-choice questions designed to assess base model capabilities, and the overall accuracy is reported.
For NIAH evaluation, we use the HELMET suite \citep{yen2024helmet}, which incorporates the RULER dataset \citep{hsieh2024ruler} as its testbed for NIAH, the average recall rate is reported. 

\begin{table*}[h]
    \centering
    \footnotesize
    \renewcommand{\arraystretch}{1.1}
    \setlength{\tabcolsep}{6pt}
    \begin{tabular}{l|ccccc|c}
    \toprule
    \textbf{Benchmark} & \makecell[c]{\textbf{Qwen3 1.7B}\\\textbf{Base}} & \makecell[c]{\textbf{SmolLM3 3B}\\\textbf{Base}} & \makecell[c]{\textbf{Gemma3 4B}\\\textbf{Base}} & \makecell[c]{\textbf{Qwen3 4B}\\\textbf{Base}} & \makecell[c]{\textbf{Llama3.1 8B}\\\textbf{Base}} & \makecell[c]{\textbf{Youtu-LLM 2B}\\\textbf{Base}} \\
    \midrule
    \multicolumn{6}{l}{\textbf{Commonsense}} \\
    \midrule
    MMLU-Pro (5-shot) & 34.9 & 35.3 & 29.4 & \underline{46.1} & 36.2 & \textbf{48.4} \\
    MLQA-Zh (3-shot) & 38.1 & 38.0 & 40.1 & \textbf{47.2} & 43.0 & \underline{43.5} \\
    MMLU-ProX-Zh (5-shot) & 32.5 & 26.7 & 24.2 & \textbf{45.2} & 25.4 & \underline{40.7} \\
    \midrule
    \multicolumn{6}{l}{\textbf{STEM}} \\
    \midrule
    GSM8K (8-shot) & 68.2 & 67.3 & 38.5 & \textbf{80.8} & 47.8 & \underline{77.6} \\
    MGSM-Zh (8-shot) & 57.1 & 40.7 & 33.0 & \textbf{69.7} & 35.9 & \underline{68.9} \\
    MATH (4-shot) & 28.1 & 40.8 & 24.4 & \textbf{44.8} & 21.5 & \underline{44.4} \\
    BBH (3-shot) & 53.0 & 60.0 & 51.6 & \textbf{70.8} & \underline{62.9} & 60.0 \\
    GPQA-MC (5-shot) & 30.4 & 26.6 & 28.6 & \textbf{37.8} & 30.1 & \underline{33.3} \\
    HLE-MC (3-shot) & 10.7 & 3.1 & 8.0 & \underline{15.0} & 11.5 & \textbf{17.4} \\
    \midrule
    \multicolumn{6}{l}{\textbf{Coding}} \\ 
    \midrule
    MBPP (3-shot) & 55.6 & 51.0 & 45.8 & \textbf{67.5} & 49.4 & \underline{66.6} \\
    MBPP+ (3-shot) & 71.0 & 66.1 & 61.9 & \underline{80.8} & 62.7 & \textbf{81.8} \\
    HumanEval (0-shot) & 49.9 & 34.8 & 36.6 & \underline{57.6} & 36.0 & \textbf{64.6} \\
    HumanEval+ (0-shot) & 41.3 & 28.1 & 28.1 & \underline{49.9} & 28.1 & \textbf{57.3} \\
    LiveCodeBench v6 (3-shot) & 5.1 & 2.9 & 2.9 & \underline{6.9} & 3.4 & \textbf{9.7} \\
    CRUXEval (1-shot) & 40.6 & 42.1 & 39.7 & \underline{54.8} & 42.3 & \textbf{55.9} \\
    RepoBench (3-shot) & 21.0 & 21.8 & 23.0 & \textbf{25.3} & \underline{25.2} & 22.7 \\
    \midrule
    \multicolumn{6}{l}{\textbf{Long Context}} \\
    \midrule
    LongBench v2 (3-shot) & \underline{28.0} & \textbf{28.8} & 26.6 & 25.8 & 27.8 & 27.2 \\
    NIAH & 79.8 & 75.0 & \underline{99.5} & 83.0 & \textbf{99.8} & 98.8 \\
    \bottomrule
    \end{tabular}
    \caption{General benchmark performance of similar-sized \textbf{base} models. The top performance is highlighted in \textbf{bold}, and the second-place performance is marked with an \underline{underline}.}
    \label{tab:baseline_for_base}
\end{table*}

We evaluate the models in Table\,\ref{tab:baseline_for_base} based on the open-source evaluation framework lm-evaluation-harness\footnote{https://github.com/EleutherAI/lm-evaluation-harness}, and using the version of Transformers recommended in their own configurations. 
Overall, in terms of general capabilities, Youtu-LLM 2B Base significantly outperforms the similar sized baselines, and achieves competitive results to the larger Qwen3-4B Base.

\subsubsection{General Evaluation of Instruct Model}

For the instruct models, we evaluate them from four perspectives: Commonsense Knowledge, Instruction Following \& Text Reasoning, STEM, and Coding. The benchmarks involved are described as follows:

\begin{table}[h]
\centering
\scriptsize
\begin{tabular}{l|ccccc|c}
\toprule
\textbf{Benchmark} & \makecell[c]{\textbf{DeepSeek-R1-}\\\textbf{Distill-Qwen 1.5B}} & \textbf{Qwen3 1.7B} & \textbf{SmolLM3 3B} & \textbf{Qwen3 4B} & \makecell[c]{\textbf{DeepSeek-R1-}\\\textbf{Distill-Llama 8B}} & \textbf{Youtu-LLM 2B} \\
\midrule
\multicolumn{7}{l}{\textbf{Commonsense Knowledge Reasoning}} \\
\midrule
MMLU-Redux (0-shot)      & 53.0 & 74.1 & 75.6 & \textbf{83.8} & \underline{78.1} & 75.8 \\
MMLU-Pro (5-shot)        & 36.5 & 54.9 & 53.0 & \textbf{69.1} & 57.5 & \underline{61.6} \\
\midrule
\multicolumn{7}{l}{\textbf{Instruction Following \& Text Reasoning}} \\
\midrule
IFEval (0-shot)          & 29.4 & 70.4 & 60.4 & \textbf{83.6} & 34.6 & \underline{81.2} \\
DROP (3-shot)         & 41.3 & 72.5 & 72.0 & \underline{82.9} & 73.1 & \textbf{86.7} \\
MUSR (0-shot)            & 43.8 & 56.6 & 54.1 & \textbf{60.5} & \underline{59.7} & 57.4 \\
\midrule
\multicolumn{7}{l}{\textbf{STEM}} \\
\midrule
MATH-500 (0-shot)        & 84.8 & 89.8 & 91.8 & \textbf{95.0} & 90.8 & \underline{93.7} \\
AIME 24 (0-shot)        & 30.2 & 44.2 & 46.7 & \textbf{73.3} & 52.5 & \underline{65.4} \\
AIME 25 (0-shot)         & 23.1 & 37.1 & 34.2 & \textbf{64.2} & 34.4 & \underline{49.8} \\
GPQA-Diamond (0-shot)    & 33.6 & 36.9 & 43.8 & \textbf{55.2} & 45.5 & \underline{48.0} \\
BBH (3-shot)             & 31.0 & 69.1 & 76.3 & \textbf{87.8} & \underline{77.8} & 77.5 \\
\midrule
\multicolumn{7}{l}{\textbf{Coding}} \\
\midrule
HumanEval (0-shot)       & 64.0 & 84.8 & 79.9 & \underline{95.4} & 88.1 & \textbf{95.9} \\
HumanEval+ (0-shot)      & 59.5 & 76.2 & 74.7 & \underline{87.8} & 82.5 & \textbf{89.0} \\
MBPP (0-shot)            & 51.5 & 80.5 & 66.7 & \textbf{92.3} & 73.9 & \underline{85.0} \\
MBPP+ (0-shot)           & 44.2 & 67.7 & 56.7 & \textbf{77.6} & 61.0 & \underline{71.7} \\
LiveCodeBench v6 (0-shot)   & 19.8 & 30.7 & 30.8 & \textbf{48.5} & 36.8 & \underline{43.7} \\
\bottomrule
\end{tabular}

\caption{Comprehensive comparison across various benchmarks of similar-sized \textbf{instruct} models. Models are evaluated on commonsense knowledge, instruction following \& text reasoning, stem, and coding capabilities. The top performance is highlighted in \textbf{bold}, and the second-place performance is marked with an \underline{underline}.}
\label{tab:general_main_results_instruct}
\end{table}

\paragraph{Commonsense Knowledge.} We evaluate the models' foundational knowledge using MMLU-Redux~\citep{gema2025we} and MMLU-Pro~\citep{wang2024mmlu}. The accuracy score is reported. 

\paragraph{Instruction Following \& Text Reasoning.} To assess how well models adhere to complex constraints and reason through linguistic tasks, we utilize IFEval~\citep{zhou2023instructionfollowingevaluationlargelanguage}, DROP~\citep{dua2019drop}, and MUSR~\citep{sprague2023musr}. IFEval focuses on objective instruction following (\eg, formatting and keyword constraints), DROP tests discrete reasoning over paragraphs requiring mathematical operations on text, and MUSR evaluates the model's ability to perform long-chain logical deduction in narrative contexts. As for the evaluation metrics, we utilize the strict-prompt accuracy for IFEval, the F1 score for DROP, and the accuracy score for MUSR.

\paragraph{STEM.} Mathematical and scientific proficiency is measured across a spectrum of difficulty. We include MATH-500~\citep{lightman2023let} for competition-level mathematics and AIME (2024 and 2025 sets)~\citep{aime_2024_2025} for elite-level mathematical problem solving. For complex logical reasoning, we use BBH~\citep{suzgun2023challenging}. Furthermore, we evaluate expert-level scientific knowledge using GPQA-Diamond~\citep{rein2024gpqa}. When evaluating STEM benchmarks, we adopt zero-shot settings, except for BBH, where we employ 3-shot setup. For BBH, we report the accuracy score. And for other benchmarks, we report the pass@1 metric. Specifically, we generate 4, 16, 16, and 8 responses per question for MATH-500, AIME 24, AIME 25, and GPQA-Diamond respectively, to obtain an unbiased estimate of the pass@1 score.

\paragraph{Coding.} We measure coding proficiency using standard functional benchmarks and real-world scenarios. This includes HumanEval~\citep{chen2021evaluating} and MBPP~\citep{austin2021program}, along with their rigorous "+" versions~\citep{liu2023your}. Additionally, we use LiveCodeBench v6~\citep{jain2024livecodebench} to evaluate the model's performance on the latest competitive programming problems. When evaluating coding benchmarks, we adopt zero-shot settings and default to report the pass@1 metric. Here, for HumanEval/HumanEval+ and MBPP/MBPP+, we generate 4 responses for each question. And 16 responses are generated for LiveCodeBench v6 benchmark.

We evaluated the instruct models based on open-source evaluation frameworks, including evalscope\footnote{https://github.com/modelscope/evalscope} and evalplus\footnote{https://github.com/evalplus/evalplus}. Here, the majority of the benchmarks were conducted using evalscope, with the exception of HumanEval, MBPP, HumanEval+, and MBPP+, which were assessed using the official evalplus repository.
For all listed models, they are evaluated in a \textit{thinking mode} (generating explicit thought processes). We utilized a sampling temperature of $1.0$, a top-p value of $0.95$, a top-k value of $20$, and a presence penalty of $1.5$. Besides, we set the maximum output length to $32,768$ tokens to ensure sufficient thinking space.

We compare our Youtu-LLM 2B instruct model with DeepSeek-R1-Distill-Qwen 1.5B, Qwen3 1.7B, SmolLM3 3B, and DeepSeek-R1-Distill-Llama 8B. The evaluation results are shown in Table~\ref{tab:general_main_results_instruct}. The Youtu-LLM 2B model outperforms similarly sized baselines (such as Qwen3-1.7B and SmolLM3 3B) and even exceeds the larger DeepSeek-R1-Distill-Llama 8B model in most capabilities. Furthermore, it exhibits competitive performance against Qwen3 4B, particularly in coding, instruction following and text reasoning (\eg, HumanEval and DROP). These results underscore the efficacy of the proposed method in developing high-performance, light-weight LLMs.

\subsection{Agentic Evaluation}
\subsubsection{Agentic Evaluation of Base Model}

To evaluate the agent-oriented capability of the base model during the agentic pre-training, we use \textbf{APTBench}~\citep{qin2025aptbench} as the benchmark\footnote{https://github.com/TencentYoutuResearch/APTBench}.
APTBench serves as a benchmark specifically designed to evaluate agent capabilities of base models. It acts as a proxy metric for agent ability without requiring additional post-training or end-to-end evaluation. APTBench adopts a combination of few-shot multiple-choice question (MCQ), text completion (TC) and true/false questions, making it suitable for efficient assessment during the pre-training/mid-training phase.
Specifically, APTBench includes data from four domains: code, deep research, tool, and math.

\paragraph{Code.} This scenario evaluates the model's ability to handle complex coding tasks across the \textbf{Environment Setup} and \textbf{Issue Fixing}. 
To comprehensively assess the model's potential as an coding agent, the benchmark measures two core agent capabilities: planning and action.
Planning measures whether the base model knows which step should be executed next while action measures its ability to write correct bash commands.
We further test other important atomic capabilities of coding tasks, including error handling, bug localization and fix/test patch identification.
Details of the APTBench-Code are shown in Table~\ref{tab:apt-code}.

\begin{table}[htbp]
    \centering
    \small
    \renewcommand{\arraystretch}{1.5}
    \begin{tabularx}{\textwidth}{@{} l >{\RaggedRight\arraybackslash}X >{\RaggedRight\arraybackslash}X c @{}}
        \toprule
        \textbf{Task Type} & \textbf{Task Definition} & \textbf{Evaluation Mechanism} & \textbf{\# Instances} \\
        \midrule
        \textbf{Planning} 
        & Stepwise environment setup strategies or SWE debugging plans. 
        & MCQ: Distinguish the correct plans from the wrong ones. Ground-truth plans are extracted from real successful trajectories. 
        & 680 \\
        
        \textbf{Action} 
        & Execution of terminal commands based on prior context. 
        & TC: Write the exact next bash command based on the given trajectory context. 
        & 1,325\\
        
        \textbf{Atomic} 
        & Focuses on core engineering tasks like error handling, bug localization and fix/test patch identification. 
        & MCQ: Select correct fixes/code snippet from options including irrelevant or failed negative choices. 
        & 1,722\\
        \bottomrule
    \end{tabularx}
    \caption{Overview of APTBench-Code tasks.}
    \label{tab:apt-code}
\end{table}

\begin{table}[htbp]
    \centering
    \small
    \renewcommand{\arraystretch}{1.5}
    \begin{tabularx}{\textwidth}{@{} l >{\RaggedRight\arraybackslash}X >{\RaggedRight\arraybackslash}X c @{}}
        \toprule
        \textbf{Task Type} & \textbf{Task Definition} & \textbf{Evaluation Mechanism} & \textbf{\# Instances} \\
        \midrule
        \textbf{Planning} 
        & Stepwise decision-making for information seeking/organizing report generation. 
        & MCQ: Choose correct plans verified by ground-truth agent trajectories from the wrong plans. 
        & 1,329\\
        
        \textbf{Action} 
        & Synthesizes gathered information into outputs, from short answers to detailed reports. 
        & TC: Generate short answers from trajectories.
        & 564\\
        
        \textbf{Atomic} 
        & Checks citation accuracy and factual grounding of statements. 
        & MCQ: Distinguish supported statements from unsupported distractors. 
        & 362\\
        \bottomrule
    \end{tabularx}
    \caption{Overview of APTBench-DR tasks.} 
    \label{tab:apt-dr}
\end{table}

\begin{table}[htbp]
    \centering
    \small
    \renewcommand{\arraystretch}{1.5}
    \begin{tabularx}{\textwidth}{@{} l >{\RaggedRight\arraybackslash}X >{\RaggedRight\arraybackslash}X c @{}}
        \toprule
        \textbf{Task Type} 
        & \textbf{Task Definition} 
        & \textbf{Evaluation Mechanism} 
        & \textbf{\# Instances} \\
        \midrule
        
        \textbf{Planning} 
        & Identify the \textbf{most critical atomic ability} required to solve a given mathematical problem. 
        & MCQ: Select one essential atomic ability from five candidates. 
        & 1,448 \\

        \textbf{Action} 
        & Execute \textbf{atomic computational subtasks} involved in mathematical reasoning, emphasizing precise symbolic and numerical manipulation. 
        & MCQ: Solve calculation problems with distractors generated via operator, fraction, and numeric perturbations from ground truth. 
        & 1,101 \\
        
        \textbf{Feedback} 
        & Perform \textbf{trajectory-level verification} by judging whether a full solution process is correct. 
        & T/F: Determine the correctness of solution trajectories from given problems. 
        & 1,198 \\
        
        \bottomrule
    \end{tabularx}
    \caption{Overview of APTBench-math Tasks.}
    \label{tab:apt-math}
\end{table}

\begin{table}[htbp]
    \centering
    \small
    \renewcommand{\arraystretch}{1.5}
    \begin{tabularx}{\textwidth}{@{} l >{\RaggedRight\arraybackslash}X >{\RaggedRight\arraybackslash}X c @{}}
        \toprule
        \textbf{Task Type} 
        & \textbf{Task Definition} 
        & \textbf{Evaluation Mechanism} 
        & \textbf{\# Instances} \\
        \midrule
        
        \textbf{Tool-Select} 
        & Select the \textbf{correct tool call} from APIs based on the given context. 
        & MCQ: Select the correct tool name.
        & 894 \\
        
        \textbf{Tool-Param} 
        & Determine the \textbf{corresponding parameters} that need to be passed to the tool function.
        & TC: Write the parameters' name and value. 
        & 2,103 \\
        \bottomrule
    \end{tabularx}
    \caption{Overview of APTBench-tool tasks.}
    \label{tab:apt-tool}
\end{table}

\paragraph{DeepResearch (DR).} This scenario assesses the model's ability to synthesize information and navigate the web to answer \textbf{Closed-ended} (concise fact-seeking) and \textbf{Open-ended} (comprehensive report generation) questions. 
Similar to the coding scenario, the benchmark measures two core agent capabilities: planning and action. 
Other critical atomic abilities such as accurate citation and factual grounding are also tested.
The detailed explanation is shown in Table~\ref{tab:apt-dr}.

\paragraph{Math.}
Most existing mathematical reasoning benchmarks emphasize \textbf{end-to-end answer correctness}, offering limited insight into the reasoning process and providing little support for evaluating \textbf{agentic capabilities}. In contrast, we evaluates mathematical reasoning through three \textbf{orthogonal agentic dimensions}: planning, action, and feedback. This design enables fine-grained, capability-oriented assessment beyond final-answer accuracy, as details shown in Table \ref{tab:apt-math}.

\paragraph{Tool.}
Since general tool-use capability is a core competency of agents, we design an additional evaluation task to assess this ability in base models, as an addition to Code, DR, and Math tasks. Specifically, we adapt AceBench \citep{chen2025acebench} and BFCL V4 \citep{patil2025bfcl} into two types of problems:
The first type presents a given context and asks the model to select the appropriate tool from a predefined list, formulated as a multiple-choice question (MCQ). The second type provides both the context and the selected tool, and requires the base model to generate the corresponding tool parameters as a text completion task (TC). 
Details of the APT-Tool are shown in Table~\ref{tab:apt-tool}.

We use APT-Bench to evaluate Youtu-LLM 2B Base and a series of similarly sized LLMs. The comparison results are shown in Table~\ref{tab:agent_baseline_for_base}. Youtu-LLM 2B base's agentic capabilities are close to those of Qwen3 4B base across all dimensions, significantly outperforming other models of similar size.

\begin{table*}[h]
    \centering
    \small
    \renewcommand{\arraystretch}{1.1}
    \setlength{\tabcolsep}{6pt}
    \begin{tabular}{l|ccccc|c}
    \toprule
     \textbf{Category} & \makecell[c]{\textbf{Qwen3 1.7B}\\\textbf{Base}} & \makecell[c]{\textbf{SmolLM3 3B}\\\textbf{Base}} & \makecell[c]{\textbf{Gemma3 4B}\\\textbf{Base}} & \makecell[c]{\textbf{Qwen3 4B}\\\textbf{Base}} & \makecell[c]{\textbf{Llama3.1 8B}\\\textbf{Base}} & \makecell[c]{\textbf{Youtu-LLM 2B}\\\textbf{Base}} \\
    \midrule
    Code & 25.1 & 24.3 & 32.8 & \textbf{41.9} & 23.6 & \underline{37.9} \\
    Deep Research & 28.5 & 27.2 & 36.4 & \textbf{40.5} & 30.0 & \underline{38.6} \\
    Math & 59.9 & 60.7 & 59.8 & \textbf{70.5} & 60.1 & \underline{68.0} \\
    Tool & 56.7 & 59.1 & 61.7 & \textbf{65.8} & 64.1 & \underline{64.2} \\
    \bottomrule
    \end{tabular}
    \caption{APTBench performance of Youtu-LLM 2B Base and other similarly-sized base models.}
    \label{tab:agent_baseline_for_base}
\end{table*}

\subsubsection{Agentic Evaluation of Instruct Model}

\paragraph{Deep Research (DR).} 
We test the instruct models on two established deep research benchmarks: GAIA~\citep{mialon2023gaia} and xbench~\citep{chen2025xbench}. Our methodology integrates a standard ReAct framework~\citep{yao2022react} for each candidate model. Evaluation of both tasks was performed using the LLM-as-a-Judge paradigm, with Pass@1 as the primary metric. We adopted the evaluation protocol and implementation details of WebDancer~\citep{Wu2025WebDancerTA}, ensuring all models were assessed in thinking mode. To enhance reliability, each test case was executed four times, and the average accuracy was reported.

\paragraph{Code.} 
For the coding agentic capability, we use two benchmarks: SWE-Bench-Verified~\citep{jimenez2023swe} and EnConda-Bench~\citep{kuang2025process}. For SWE-Bench-Verified, we use R2E-Gym~\citep{jain2025r2e} and Openhands~\citep{wang2024openhands} scaffold with the maximum step of 50 for evaluation. Across all models, we employ the non-thinking mode and report the average resolve rate across 8 trials. For EnConda-Bench, we test models in non-thinking mode and report the accuracy of the fix suggestion in environment setup tasks.

\paragraph{Tool.} 
We evaluate instruct models on two established Tool benchmarks: BFCL V3~\citep{patil2025bfcl} and $\tau^2$-bench~\citep{barres2025tau2}. For BFCL V3, all models are evaluated with the FC format in the \textit{thinking mode}. And we utilize the following sampling parameters to generate responses, i.e., temperature = 1.0, top-p = 0.95, top-k = 20, and presence penalty = 1.5.
For $\tau^2$-bench, we use the non-thinking mode of the compared models with other generation parameters being kept the same with BFCL testing.

The evaluation results are presented in Table~\ref{tab:agent_results_for_Instruct}. Youtu-LLM 2B achieves superior performance compared to the best baseline on most benchmarks, with significant improvements.
It is important to note that, to date, no suitable agentic mathematical benchmark exists to effectively evaluate the corresponding capabilities of instruction-tuned models. 
This remains an open challenge in the field.
Additionally, Gemma3 4B and Llama3.1 8B do not natively support a tool-calling mechanism, which precludes their evaluation on the Deep Research and Tool tasks.

\begin{table*}[t]
    \centering
    \small
    \renewcommand{\arraystretch}{1.1}
    \setlength{\tabcolsep}{6pt}
    \begin{tabular}{l|ccc|c}
    \toprule
     \textbf{Benchmark} & \textbf{Qwen3 1.7B} & \textbf{SmolLM3 3B} & \textbf{Qwen3 4B}  & \textbf{Youtu-LLM 2B} \\
    \midrule
    \multicolumn{5}{l}{\textbf{Deep Research}} \\
    \midrule
    GAIA& 11.4 & 11.7 & \underline{25.5} & \textbf{33.9} \\
    xbench& 11.7 & 13.9 & \underline{18.4} & \textbf{19.5} \\
    \midrule
    \multicolumn{5}{l}{\textbf{Code}} \\
    \midrule
    SWE-Bench-Verified& 0.6 & \underline{7.2} & 5.7 & \textbf{17.7} \\
    EnConda-Bench& 10.8 & 3.5 & \underline{16.1} & \textbf{21.5} \\
    \midrule
    \multicolumn{5}{l}{\textbf{Tool}} \\
    \midrule
    BFCL V3 & 55.5 & 31.5 & \textbf{61.7} & \underline{58.0} \\
    $\tau^2$-Bench & 2.6 & 9.7 & \underline{10.9} & \textbf{15.0} \\
    \bottomrule
    \end{tabular}
    \caption{Agent benchmark performance of similarly-sized SOTA instruct models.} 
    \label{tab:agent_results_for_Instruct}
\end{table*}

\subsection{Analysis of Agentic Mid-Training}
\label{sec:agent_skill}

\begin{figure}[t]
    \centering
    \includegraphics[width=0.6\linewidth]{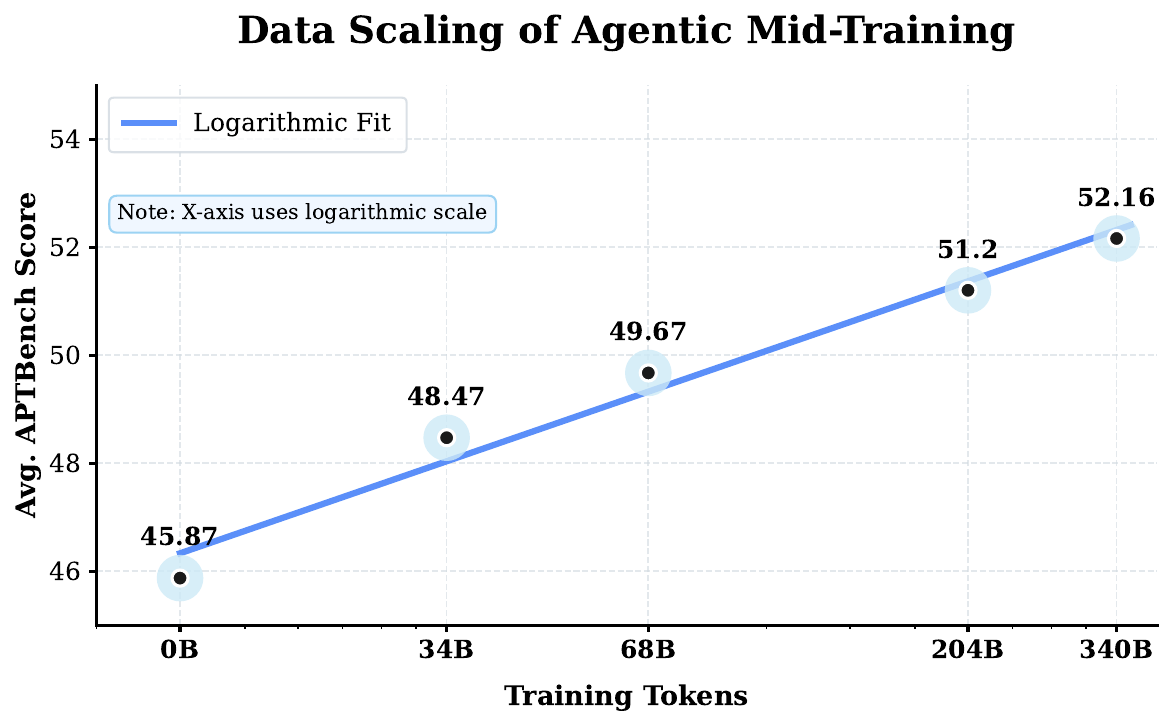}
    \caption{The impact of training data scaling on APTBench performance.}
    \label{fig:agent-scaling}
\end{figure}

As described in Section~\ref{sec:multi-stage pretrain}, we use the model after general mid-training (Stage 3) as the starting point to conduct native agentic mid-training on Youtu-LLM. 
The model’s performance is evaluated using the average APTBench score to analyze scaling behavior and training effectiveness. 
We further compare the post-trained models with and without agentic mid-training on the critical agent benchmarks.

\paragraph{Mid-Training Scaling.}
As shown in Figure~\ref{fig:agent-scaling}, we observe a clear logarithmic trend between the number of training tokens and model agentic performance. A significant improvement occurs within the first 34B tokens (approximately 20\% of the total), indicating that agentic behaviors can be efficiently acquired through agentic mid-training. Moreover, agentic mid-training yields consistent performance gains across the entire 340B-token training budget, achieving an overall improvement of more than 6\%. 
These results validate the effectiveness of our agentic mid-training paradigm.

\begin{table*}[h]
    \centering
    \small
    \renewcommand{\arraystretch}{1.1}
    \setlength{\tabcolsep}{6pt}
    \begin{tabular}{l|cccc}
    \toprule
     \textbf{Benchmark} & \textbf{Youtu-LLM 2B w/o AMT}& \textbf{Youtu-LLM 2B} & \textbf{Relative $\Delta$} & \textbf{$\Delta$ Ratio} \\
    \midrule
    GAIA & 31.1 & \textbf{33.9} & \textcolor{blue}{\textbf{+2.8}} & \textcolor{blue}{\textbf{+9.0\%}}\\
    xbench & 18.0 & \textbf{19.5} & \textcolor{blue}{\textbf{+1.5}} & \textcolor{blue}{\textbf{+8.3\%}} \\
    \midrule
    SWE-Bench-Verified& 12.4 & \textbf{17.7} & \textcolor{blue}{\textbf{+5.3}} & \textcolor{blue}{\textbf{+42.7\%}} \\
    EnConda-Bench & 19.8 & \textbf{21.5} & \textcolor{blue}{\textbf{+1.7}} & \textcolor{blue}{\textbf{+8.6\%}}\\
    \midrule
    BFCL V3 & \textbf{59.3} & 58.0 & -1.3 & -2.2\%\\
    $\tau^2$-Bench & 12.5 & \textbf{15.0} & \textcolor{blue}{\textbf{+2.5}} & \textcolor{blue}{\textbf{+20.0\%}}\\
    \bottomrule
    \end{tabular}
    \caption{Performance of post-trained models with and without agentic mid-training (AMT).}
    \label{tab:agent-mid-train-comp}
\end{table*}

To systematically investigate the impact of individual trajectory data in Section ~\ref{sec:agentic_data} on overall model training, we conduct a dedicated ablation study focusing exclusively on separate trajectory pre-training of each category, including Appendix \textsection{\ref{app:agent_reasoning}} for Agentic-CoT trajectory, Appendix \textsection{\ref{app:agent_math}} for math trajectory, Appendix \textsection{\ref{app:agent_deepresearch}} for deep research trajectory, Appendix \textsection{\ref{app:agent_code} for code trajectory and Appendix \textsection{\ref{app:agent_tool}} for tool-use trajectory. Specifically, we take the base model after completing the 8k-stage (Stage 2) training as the starting point for all ablation experiments.

\paragraph{Agentic Performance Analysis.}
The improvements brought by the \textbf{agentic mid-training} stage on APTBench also effectively transfer to downstream post-training performance.
We evaluate models with and without agentic mid-training under aligned post-training settings, as described in Section~\ref{chp:4_posttrain}, and compare their performance across multiple agent-specific evaluation benchmarks, which are widely adopted in Deep Research, software engineering, and tool-use evaluation tasks.
As shown in Table~\ref{tab:agent-mid-train-comp}, the base model equipped with agentic mid-training demonstrates significant performance improvements on most of the benchmarks.
The average relative improvement rate of the six agent benchmarks is 14.4\%, which aligns perfectly with the 13.7\% improvement on APTBench as shown in Figure~\ref{fig:agent-scaling}.

We further conduct a pass@k evaluation on SWE-Bench-Verified as an example to demonstrate the effectiveness of AMT. As illustrated in Figure~\ref{fig:pass_at_k_comparison}, the illusion of AMT yields a consistent and substantial performance uplift across all $k$ values. Specifically, at $k=1$, the AMT-enhanced model achieves a resolve rate of 17.7\%, a relative improvement of 42.7\% over the 12.4\% baseline. This margin widens as $k$ increases, reaching 33.4\% at $k=8$ compared to 26.2\% for the version without AMT. Notably, the 95\% confidence intervals for these two variants remain largely non-overlapping, particularly at higher $k$ values, signifying that the performance gains are statistically significant and not attributable to sampling variance. AMT endows the model with better long-horizon planning and error-correction priors, resulting in a model that is not only more accurate on the first attempt but scales more effectively with repeated sampling.

\begin{figure}[h]
    \centering
    \includegraphics[width=0.70\linewidth]{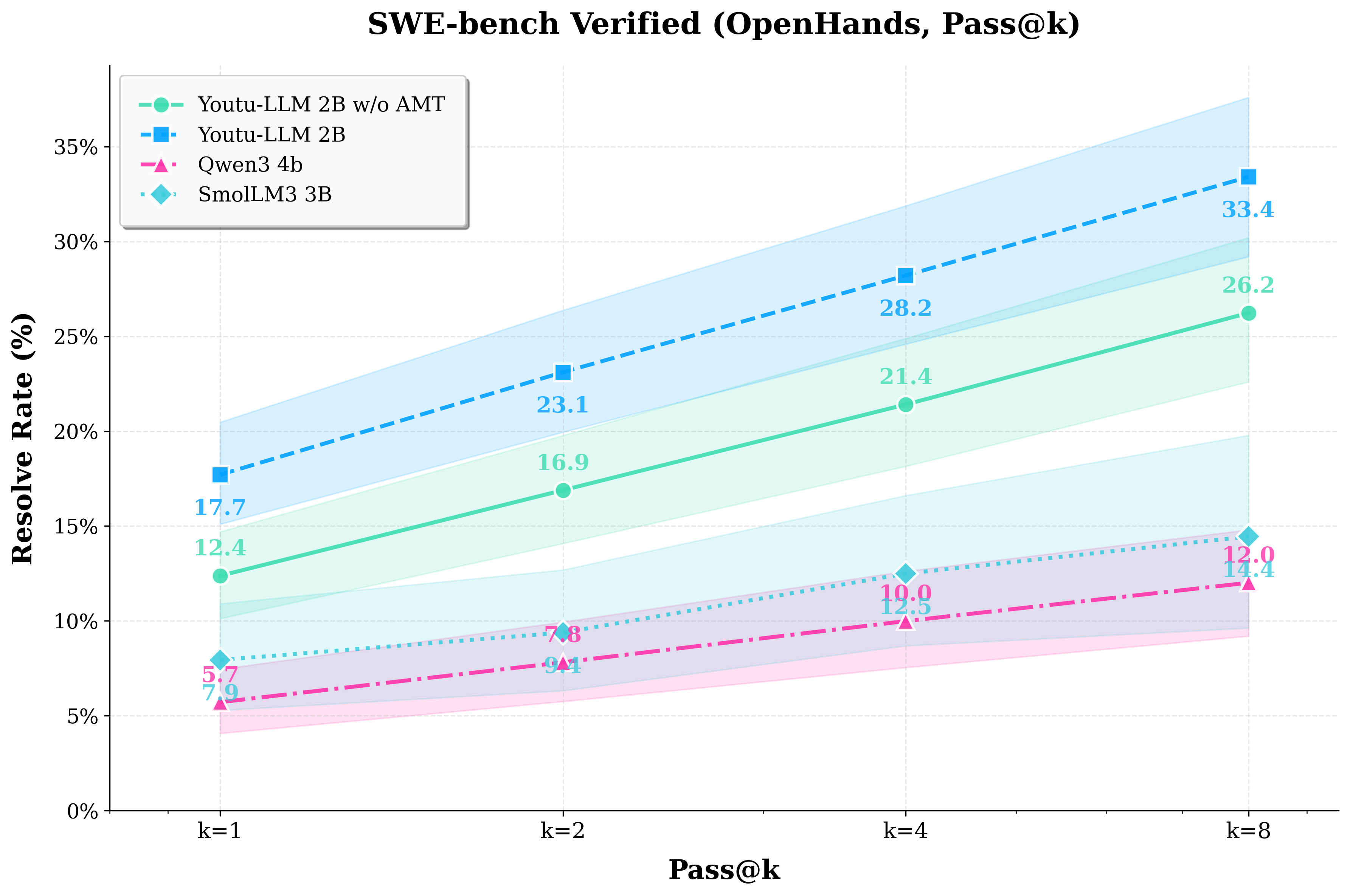}
    \caption{Pass@k comparison on SWE-Bench-Verified with Openhands scaffold across different models (with 95\% confidence intervals smeared).}
    \label{fig:pass_at_k_comparison}
\end{figure}

%% file: youtu_template.bbl
\begin{thebibliography}{95}
\providecommand{\natexlab}[1]{#1}
\providecommand{\url}[1]{\texttt{#1}}
\expandafter\ifx\csname urlstyle\endcsname\relax
  \providecommand{\doi}[1]{doi: #1}\else
  \providecommand{\doi}{doi: \begingroup \urlstyle{rm}\Url}\fi

\bibitem[Hendrycks et~al.(2025)Hendrycks, Song, Szegedy, Lee, Gal, Brynjolfsson, Li, Zou, Levine, Han, et~al.]{hendrycks2025definition}
Dan Hendrycks, Dawn Song, Christian Szegedy, Honglak Lee, Yarin Gal, Erik Brynjolfsson, Sharon Li, Andy Zou, Lionel Levine, Bo~Han, et~al.
\newblock A definition of agi.
\newblock \emph{arXiv preprint arXiv:2510.18212}, 2025.

\bibitem[Guo et~al.(2025)Guo, Yang, Zhang, Song, Zhang, Xu, Zhu, Ma, Wang, Bi, et~al.]{guo2025deepseek}
Daya Guo, Dejian Yang, Haowei Zhang, Junxiao Song, Ruoyu Zhang, Runxin Xu, Qihao Zhu, Shirong Ma, Peiyi Wang, Xiao Bi, et~al.
\newblock Deepseek-r1: Incentivizing reasoning capability in llms via reinforcement learning.
\newblock \emph{arXiv preprint arXiv:2501.12948}, 2025.

\bibitem[Yang et~al.(2025{\natexlab{a}})Yang, Li, Yang, Zhang, Hui, Zheng, Yu, Gao, Huang, Lv, et~al.]{qwen3technicalreport}
An~Yang, Anfeng Li, Baosong Yang, Beichen Zhang, Binyuan Hui, Bo~Zheng, Bowen Yu, Chang Gao, Chengen Huang, Chenxu Lv, et~al.
\newblock Qwen3 technical report, 2025{\natexlab{a}}.

\bibitem[Plaat et~al.(2025)Plaat, van Duijn, van Stein, Preuss, van~der Putten, and Batenburg]{plaat2025agentic}
Aske Plaat, Max van Duijn, Niki van Stein, Mike Preuss, Peter van~der Putten, and Kees~Joost Batenburg.
\newblock Agentic large language models, a survey.
\newblock \emph{arXiv preprint arXiv:2503.23037}, 2025.

\bibitem[Belcak et~al.(2025)Belcak, Heinrich, Diao, Fu, Dong, Muralidharan, Lin, and Molchanov]{belcak2025small}
Peter Belcak, Greg Heinrich, Shizhe Diao, Yonggan Fu, Xin Dong, Saurav Muralidharan, Yingyan~Celine Lin, and Pavlo Molchanov.
\newblock Small language models are the future of agentic ai.
\newblock \emph{arXiv preprint arXiv:2506.02153}, 2025.

\bibitem[Hu et~al.(2024)Hu, Tu, Han, He, Cui, Long, Zheng, Fang, Huang, Zhao, et~al.]{hu2024minicpm}
Shengding Hu, Yuge Tu, Xu~Han, Chaoqun He, Ganqu Cui, Xiang Long, Zhi Zheng, Yewei Fang, Yuxiang Huang, Weilin Zhao, et~al.
\newblock Minicpm: Unveiling the potential of small language models with scalable training strategies.
\newblock \emph{arXiv preprint arXiv:2404.06395}, 2024.

\bibitem[Bakouch et~al.(2025)Bakouch, Ben~Allal, Lozhkov, Tazi, Tunstall, Patiño, Beeching, Roucher, Reedi, Gallouédec, Rasul, Habib, Fourrier, Kydlicek, Penedo, Larcher, Morlon, Srivastav, Lochner, Nguyen, Raffel, von Werra, and Wolf]{bakouch2025smollm3}
Elie Bakouch, Loubna Ben~Allal, Anton Lozhkov, Nouamane Tazi, Lewis Tunstall, Carlos~Miguel Patiño, Edward Beeching, Aymeric Roucher, Aksel~Joonas Reedi, Quentin Gallouédec, Kashif Rasul, Nathan Habib, Clémentine Fourrier, Hynek Kydlicek, Guilherme Penedo, Hugo Larcher, Mathieu Morlon, Vaibhav Srivastav, Joshua Lochner, Xuan-Son Nguyen, Colin Raffel, Leandro von Werra, and Thomas Wolf.
\newblock {SmolLM3: smol, multilingual, long-context reasoner}.
\newblock \url{https://huggingface.co/blog/smollm3}, 2025.

\bibitem[Sharma and Mehta(2025)]{sharma2025small}
Raghav Sharma and Manan Mehta.
\newblock Small language models for agentic systems: A survey of architectures, capabilities, and deployment trade offs.
\newblock \emph{arXiv preprint arXiv:2510.03847}, 2025.

\bibitem[Luo et~al.(2025)Luo, Zhang, Yuan, Zhao, Yang, Gu, Wu, Chen, Qiao, Long, et~al.]{luo2025large}
Junyu Luo, Weizhi Zhang, Ye~Yuan, Yusheng Zhao, Junwei Yang, Yiyang Gu, Bohan Wu, Binqi Chen, Ziyue Qiao, Qingqing Long, et~al.
\newblock Large language model agent: A survey on methodology, applications and challenges.
\newblock \emph{arXiv preprint arXiv:2503.21460}, 2025.

\bibitem[Zeng et~al.(2025)Zeng, Lv, Zheng, Hou, Chen, Xie, Wang, Yin, Zeng, Zhang, et~al.]{zeng2025glm}
Aohan Zeng, Xin Lv, Qinkai Zheng, Zhenyu Hou, Bin Chen, Chengxing Xie, Cunxiang Wang, Da~Yin, Hao Zeng, Jiajie Zhang, et~al.
\newblock Glm-4.5: Agentic, reasoning, and coding (arc) foundation models.
\newblock \emph{arXiv preprint arXiv:2508.06471}, 2025.

\bibitem[Li et~al.(2025{\natexlab{a}})Li, Zhang, Zhang, Huang, Li, Chen, Yin, Wu, Zhou, et~al.]{team2025tongyi}
Baixuan Li, Bo~Zhang, Dingchu Zhang, Fei Huang, Guangyu Li, Guoxin Chen, Huifeng Yin, Jialong Wu, Jingren Zhou, et~al.
\newblock Tongyi deepresearch technical report.
\newblock \emph{arXiv preprint arXiv:2510.24701}, 2025{\natexlab{a}}.

\bibitem[Liu et~al.(2025{\natexlab{a}})Liu, Feng, Xue, Wang, Wu, Lu, Zhao, Deng, Zhang, Ruan, Dai, and et~al.]{deepseekai2025deepseekv3technicalreport}
Aixin Liu, Bei Feng, Bing Xue, Bingxuan Wang, Bochao Wu, Chengda Lu, Chenggang Zhao, Chengqi Deng, Chenyu Zhang, Chong Ruan, Damai Dai, and et~al.
\newblock Deepseek-v3 technical report, 2025{\natexlab{a}}.
\newblock URL \url{https://arxiv.org/abs/2412.19437}.

\bibitem[Broder(1997)]{Broder1997OnTR}
Andrei~Z. Broder.
\newblock On the resemblance and containment of documents.
\newblock \emph{Proceedings. Compression and Complexity of SEQUENCES 1997 (Cat. No.97TB100171)}, pages 21--29, 1997.
\newblock URL \url{https://api.semanticscholar.org/CorpusID:11748509}.

\bibitem[Har-Peled et~al.(2012)Har-Peled, Indyk, and Motwani]{HarPeled2012ApproximateNN}
Sariel Har-Peled, Piotr Indyk, and Rajeev Motwani.
\newblock Approximate nearest neighbor: Towards removing the curse of dimensionality.
\newblock \emph{Theory Comput.}, 8:\penalty0 321--350, 2012.
\newblock URL \url{https://api.semanticscholar.org/CorpusID:6110572}.

\bibitem[Aho and Corasick(1975)]{aho1975efficient}
Alfred~V Aho and Margaret~J Corasick.
\newblock Efficient string matching: an aid to bibliographic search.
\newblock \emph{Communications of the ACM}, 18\penalty0 (6):\penalty0 333--340, 1975.

\bibitem[Wei et~al.(2022)Wei, Wang, Schuurmans, Bosma, Xia, Chi, Le, Zhou, et~al.]{wei2022chain}
Jason Wei, Xuezhi Wang, Dale Schuurmans, Maarten Bosma, Fei Xia, Ed~Chi, Quoc~V Le, Denny Zhou, et~al.
\newblock Chain-of-thought prompting elicits reasoning in large language models.
\newblock \emph{Advances in neural information processing systems}, 35:\penalty0 24824--24837, 2022.

\bibitem[Shao et~al.(2024)Shao, Wang, Zhu, Xu, Song, Bi, Zhang, Zhang, Li, Wu, et~al.]{shao2024deepseekmath}
Zhihong Shao, Peiyi Wang, Qihao Zhu, Runxin Xu, Junxiao Song, Xiao Bi, Haowei Zhang, Mingchuan Zhang, YK~Li, Yang Wu, et~al.
\newblock Deepseekmath: Pushing the limits of mathematical reasoning in open language models.
\newblock \emph{arXiv preprint arXiv:2402.03300}, 2024.

\bibitem[Shao et~al.(2025)Shao, Luo, Lu, Ren, Hu, Ye, Gou, Ma, and Zhang]{shao2025deepseekmath}
Zhihong Shao, Yuxiang Luo, Chengda Lu, ZZ~Ren, Jiewen Hu, Tian Ye, Zhibin Gou, Shirong Ma, and Xiaokang Zhang.
\newblock Deepseekmath-v2: Towards self-verifiable mathematical reasoning.
\newblock \emph{arXiv preprint arXiv:2511.22570}, 2025.

\bibitem[Liu et~al.(2025{\natexlab{b}})Liu, Liu, Wang, Cheng, Estornell, Zhao, and Wei]{liu2025agenticmath}
Xianyang Liu, Yilin Liu, Shuai Wang, Hao Cheng, Andrew Estornell, Yuzhi Zhao, and Jiaheng Wei.
\newblock Agenticmath: Enhancing llm reasoning via agentic-based math data generation.
\newblock \emph{arXiv preprint arXiv:2510.19361}, 2025{\natexlab{b}}.

\bibitem[Gao et~al.(2025)Gao, Gu, Wu, Kong, Zhang, Cai, Zheng, Ma, Shen, Zhao, et~al.]{gao2025long}
Songyang Gao, Yuzhe Gu, Zijian Wu, Lingkai Kong, Wenwei Zhang, Zhongrui Cai, Fan Zheng, Tianyou Ma, Junhao Shen, Haiteng Zhao, et~al.
\newblock Long-horizon reasoning agent for olympiad-level mathematical problem solving.
\newblock \emph{arXiv preprint arXiv:2512.10739}, 2025.

\bibitem[Zhao et~al.(2025)Zhao, Shen, Zhang, Gao, Liu, Ma, Zheng, Lin, Zhang, and Chen]{zhao2025achieving}
Haiteng Zhao, Junhao Shen, Yiming Zhang, Songyang Gao, Kuikun Liu, Tianyou Ma, Fan Zheng, Dahua Lin, Wenwei Zhang, and Kai Chen.
\newblock Achieving olympia-level geometry large language model agent via complexity boosting reinforcement learning.
\newblock \emph{arXiv preprint arXiv:2512.10534}, 2025.

\bibitem[Gou et~al.(2024)Gou, Shao, Gong, Shen, Yang, Huang, Duan, and Chen]{ToRA2024GouSGSYHDC}
Zhibin Gou, Zhihong Shao, Yeyun Gong, Yelong Shen, Yujiu Yang, Minlie Huang, Nan Duan, and Weizhu Chen.
\newblock Tora: {A} tool-integrated reasoning agent for mathematical problem solving.
\newblock In \emph{The Twelfth International Conference on Learning Representations, {ICLR} 2024, Vienna, Austria, May 7-11, 2024}. OpenReview.net, 2024.
\newblock URL \url{https://openreview.net/forum?id=Ep0TtjVoap}.

\bibitem[Liu et~al.(2025{\natexlab{c}})Liu, Yang, Liu, Song, Gao, and Liu]{MM-Agent/corr/abs-2505-14148}
Fan Liu, Zherui Yang, Cancheng Liu, Tianrui Song, Xiaofeng Gao, and Hao Liu.
\newblock Mm-agent: {LLM} as agents for real-world mathematical modeling problem.
\newblock \emph{CoRR}, abs/2505.14148, 2025{\natexlab{c}}.
\newblock \doi{10.48550/ARXIV.2505.14148}.
\newblock URL \url{https://doi.org/10.48550/arXiv.2505.14148}.

\bibitem[Shen et~al.(2025)Shen, Huang, Yang, Wang, Gao, Xu, Jiang, He, Yang, Sun, Ju, Wu, Dai, and Dong]{REAL-Prover/corr/abs-2505-20613}
Ziju Shen, Naohao Huang, Fanyi Yang, Yutong Wang, Guoxiong Gao, Tianyi Xu, Jiedong Jiang, Wanyi He, Pu~Yang, Mengzhou Sun, Haocheng Ju, Peihao Wu, Bryan Dai, and Bin Dong.
\newblock Real-prover: Retrieval augmented lean prover for mathematical reasoning.
\newblock \emph{CoRR}, abs/2505.20613, 2025.
\newblock \doi{10.48550/ARXIV.2505.20613}.
\newblock URL \url{https://doi.org/10.48550/arXiv.2505.20613}.

\bibitem[Zhang et~al.(2025{\natexlab{a}})Zhang, Zhang, Ren, Saffat, Liu, Yang, Zhu, and Ma]{GAUSS:journals/corr/abs-2509-18122}
Yue Zhang, Jiaxin Zhang, Qiuyu Ren, Tahsin Saffat, Xiaoxuan Liu, Zitong Yang, Banghua Zhu, and Yi~Ma.
\newblock {GAUSS:} benchmarking structured mathematical skills for large language models.
\newblock \emph{CoRR}, abs/2509.18122, 2025{\natexlab{a}}.
\newblock \doi{10.48550/ARXIV.2509.18122}.
\newblock URL \url{https://doi.org/10.48550/arXiv.2509.18122}.

\bibitem[Kuang et~al.(2025{\natexlab{a}})Kuang, Huang, Li, Liang, Xu, Li, Tan, Qu, Zhang, Shen, and Yu]{kuang2025atomic}
Jiayi Kuang, Haojing Huang, Yinghui Li, Xinnian Liang, Zhikun Xu, Yangning Li, Xiaoyu Tan, Chao Qu, Meishan Zhang, Ying Shen, and Philip~S. Yu.
\newblock Atomic thinking of {LLM}s: Decoupling and exploring mathematical reasoning abilities.
\newblock In \emph{The Thirty-ninth Annual Conference on Neural Information Processing Systems}, 2025{\natexlab{a}}.
\newblock URL \url{https://openreview.net/forum?id=iBFfb6bGOz}.

\bibitem[Copet et~al.(2025)Copet, Carbonneaux, Cohen, Gehring, Kahn, Kossen, Kreuk, McMilin, Meyer, Wei, et~al.]{copet2025cwm}
Jade Copet, Quentin Carbonneaux, Gal Cohen, Jonas Gehring, Jacob Kahn, Jannik Kossen, Felix Kreuk, Emily McMilin, Michel Meyer, Yuxiang Wei, et~al.
\newblock Cwm: An open-weights llm for research on code generation with world models.
\newblock \emph{arXiv preprint arXiv:2510.02387}, 2025.

\bibitem[Pan et~al.(2024)Pan, Wang, Neubig, Jaitly, Ji, Suhr, and Zhang]{pan2024training}
Jiayi Pan, Xingyao Wang, Graham Neubig, Navdeep Jaitly, Heng Ji, Alane Suhr, and Yizhe Zhang.
\newblock Training software engineering agents and verifiers with swe-gym.
\newblock \emph{arXiv preprint arXiv:2412.21139}, 2024.

\bibitem[Yang et~al.(2025{\natexlab{b}})Yang, Lieret, Jimenez, Wettig, Khandpur, Zhang, Hui, Press, Schmidt, and Yang]{yang2025swe}
John Yang, Kilian Lieret, Carlos~E Jimenez, Alexander Wettig, Kabir Khandpur, Yanzhe Zhang, Binyuan Hui, Ofir Press, Ludwig Schmidt, and Diyi Yang.
\newblock Swe-smith: Scaling data for software engineering agents.
\newblock \emph{arXiv preprint arXiv:2504.21798}, 2025{\natexlab{b}}.

\bibitem[Badertdinov et~al.(2025)Badertdinov, Golubev, Nekrashevich, Shevtsov, Karasik, Andriushchenko, Trofimova, Litvintseva, and Yangel]{badertdinov2025swe}
Ibragim Badertdinov, Alexander Golubev, Maksim Nekrashevich, Anton Shevtsov, Simon Karasik, Andrei Andriushchenko, Maria Trofimova, Daria Litvintseva, and Boris Yangel.
\newblock Swe-rebench: An automated pipeline for task collection and decontaminated evaluation of software engineering agents.
\newblock \emph{arXiv preprint arXiv:2505.20411}, 2025.

\bibitem[Ross et~al.(2011)Ross, Gordon, and Bagnell]{ross2011reduction}
St{\'e}phane Ross, Geoffrey Gordon, and Drew Bagnell.
\newblock A reduction of imitation learning and structured prediction to no-regret online learning.
\newblock In \emph{Proceedings of the fourteenth international conference on artificial intelligence and statistics}, pages 627--635. JMLR Workshop and Conference Proceedings, 2011.

\bibitem[Youtu(2025)]{youtu-agent-2025}
Tencent Youtu.
\newblock Youtu-agent: A simple yet powerful agent framework.
\newblock \url{https://github.com/TencentCloudADP/youtu-agent}, 2025.

\bibitem[Wu et~al.(2025)Wu, Li, Fang, Yin, Zhang, Tao, Zhang, Xi, Jiang, Xie, Huang, and Zhou]{Wu2025WebDancerTA}
Jialong Wu, Baixuan Li, Runnan Fang, Wenbiao Yin, Liwen Zhang, Zhengwei Tao, Dingchu Zhang, Zekun Xi, Yong Jiang, Pengjun Xie, Fei Huang, and Jingren Zhou.
\newblock Webdancer: Towards autonomous information seeking agency.
\newblock \emph{ArXiv}, abs/2505.22648, 2025.
\newblock URL \url{https://api.semanticscholar.org/CorpusID:278959248}.

\bibitem[Xu et~al.(2025{\natexlab{a}})Xu, Soria, Tan, Roy, Agrawal, Poovendran, and Panda]{xu2025toucan}
Zhangchen Xu, Adriana~Meza Soria, Shawn Tan, Anurag Roy, Ashish~Sunil Agrawal, Radha Poovendran, and Rameswar Panda.
\newblock Toucan: Synthesizing 1.5 m tool-agentic data from real-world mcp environments.
\newblock \emph{arXiv preprint arXiv:2510.01179}, 2025{\natexlab{a}}.

\bibitem[Xu et~al.(2025{\natexlab{b}})Xu, Hao, Wang, Wen, Xu, Liu, Chen, Wang, Wang, Zhao, et~al.]{xu2025funreason}
Zengzhuang Xu, Bingguang Hao, Zechuan Wang, Yuntao Wen, Xinyi Xu, Yang Liu, Long Chen, Dong Wang, Maolin Wang, Tong Zhao, et~al.
\newblock Funreason-mt technical report: Advanced data synthesis solution for real-world multi-turn tool-use.
\newblock \emph{arXiv preprint arXiv:2510.24645}, 2025{\natexlab{b}}.

\bibitem[Sennrich et~al.(2016)Sennrich, Haddow, and Birch]{sennrich2016neural}
Rico Sennrich, Barry Haddow, and Alexandra Birch.
\newblock Neural machine translation of rare words with subword units.
\newblock In \emph{Proceedings of the 54th annual meeting of the association for computational linguistics (volume 1: long papers)}, pages 1715--1725, 2016.

\bibitem[Wang et~al.(2020)Wang, Cho, and Gu]{wang2020neural}
Changhan Wang, Kyunghyun Cho, and Jiatao Gu.
\newblock Neural machine translation with byte-level subwords.
\newblock In \emph{Proceedings of the AAAI conference on artificial intelligence}, volume~34, pages 9154--9160, 2020.

\bibitem[Liu et~al.(2025{\natexlab{d}})Liu, Hayase, Hofmann, Oh, Smith, and Choi]{liu2025superbpe}
Alisa Liu, Jonathan Hayase, Valentin Hofmann, Sewoong Oh, Noah~A Smith, and Yejin Choi.
\newblock {SuperBPE}: Space travel for language models.
\newblock In \emph{Second Conference on Language Modeling}, 2025{\natexlab{d}}.
\newblock URL \url{https://arxiv.org/abs/2503.13423}.

\bibitem[Yang et~al.(2025{\natexlab{c}})Yang, Hu, Kang, Lin, and Zhang]{yang2025number}
Haotong Yang, Yi~Hu, Shijia Kang, Zhouchen Lin, and Muhan Zhang.
\newblock Number cookbook: Number understanding of language models and how to improve it.
\newblock In \emph{The Thirteenth International Conference on Learning Representations}, 2025{\natexlab{c}}.
\newblock URL \url{https://openreview.net/forum?id=BWS5gVjgeY}.

\bibitem[Ding et~al.(2025)Ding, McLeish, Meidani, Melnyk, Nguyen, Bruss, and Huang]{ding2025you}
Mucong Ding, Sean~Michael McLeish, Kazem Meidani, Igor Melnyk, Nam~H Nguyen, C~Bayan Bruss, and Furong Huang.
\newblock You only train once: Efficient tokenizer selection for arithmetic in language models.
\newblock In \emph{Tokenization Workshop}, 2025.
\newblock URL \url{https://openreview.net/forum?id=syvdPYbdTI}.

\bibitem[Zhang et~al.(2025{\natexlab{b}})Zhang, Wang, Qian, Yan, Zhang, Xu, Li, Huang, Li, and Qiu]{zhang2025speculating}
Qingjie Zhang, Di~Wang, Haoting Qian, Liu Yan, Tianwei Zhang, Ke~Xu, Qi~Li, Minlie Huang, Hewu Li, and Han Qiu.
\newblock Speculating llms’ chinese training data pollution from their tokens.
\newblock In \emph{Proceedings of the 2025 Conference on Empirical Methods in Natural Language Processing}, pages 26124--26144, 2025{\natexlab{b}}.

\bibitem[Grattafiori et~al.(2024)Grattafiori, Dubey, Jauhri, Pandey, Kadian, Al-Dahle, Letman, Mathur, Schelten, Vaughan, Yang, Fan, Goyal, and et~al.]{grattafiori2024llama3herdmodels}
Aaron Grattafiori, Abhimanyu Dubey, Abhinav Jauhri, Abhinav Pandey, Abhishek Kadian, Ahmad Al-Dahle, Aiesha Letman, Akhil Mathur, Alan Schelten, Alex Vaughan, Amy Yang, Angela Fan, Anirudh Goyal, and et~al.
\newblock The llama 3 herd of models, 2024.
\newblock URL \url{https://arxiv.org/abs/2407.21783}.

\bibitem[Bai et~al.(2023)Bai, Bai, Chu, Cui, Dang, Deng, Fan, Ge, Han, Huang, Hui, Ji, Li, Lin, Lin, Liu, and et~al.]{bai2023qwentechnicalreport}
Jinze Bai, Shuai Bai, Yunfei Chu, Zeyu Cui, Kai Dang, Xiaodong Deng, Yang Fan, Wenbin Ge, Yu~Han, Fei Huang, Binyuan Hui, Luo Ji, Mei Li, Junyang Lin, Runji Lin, Dayiheng Liu, and et~al.
\newblock Qwen technical report, 2023.
\newblock URL \url{https://arxiv.org/abs/2309.16609}.

\bibitem[Liu et~al.(2024{\natexlab{a}})Liu, Feng, Wang, Wang, Liu, Zhao, Dengr, Ruan, Dai, Guo, Yang, Chen, Ji, and et~al.]{deepseekai2024deepseekv2strongeconomicalefficient}
Aixin Liu, Bei Feng, Bin Wang, Bingxuan Wang, Bo~Liu, Chenggang Zhao, Chengqi Dengr, Chong Ruan, Damai Dai, Daya Guo, Dejian Yang, Deli Chen, Dongjie Ji, and et~al.
\newblock Deepseek-v2: A strong, economical, and efficient mixture-of-experts language model, 2024{\natexlab{a}}.
\newblock URL \url{https://arxiv.org/abs/2405.04434}.

\bibitem[Cai et~al.(2025)Cai, Jiang, Wang, Tang, Kim, and Huang]{cai2025surveymixtureexpertslarge}
Weilin Cai, Juyong Jiang, Fan Wang, Jing Tang, Sunghun Kim, and Jiayi Huang.
\newblock A survey on mixture of experts in large language models, 2025.
\newblock URL \url{https://arxiv.org/abs/2407.06204}.

\bibitem[Yi et~al.(2023)Yi, Guo, Wei, Zhou, Wang, and Xu]{yi2023edgemoe}
Rongjie Yi, Liwei Guo, Shiyun Wei, Ao~Zhou, Shangguang Wang, and Mengwei Xu.
\newblock Edgemoe: Fast on-device inference of moe-based large language models.
\newblock \emph{arXiv preprint arXiv:2308.14352}, 2023.

\bibitem[Bai et~al.(2025{\natexlab{a}})Bai, Bao, Chen, Chen, Chen, Chen, Chen, Chen, Chen, Chen, Cui, and et~al.]{kimiteam2025kimik2openagentic}
Yifan Bai, Yiping Bao, Guanduo Chen, Jiahao Chen, Ningxin Chen, Ruijue Chen, Yanru Chen, Yuankun Chen, Yutian Chen, Zhuofu Chen, Jialei Cui, and et~al.
\newblock Kimi k2: Open agentic intelligence, 2025{\natexlab{a}}.
\newblock URL \url{https://arxiv.org/abs/2507.20534}.

\bibitem[Bruner(1960)]{bruner1960process}
Jerome~S Bruner.
\newblock The process of education, harvard, univ.
\newblock \emph{Press, Cambridge, Mass}, 1960.

\bibitem[Ausubel(1960)]{ausubel1960use}
David~P Ausubel.
\newblock The use of advance organizers in the learning and retention of meaningful verbal material.
\newblock \emph{Journal of educational psychology}, 51\penalty0 (5):\penalty0 267, 1960.

\bibitem[Walsh et~al.(2024)Walsh, Soldaini, Groeneveld, Lo, Arora, Bhagia, Gu, Huang, Jordan, et~al.]{olmo20242}
Pete Walsh, Luca Soldaini, Dirk Groeneveld, Kyle Lo, Shane Arora, Akshita Bhagia, Yuling Gu, Shengyi Huang, Matt Jordan, et~al.
\newblock 2 olmo 2 furious.
\newblock \emph{arXiv preprint arXiv:2501.00656}, 2024.

\bibitem[Guha et~al.(2025)Guha, Marten, Keh, Raoof, Smyrnis, Bansal, Nezhurina, Mercat, Vu, Sprague, Suvarna, Feuer, Chen, Khan, Frankel, Grover, Choi, Muennighoff, Su, Zhao, Yang, Pimpalgaonkar, Sharma, Ji, Deng, Pratt, Ramanujan, Saad-Falcon, Li, Dave, Albalak, Arora, Wulfe, Hegde, Durrett, Oh, Bansal, Gabriel, Grover, Chang, Shankar, Gokaslan, Merrill, Hashimoto, Choi, Jitsev, Heckel, Sathiamoorthy, Dimakis, and Schmidt]{guha2025openthoughtsdatarecipesreasoning}
Etash Guha, Ryan Marten, Sedrick Keh, Negin Raoof, Georgios Smyrnis, Hritik Bansal, Marianna Nezhurina, Jean Mercat, Trung Vu, Zayne Sprague, Ashima Suvarna, Benjamin Feuer, Liangyu Chen, Zaid Khan, Eric Frankel, Sachin Grover, Caroline Choi, Niklas Muennighoff, Shiye Su, Wanjia Zhao, John Yang, Shreyas Pimpalgaonkar, Kartik Sharma, Charlie Cheng-Jie Ji, Yichuan Deng, Sarah Pratt, Vivek Ramanujan, Jon Saad-Falcon, Jeffrey Li, Achal Dave, Alon Albalak, Kushal Arora, Blake Wulfe, Chinmay Hegde, Greg Durrett, Sewoong Oh, Mohit Bansal, Saadia Gabriel, Aditya Grover, Kai-Wei Chang, Vaishaal Shankar, Aaron Gokaslan, Mike~A. Merrill, Tatsunori Hashimoto, Yejin Choi, Jenia Jitsev, Reinhard Heckel, Maheswaran Sathiamoorthy, Alexandros~G. Dimakis, and Ludwig Schmidt.
\newblock Openthoughts: Data recipes for reasoning models, 2025.
\newblock URL \url{https://arxiv.org/abs/2506.04178}.

\bibitem[Liu et~al.(2025{\natexlab{e}})Liu, Mei, Lin, Xue, Wang, Xu, Wu, Zhang, Lin, Dong, et~al.]{liu2025deepseek}
Aixin Liu, Aoxue Mei, Bangcai Lin, Bing Xue, Bingxuan Wang, Bingzheng Xu, Bochao Wu, Bowei Zhang, Chaofan Lin, Chen Dong, et~al.
\newblock Deepseek-v3. 2: Pushing the frontier of open large language models.
\newblock \emph{arXiv preprint arXiv:2512.02556}, 2025{\natexlab{e}}.

\bibitem[Yang et~al.(2025{\natexlab{d}})Yang, Peng, Zhu, Le, Feng, Zhang, Ruan, Liu, Cheng, Xu, et~al.]{yang2025nanbeige4}
Chen Yang, Guangyue Peng, Jiaying Zhu, Ran Le, Ruixiang Feng, Tao Zhang, Wei Ruan, Xiaoqi Liu, Xiaoxue Cheng, Xiyun Xu, et~al.
\newblock Nanbeige4-3b technical report: Exploring the frontier of small language models.
\newblock \emph{arXiv preprint arXiv:2512.06266}, 2025{\natexlab{d}}.

\bibitem[He et~al.(2025)He, Liu, Liu, Yan, Wang, Cheng, Zhang, Zhang, Xu, Shen, et~al.]{he2025skywork}
Jujie He, Jiacai Liu, Chris~Yuhao Liu, Rui Yan, Chaojie Wang, Peng Cheng, Xiaoyu Zhang, Fuxiang Zhang, Jiacheng Xu, Wei Shen, et~al.
\newblock Skywork open reasoner 1 technical report.
\newblock \emph{arXiv preprint arXiv:2505.22312}, 2025.

\bibitem[Li et~al.(2025{\natexlab{b}})Li, Guo, Yang, Xu, Wu, and He]{li2025codei}
Junlong Li, Daya Guo, Dejian Yang, Runxin Xu, Yu~Wu, and Junxian He.
\newblock Codei/o: Condensing reasoning patterns via code input-output prediction.
\newblock \emph{arXiv preprint arXiv:2502.07316}, 2025{\natexlab{b}}.

\bibitem[Gunjal et~al.(2025)Gunjal, Wang, Lau, Nath, He, Liu, and Hendryx]{gunjal2025rubrics}
Anisha Gunjal, Anthony Wang, Elaine Lau, Vaskar Nath, Yunzhong He, Bing Liu, and Sean Hendryx.
\newblock Rubrics as rewards: Reinforcement learning beyond verifiable domains.
\newblock \emph{arXiv preprint arXiv:2507.17746}, 2025.

\bibitem[Yu et~al.(2025)Yu, Zhang, Zhu, Yuan, Zuo, Yue, Dai, Fan, Liu, Liu, et~al.]{yu2025dapo}
Qiying Yu, Zheng Zhang, Ruofei Zhu, Yufeng Yuan, Xiaochen Zuo, Yu~Yue, Weinan Dai, Tiantian Fan, Gaohong Liu, Lingjun Liu, et~al.
\newblock Dapo: An open-source llm reinforcement learning system at scale.
\newblock \emph{arXiv preprint arXiv:2503.14476}, 2025.

\bibitem[Yao et~al.(2025)Yao, Liu, Zhang, Dong, Shang, and Gao]{yao2025offpolicy}
Feng Yao, Liyuan Liu, Dinghuai Zhang, Chengyu Dong, Jingbo Shang, and Jianfeng Gao.
\newblock Your efficient rl framework secretly brings you off-policy rl training, August 2025.
\newblock URL \url{https://fengyao.notion.site/off-policy-rl}.

\bibitem[Liu et~al.(2025{\natexlab{f}})Liu, Li, Fu, Wang, Liu, and Shen]{liu-li-2025-rl-collapse}
Jiacai Liu, Yingru Li, Yuqian Fu, Jiawei Wang, Qian Liu, and Yu~Shen.
\newblock When speed kills stability: Demystifying {RL} collapse from the training-inference mismatch, September 2025{\natexlab{f}}.
\newblock URL \url{https://richardli.xyz/rl-collapse}.

\bibitem[Qi et~al.(2025)Qi, Liu, Zhou, Pang, Du, Lee, and Lin]{qi2025defeating}
Penghui Qi, Zichen Liu, Xiangxin Zhou, Tianyu Pang, Chao Du, Wee~Sun Lee, and Min Lin.
\newblock Defeating the training-inference mismatch via fp16.
\newblock \emph{arXiv preprint arXiv:2510.26788}, 2025.

\bibitem[Lewis et~al.(2020)Lewis, Oguz, Rinott, Riedel, and Schwenk]{lewis2020mlqa}
Patrick Lewis, Barlas Oguz, Ruty Rinott, Sebastian Riedel, and Holger Schwenk.
\newblock Mlqa: Evaluating cross-lingual extractive question answering.
\newblock In \emph{Proceedings of the 58th annual meeting of the association for computational linguistics}, pages 7315--7330, 2020.

\bibitem[Xuan et~al.(2025)Xuan, Yang, Qi, Zeng, Xiao, Feng, Liu, Xing, Wang, Gao, et~al.]{xuan2025mmlu}
Weihao Xuan, Rui Yang, Heli Qi, Qingcheng Zeng, Yunze Xiao, Aosong Feng, Dairui Liu, Yun Xing, Junjue Wang, Fan Gao, et~al.
\newblock Mmlu-prox: A multilingual benchmark for advanced large language model evaluation.
\newblock \emph{arXiv preprint arXiv:2503.10497}, 2025.

\bibitem[Wang et~al.(2024{\natexlab{a}})Wang, Ma, Zhang, Ni, Chandra, Guo, Ren, Arulraj, He, Jiang, et~al.]{wang2024mmlu}
Yubo Wang, Xueguang Ma, Ge~Zhang, Yuansheng Ni, Abhranil Chandra, Shiguang Guo, Weiming Ren, Aaran Arulraj, Xuan He, Ziyan Jiang, et~al.
\newblock Mmlu-pro: A more robust and challenging multi-task language understanding benchmark.
\newblock \emph{Advances in Neural Information Processing Systems}, 37:\penalty0 95266--95290, 2024{\natexlab{a}}.

\bibitem[Cobbe et~al.(2021)Cobbe, Kosaraju, Bavarian, Chen, Jun, Kaiser, Plappert, Tworek, Hilton, Nakano, et~al.]{cobbe2021training}
Karl Cobbe, Vineet Kosaraju, Mohammad Bavarian, Mark Chen, Heewoo Jun, Lukasz Kaiser, Matthias Plappert, Jerry Tworek, Jacob Hilton, Reiichiro Nakano, et~al.
\newblock Training verifiers to solve math word problems.
\newblock \emph{arXiv preprint arXiv:2110.14168}, 2021.

\bibitem[Shi et~al.(2022)Shi, Suzgun, Freitag, Wang, Srivats, Vosoughi, Chung, Tay, Ruder, Zhou, et~al.]{shi2022language}
Freda Shi, Mirac Suzgun, Markus Freitag, Xuezhi Wang, Suraj Srivats, Soroush Vosoughi, Hyung~Won Chung, Yi~Tay, Sebastian Ruder, Denny Zhou, et~al.
\newblock Language models are multilingual chain-of-thought reasoners.
\newblock \emph{arXiv preprint arXiv:2210.03057}, 2022.

\bibitem[Hendrycks et~al.(2021)Hendrycks, Burns, Kadavath, Arora, Basart, Tang, Song, and Steinhardt]{hendrycks2021measuring}
Dan Hendrycks, Collin Burns, Saurav Kadavath, Akul Arora, Steven Basart, Eric Tang, Dawn Song, and Jacob Steinhardt.
\newblock Measuring mathematical problem solving with the math dataset.
\newblock \emph{arXiv preprint arXiv:2103.03874}, 2021.

\bibitem[Suzgun et~al.(2023)Suzgun, Scales, Sch{\"a}rli, Gehrmann, Tay, Chung, Chowdhery, Le, Chi, Zhou, et~al.]{suzgun2023challenging}
Mirac Suzgun, Nathan Scales, Nathanael Sch{\"a}rli, Sebastian Gehrmann, Yi~Tay, Hyung~Won Chung, Aakanksha Chowdhery, Quoc Le, Ed~Chi, Denny Zhou, et~al.
\newblock Challenging big-bench tasks and whether chain-of-thought can solve them.
\newblock In \emph{Findings of the Association for Computational Linguistics: ACL 2023}, pages 13003--13051, 2023.

\bibitem[Rein et~al.(2024)Rein, Hou, Stickland, Petty, Pang, Dirani, Michael, and Bowman]{rein2024gpqa}
David Rein, Betty~Li Hou, Asa~Cooper Stickland, Jackson Petty, Richard~Yuanzhe Pang, Julien Dirani, Julian Michael, and Samuel~R Bowman.
\newblock Gpqa: A graduate-level google-proof q\&a benchmark.
\newblock In \emph{First Conference on Language Modeling}, 2024.

\bibitem[Phan et~al.(2025)Phan, Gatti, Han, Li, Hu, Zhang, Zhang, Shaaban, Ling, Shi, et~al.]{phan2025humanity}
Long Phan, Alice Gatti, Ziwen Han, Nathaniel Li, Josephina Hu, Hugh Zhang, Chen Bo~Calvin Zhang, Mohamed Shaaban, John Ling, Sean Shi, et~al.
\newblock Humanity's last exam.
\newblock \emph{arXiv preprint arXiv:2501.14249}, 2025.

\bibitem[Austin et~al.(2021)Austin, Odena, Nye, Bosma, Michalewski, Dohan, Jiang, Cai, Terry, Le, et~al.]{austin2021program}
Jacob Austin, Augustus Odena, Maxwell Nye, Maarten Bosma, Henryk Michalewski, David Dohan, Ellen Jiang, Carrie Cai, Michael Terry, Quoc Le, et~al.
\newblock Program synthesis with large language models.
\newblock \emph{arXiv preprint arXiv:2108.07732}, 2021.

\bibitem[Liu et~al.(2023)Liu, Xia, Wang, and Zhang]{liu2023your}
Jiawei Liu, Chunqiu~Steven Xia, Yuyao Wang, and Lingming Zhang.
\newblock Is your code generated by chatgpt really correct? rigorous evaluation of large language models for code generation.
\newblock \emph{Advances in Neural Information Processing Systems}, 36:\penalty0 21558--21572, 2023.

\bibitem[Chen(2021)]{chen2021evaluating}
Mark Chen.
\newblock Evaluating large language models trained on code.
\newblock \emph{arXiv preprint arXiv:2107.03374}, 2021.

\bibitem[Jain et~al.(2024)Jain, Han, Gu, Li, Yan, Zhang, Wang, Solar-Lezama, Sen, and Stoica]{jain2024livecodebench}
Naman Jain, King Han, Alex Gu, Wen-Ding Li, Fanjia Yan, Tianjun Zhang, Sida Wang, Armando Solar-Lezama, Koushik Sen, and Ion Stoica.
\newblock Livecodebench: Holistic and contamination free evaluation of large language models for code.
\newblock \emph{arXiv preprint arXiv:2403.07974}, 2024.

\bibitem[Gu et~al.(2024)Gu, Rozière, Leather, Solar-Lezama, Synnaeve, and Wang]{gu2024cruxeval}
Alex Gu, Baptiste Rozière, Hugh Leather, Armando Solar-Lezama, Gabriel Synnaeve, and Sida~I. Wang.
\newblock Cruxeval: A benchmark for code reasoning, understanding and execution.
\newblock \emph{arXiv preprint arXiv:2401.03065}, 2024.

\bibitem[Liu et~al.(2024{\natexlab{b}})Liu, Xu, and McAuley]{liu2023repobench}
Tianyang Liu, Canwen Xu, and Julian McAuley.
\newblock Repobench: Benchmarking repository-level code auto-completion systems, 2024{\natexlab{b}}.
\newblock URL \url{https://arxiv.org/abs/2306.03091}.

\bibitem[Bai et~al.(2025{\natexlab{b}})Bai, Tu, Zhang, Peng, Wang, Lv, Cao, Xu, Hou, Dong, et~al.]{bai2025longbench}
Yushi Bai, Shangqing Tu, Jiajie Zhang, Hao Peng, Xiaozhi Wang, Xin Lv, Shulin Cao, Jiazheng Xu, Lei Hou, Yuxiao Dong, et~al.
\newblock Longbench v2: Towards deeper understanding and reasoning on realistic long-context multitasks.
\newblock In \emph{Proceedings of the 63rd Annual Meeting of the Association for Computational Linguistics (Volume 1: Long Papers)}, pages 3639--3664, 2025{\natexlab{b}}.

\bibitem[Hsieh et~al.(2024)Hsieh, Sun, Kriman, Acharya, Rekesh, Jia, Zhang, and Ginsburg]{hsieh2024ruler}
Cheng-Ping Hsieh, Simeng Sun, Samuel Kriman, Shantanu Acharya, Dima Rekesh, Fei Jia, Yang Zhang, and Boris Ginsburg.
\newblock Ruler: What's the real context size of your long-context language models?
\newblock \emph{arXiv preprint arXiv:2404.06654}, 2024.

\bibitem[Yen et~al.(2024)Yen, Gao, Hou, Ding, Fleischer, Izsak, Wasserblat, and Chen]{yen2024helmet}
Howard Yen, Tianyu Gao, Minmin Hou, Ke~Ding, Daniel Fleischer, Peter Izsak, Moshe Wasserblat, and Danqi Chen.
\newblock Helmet: How to evaluate long-context language models effectively and thoroughly.
\newblock \emph{arXiv preprint arXiv:2410.02694}, 2024.

\bibitem[Gema et~al.(2025)Gema, Leang, Hong, Devoto, Mancino, Saxena, He, Zhao, Du, Madani, et~al.]{gema2025we}
Aryo~Pradipta Gema, Joshua Ong~Jun Leang, Giwon Hong, Alessio Devoto, Alberto Carlo~Maria Mancino, Rohit Saxena, Xuanli He, Yu~Zhao, Xiaotang Du, Mohammad Reza~Ghasemi Madani, et~al.
\newblock Are we done with mmlu?
\newblock In \emph{Proceedings of the 2025 Conference of the Nations of the Americas Chapter of the Association for Computational Linguistics: Human Language Technologies (Volume 1: Long Papers)}, pages 5069--5096, 2025.

\bibitem[Zhou et~al.(2023)Zhou, Lu, Mishra, Brahma, Basu, Luan, Zhou, and Hou]{zhou2023instructionfollowingevaluationlargelanguage}
Jeffrey Zhou, Tianjian Lu, Swaroop Mishra, Siddhartha Brahma, Sujoy Basu, Yi~Luan, Denny Zhou, and Le~Hou.
\newblock Instruction-following evaluation for large language models, 2023.
\newblock URL \url{https://arxiv.org/abs/2311.07911}.

\bibitem[Dua et~al.(2019)Dua, Wang, Dasigi, Stanovsky, Singh, and Gardner]{dua2019drop}
Dheeru Dua, Yizhong Wang, Pradeep Dasigi, Gabriel Stanovsky, Sameer Singh, and Matt Gardner.
\newblock Drop: A reading comprehension benchmark requiring discrete reasoning over paragraphs.
\newblock \emph{arXiv preprint arXiv:1903.00161}, 2019.

\bibitem[Sprague et~al.(2023)Sprague, Ye, Bostrom, Chaudhuri, and Durrett]{sprague2023musr}
Zayne Sprague, Xi~Ye, Kaj Bostrom, Swarat Chaudhuri, and Greg Durrett.
\newblock Musr: Testing the limits of chain-of-thought with multistep soft reasoning.
\newblock \emph{arXiv preprint arXiv:2310.16049}, 2023.

\bibitem[Lightman et~al.(2023)Lightman, Kosaraju, Burda, Edwards, Baker, Lee, Leike, Schulman, Sutskever, and Cobbe]{lightman2023let}
Hunter Lightman, Vineet Kosaraju, Yuri Burda, Harrison Edwards, Bowen Baker, Teddy Lee, Jan Leike, John Schulman, Ilya Sutskever, and Karl Cobbe.
\newblock Let's verify step by step.
\newblock In \emph{The Twelfth International Conference on Learning Representations}, 2023.

\bibitem[AIME(2025)]{aime_2024_2025}
AIME.
\newblock Aime problems and solutions.
\newblock \url{https://artofproblemsolving.com/wiki/index.php/AIME_Problems_and_Solutions}, 2025.

\bibitem[Qin et~al.(2025)Qin, Xi, Huang, Rui, Yin, Liu, Yu, Zhang, and Sun]{qin2025aptbench}
Jiarui Qin, Yunjia Xi, Junjie Huang, Renting Rui, Di~Yin, Weiwen Liu, Yong Yu, Weinan Zhang, and Xing Sun.
\newblock Aptbench: Benchmarking agentic potential of base llms during pre-training.
\newblock \emph{arXiv preprint arXiv:2510.24397}, 2025.

\bibitem[Chen et~al.(2025{\natexlab{a}})Chen, Hao, Liu, Huang, Zeng, Yu, Li, Wang, Gan, Huang, et~al.]{chen2025acebench}
Chen Chen, Xinlong Hao, Weiwen Liu, Xu~Huang, Xingshan Zeng, Shuai Yu, Dexun Li, Shuai Wang, Weinan Gan, Yuefeng Huang, et~al.
\newblock Acebench: Who wins the match point in tool usage?
\newblock \emph{arXiv preprint arXiv:2501.12851}, 2025{\natexlab{a}}.

\bibitem[Patil et~al.(2025)Patil, Mao, Cheng-Jie~Ji, Yan, Suresh, Stoica, and E.~Gonzalez]{patil2025bfcl}
Shishir~G. Patil, Huanzhi Mao, Charlie Cheng-Jie~Ji, Fanjia Yan, Vishnu Suresh, Ion Stoica, and Joseph E.~Gonzalez.
\newblock The berkeley function calling leaderboard (bfcl): From tool use to agentic evaluation of large language models.
\newblock In \emph{Forty-second International Conference on Machine Learning}, 2025.

\bibitem[Mialon et~al.(2023)Mialon, Fourrier, Wolf, LeCun, and Scialom]{mialon2023gaia}
Gr{\'e}goire Mialon, Cl{\'e}mentine Fourrier, Thomas Wolf, Yann LeCun, and Thomas Scialom.
\newblock Gaia: a benchmark for general ai assistants.
\newblock In \emph{The Twelfth International Conference on Learning Representations}, 2023.

\bibitem[Chen et~al.(2025{\natexlab{b}})Chen, Ren, Liu, Hu, Tian, Xie, Liu, Zhang, Liu, Gong, et~al.]{chen2025xbench}
Kaiyuan Chen, Yixin Ren, Yang Liu, Xiaobo Hu, Haotong Tian, Tianbao Xie, Fangfu Liu, Haoye Zhang, Hongzhang Liu, Yuan Gong, et~al.
\newblock xbench: Tracking agents productivity scaling with profession-aligned real-world evaluations.
\newblock \emph{arXiv preprint arXiv:2506.13651}, 2025{\natexlab{b}}.

\bibitem[Yao et~al.(2022)Yao, Zhao, Yu, Du, Shafran, Narasimhan, and Cao]{yao2022react}
Shunyu Yao, Jeffrey Zhao, Dian Yu, Nan Du, Izhak Shafran, Karthik~R Narasimhan, and Yuan Cao.
\newblock React: Synergizing reasoning and acting in language models.
\newblock In \emph{The eleventh international conference on learning representations}, 2022.

\bibitem[Jimenez et~al.(2023)Jimenez, Yang, Wettig, Yao, Pei, Press, and Narasimhan]{jimenez2023swe}
Carlos~E Jimenez, John Yang, Alexander Wettig, Shunyu Yao, Kexin Pei, Ofir Press, and Karthik Narasimhan.
\newblock Swe-bench: Can language models resolve real-world github issues?
\newblock \emph{arXiv preprint arXiv:2310.06770}, 2023.

\bibitem[Kuang et~al.(2025{\natexlab{b}})Kuang, Li, Zhang, Li, Yin, Sun, Shen, and Yu]{kuang2025process}
Jiayi Kuang, Yinghui Li, Xin Zhang, Yangning Li, Di~Yin, Xing Sun, Ying Shen, and Philip~S Yu.
\newblock Process-level trajectory evaluation for environment configuration in software engineering agents.
\newblock \emph{arXiv preprint arXiv:2510.25694}, 2025{\natexlab{b}}.

\bibitem[Jain et~al.(2025)Jain, Singh, Shetty, Zhang, Zheng, Sen, and Stoica]{jain2025r2e}
Naman Jain, Jaskirat Singh, Manish Shetty, Tianjun Zhang, Liang Zheng, Koushik Sen, and Ion Stoica.
\newblock R2e-gym: Procedural environment generation and hybrid verifiers for scaling open-weights swe agents.
\newblock In \emph{Second Conference on Language Modeling}, 2025.

\bibitem[Wang et~al.(2024{\natexlab{b}})Wang, Li, Song, Xu, Tang, Zhuge, Pan, Song, Li, Singh, et~al.]{wang2024openhands}
Xingyao Wang, Boxuan Li, Yufan Song, Frank~F Xu, Xiangru Tang, Mingchen Zhuge, Jiayi Pan, Yueqi Song, Bowen Li, Jaskirat Singh, et~al.
\newblock Openhands: An open platform for ai software developers as generalist agents.
\newblock \emph{arXiv preprint arXiv:2407.16741}, 2024{\natexlab{b}}.

\bibitem[Barres et~al.(2025)Barres, Dong, Ray, Si, and Narasimhan]{barres2025tau2}
Victor Barres, Honghua Dong, Soham Ray, Xujie Si, and Karthik Narasimhan.
\newblock $\tau^2$-bench: Evaluating conversational agents in a dual-control environment, 2025.
\newblock URL \url{https://arxiv.org/abs/2506.07982}.

\end{thebibliography}
